\documentclass[10pt,twocolumn,letterpaper]{article}

\usepackage[pagenumbers]{cvpr} 

\usepackage{bm}
\usepackage{xspace}
\def\etal{et al.\xspace}
\def\eg{e.g.\xspace}
\def\ie{i.e.\xspace}

\DeclareMathOperator{\erf}{erf}

\newcommand{\dd}{\mathrm{d}}
\newcommand{\pos}{\bm{x}}

\newcommand{\origin}{\bm{o}}
\newcommand{\dir}{\bm{d}}
\newcommand{\mean}{\bm{\mu}}
\newcommand{\rot}{\bm{R}}
\newcommand{\scale}{\bm{S}}
\newcommand{\cov}{\bm{\Sigma}}
\newcommand{\invcov}{\cov^{-1}}

\newcommand{\m}{\mu}
\newcommand{\s}{\Sigma}
\newcommand{\w}{\omega}
\newcommand{\mi}{\m_i}
\newcommand{\si}{\s_i}
\newcommand{\wi}{\w_i}
\newcommand{\mj}{\m_j}
\newcommand{\sj}{\s_j}
\newcommand{\wj}{\w_j}

\let\originalleft\left
\let\originalright\right
\renewcommand{\left}{\mathopen{}\mathclose\bgroup\originalleft}
\renewcommand{\right}{\aftergroup\egroup\originalright}

\usepackage{tikz}
\usetikzlibrary{spy}
\usetikzlibrary{arrows.meta} 
\usetikzlibrary{calc} 

\usepackage{booktabs}
\usepackage{multirow}
\usepackage{graphicx}

\definecolor{Best}{RGB}{255,235,150}     
\definecolor{Second}{RGB}{220,240,255}   

\newcommand{\imagewithzoom}[8]{%
    \begin{tikzpicture}[spy using outlines={lens={scale=#7}, size=#6}]
        \node[draw=black, line width=2pt, inner sep=0pt] at (0, 0)  
            {\includegraphics[width=\textwidth, keepaspectratio, trim={#8}, clip]{#1}};
        \spy [red] on (#2,#3) in node at (#4,#5);
    \end{tikzpicture}%
}

\definecolor{cvprblue}{rgb}{0.21,0.49,0.74}
\usepackage[pagebackref,breaklinks,colorlinks,allcolors=cvprblue]{hyperref}

\def\paperID{} 
\def\confName{}
\def\confYear{}

\title{Moment-Based 3D Gaussian Splatting: Resolving Volumetric Occlusion with Order-Independent Transmittance}

\author{Jan U. Müller
\quad
Robin Tim Landsgesell
\quad
Leif Van Holland 
\quad
Patrick Stotko 
\quad
Reinhard Klein\\
University of Bonn \\
{\tt\small muellerj@cs.uni-bonn.de, landsgesell@uni-bonn.de, \{holland,stotko,rk\}@cs.uni-bonn.de}\\
}

\begin{document}
\maketitle
\maketitle

\begin{abstract}
The recent success of 3D Gaussian Splatting (3DGS) has reshaped novel view synthesis by enabling fast optimization and real-time rendering of high-quality radiance fields.
However, it relies on simplified, order-dependent alpha blending and coarse approximations of the density integral within the rasterizer, thereby limiting its ability to render complex, overlapping semi-transparent objects.
In this paper, we extend rasterization-based rendering of 3D Gaussian representations with a novel method for high-fidelity transmittance computation, entirely avoiding the need for ray tracing or per-pixel sample sorting.
Building on prior work in moment-based order-independent transparency, our key idea is to characterize the density distribution along each camera ray with a compact and continuous representation based on statistical moments.
To this end, we analytically derive and compute a set of per-pixel moments from all contributing 3D Gaussians.
From these moments, a continuous transmittance function is reconstructed for each ray, which is then independently sampled within each Gaussian.
As a result, our method bridges the gap between rasterization and physical accuracy by modeling light attenuation in complex translucent media, significantly improving overall reconstruction and rendering quality. 
\end{abstract}  
\section{Introduction}

Novel view synthesis has witnessed tremendous progress in recent years, driven initially by volumetric approaches such as Neural Radiance Fields (NeRF)~\cite{mildenhall2020nerf} and its many extensions\cite{barron2021mip, barron2022mip, barron2023zip}.
These implicit radiance-field formulations substantially improved visual fidelity through physically motivated volumetric integration.
More recently, the emergence of 3D Gaussian Splatting (3DGS)~\cite{kerbl20233d} has shifted the paradigm toward explicit representations.
By modeling scenes as collections of 3D Gaussians, 3DGS enables remarkably fast training and real-time rendering while continuing to achieve state-of-the-art image quality.
However, this efficiency comes at the cost of physical accuracy.
Core approximations limit the faithfulness and robustness of the representation, which includes replacing volumetric integration with splatting, assuming non-overlapping Gaussians with a correct front-to-back ordering, and modeling opacity independently of the spatial extent of Gaussians.

Recent works have begun addressing these shortcomings from two opposing directions.
Some methods abandon splatting entirely and instead adopt ray-tracing-based formulations that avoid the above approximations and provide physically accurate volumetric rendering of Gaussian primitives~\cite{moenne20243d, condor2025don}.
Others retain splatting to preserve its performance advantages, but target specific limitations.
StopThePop~\cite{radl2024stopthepop} mitigates popping artifacts caused by incorrect Gaussian ordering under view changes, whereas Vol3DGS~\cite{talegaonkar2025volumetrically} reintroduces proper volumetric integration of density, yet still assumes non-overlapping Gaussians and requires a correct rendering order.
Despite this progress, achieving the \emph{physical accuracy} of ray-traced approaches while maintaining the \emph{high efficiency} of rasterization-based splatting remains an open problem.

In this work, we present MB3DGS, a splatting-based method that performs accurate, order-independent rasterization of 3D Gaussians via moments.
In contrast to prior approaches, we treat opacity in a fully volumetric manner by modeling the combined density of potentially overlapping Gaussians to compute the exact emitted radiance.
Assuming only piecewise-constant density, analogous to volumetric methods such as NeRF, we derive an efficient numerical quadrature rule for radiance computation.
To achieve order-independent rendering, we reconstruct the moments of the transmittance function.
Since a naive formulation introduces numerical instability, we employ a power transform and derive a closed-form recurrence relation between moments. 
Combined with a confidence-interval-based formulation that produces correct screen-space bounds for more efficient rasterization, MB3DGS yields more stable and consistent results, particularly in visually complex regions where accurate volumetric modeling is essential.

In summary, our key contributions are:

\begin{itemize}
    \item A splatting-based, physically accurate rendering formulation that computes emitted radiance from the combined density of potentially overlapping 3D Gaussians.
    \item An efficient numerical quadrature rule derived under the assumption of piecewise-constant density, enabling volumetric radiance computation.
    \item A power-transform-based moment representation with a closed-form recurrence, resolving numerical instability in transmittance-moment computation, enabling order-independent rasterization without requiring Gaussian sorting.
    \item A confidence-interval-based screen-space bounding strategy that enables robust and faster rasterization and improves consistency in visually complex regions.
\end{itemize}
We release our code, data, and additional results at: \url{https://vc-bonn.github.io/mb3dgs/}

\section{Related Work}
\begin{figure*}
    \centering
    \includegraphics[width=\linewidth]{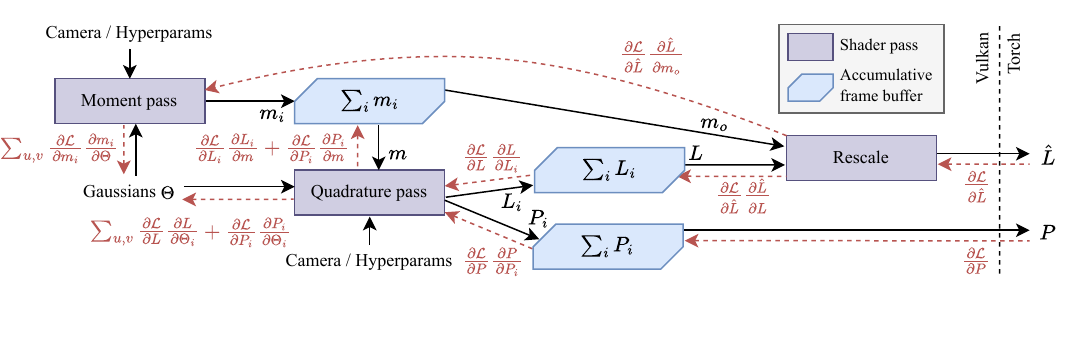}
    \vspace{-1.5cm}
    \caption{Overview of our order-independent differentiable rasterization pipeline. The forward pass features two separate accumulation passes: a Moment pass computes and sums per-Gaussian density moments to derive a continuous transmittance function. A Quadrature pass then independently evaluates the volume rendering integral for each Gaussian, computing and summing its radiance and penalty contributions. Also depicted is the adjoint gradient flow, which propagates derivatives from the final loss back through all stages. Notably, this includes backpropagation through both the radiance contributions and the transmittance to optimize the Gaussian parameters.}
    \label{fig:fwd-naive-bwd}
\end{figure*}
\paragraph*{Splatting-based Approaches.}
3D Gaussian Splatting (3DGS)~\cite{kerbl20233d} enables real-time, high-quality radiance field rendering and has sparked extensive follow-up work addressing its artifacts and physical limitations.
Several methods reduce view-dependent popping via improved or sort-free rasterization~\cite{radl2024stopthepop,housort,kheradmand2025stochasticsplats}, while others mitigate aliasing and scale distortions using Mip-filtering or analytic integration~\cite{yu2024mip,liang2024analytic}.
More physically grounded variants introduce volumetrically consistent integration~\cite{talegaonkar2025volumetrically,hahlbohm2025efficient}, explore alternative primitive parameterizations~\cite{huang20242d}, or extend splatting to complex cameras and secondary effects~\cite{wu20253dgut}.
Some works argue that optimization, rather than volumetric correctness, is the primary driver of fidelity~\cite{celarek2025does}.

Our method differs by explicitly modeling the combined density of overlapping Gaussians to compute physically accurate transmittance within a rasterization pipeline.

\paragraph*{Ray Tracing-based Approaches.}
To overcome the inherent limitations of splatting, another line of research adopts ray tracing for Gaussian primitives.
Moenne \etal~\cite{moenne20243d} and Mai \etal~\cite{mai2025ever} introduce efficient frameworks for volumetric or ellipsoidal ray tracing, eliminating popping and enabling physically based effects
Blanc \etal~\cite{blanc2025raygauss} and Condor \etal~\cite{condor2025don} propose Gaussian primitives tailored for volumetric ray-traced rendering, while other works address transparency via stochastic sampling~\cite{sun2025stochastic} or introduce Gaussian opacity fields for volumetric geometry extraction~\cite{yu2024gaussian}.
While these approaches offer high physical accuracy, they rely on the computationally heavier ray-tracing pipeline.

In contrast, our method achieves volumetric fidelity through moment-based transmittance reconstruction while retaining the efficiency of rasterization.

\paragraph*{Order-Independent Transparency.}
Order-independent transparency (OIT) techniques seek correct blending of semi-transparent geometry without sorting.
Classic exact methods such as the A-buffer~\cite{carpenter1984buffer} and depth peeling~\cite{everitt2001interactive} are accurate but costly, motivating real-time approximations including weighted averaging~\cite{bavoil2008order}, physically inspired blending~\cite{mcguire2013weighted}, and moment-based transparency~\cite{munstermann2018moment}, as well as recent learning-based methods~\cite{tsopouridis2024deep}.

Our work is conceptually related to moment-based OIT but adapts these ideas to volumetric Gaussian primitives, enabling continuous transmittance reconstruction along each ray within a splatting framework.
\section{Moment-based 3D Gaussian Splatting}
This section presents our method for approximating the volume rendering of 3D Gaussians (see \cref{ssec:quadrature}) while rasterizing each one individually.
Our approach first computes per-ray density moments to recover a continuous, order-independent transmittance function.
This function enables the independent evaluation of the volume rendering integral for each Gaussian via numerical quadrature (see \cref{ssec:transmittance}).
We then derive a geometric proxy for rasterization that accurately models perspective distortion in anisotropic Gaussians and detail the use of adjoint rendering for gradient computation (see \cref{ssec:rasterization}).
Finally, we describe the optimization and adaptive densification process for this volumetric representation (see \cref{ssec:optimisation}).
An overview of the process in described in \cref{fig:fwd-naive-bwd}.
\subsection{Volume Rendering} 
\label{ssec:quadrature}

In volume rendering, a participating medium is defined by a volume in space containing particles that interact with light through absorption, emission, and scattering.
The spatial distribution of these particles is described by a density function, representing extinction, which modulates the intensity of these light interactions.
Following most work in this area \cite{mildenhall2020nerf, barron2022mip, barron2023zip, kerbl20233d}, we only consider absorption and emission.
Thus, the final appearance of the object is determined by integrating these interactions along camera rays passing through the medium.
To solve this integration problem, we assume density is piecewise-constant within small intervals and recovery of the transmittance is possible from the statistical moments of the density, and we explicitly make no simplifying assumptions regarding self-occlusion or occlusion between Gaussians.

Each Gaussian particle defines both a localized density distribution and its appearance.
The density $\sigma$ at a position $\pos$ is the weighted sum of all Gaussian contributions:
\begin{equation}
    \sigma(\pos) = \sum_{i} w_i \, G(\pos \,|\, \mean_i, \cov_i)
\end{equation}
where $ { G(\pos) = e^{-\frac{1}{2}(\pos - \mean_i)^T \cov_i^{-1} (\pos - \mean_i)} } $.
In contrast to 3DGS~\cite{kerbl20233d}, which uses an opacity-centric model ($ { w_i \in [0, 1] } $), our physically-motivated approach only requires non-negative weights ($ { w_i \ge 0 } $), enforced via a softplus function.
The appearance is modeled by an emission term $(L_e)_i(\dir)$, which depends on direction $\dir$ and is represented using spherical harmonic (SH) coefficients $ { \bm{f}_i \in \mathbb{R}^{48} } $ up to degree $ { l = 3 } $.
Each Gaussian primitive is thus defined by a weight $ { w_i \in \mathbb{R}_{\ge 0} } $, a mean $ { \mean_i \in \mathbb{R}^3 } $, a covariance $\cov_i$, and the SH coefficients $\bm{f}_i$.
To ensure that the covariance matrix remains positive semi-definite during optimization, it is parameterized via $ { \cov = \rot \scale \scale^T \rot^T } $ with a rotation $ {\rot \in \text{SO}(3) }$ and a diagonal scaling matrix $ { \scale \in \mathbb{R}^{3 \times 3} } $.
For all remaining details regarding the parameterization, we refer the reader to the original publication \cite{kerbl20233d}.

The observed radiance $L$ along a camera ray $ { \bm{r}(t) = \origin + t \cdot \dir } $ from origin $\origin$ in direction $\dir$ within an emission-absorption medium is given by the volume rendering equation:
\begin{equation}
    L = \int_{t_n}^{t_f} T(t) \, \sigma(\pos_t) \, L_e(\pos_t, \dir) \, \dd t + T(t_f) \, L_{\textrm{bg}}(\pos_{t_f}, \dir)
    \label{eq:vre}
\end{equation}
where the transmittance $ { T(t) = e^{-\int_{t_n}^{t} \sigma(\pos_s) \, \dd s} } $ describes the mediums' permeability, $L_e(\pos, \dir)$ is the emitted radiance, $\pos_t$ is shorthand for $\bm{r}(t)$, $t_n$ and $t_f$ are the near and far integration bounds, and $L_{\text{bg}}$ is the incident radiance from the background.
To ensure that the total light emitted per unit distance at $\pos$ from all particles is the sum of their individual contributions, we define the emitted radiance as
\begin{equation}
    L_e(\pos, \dir) = \frac{1}{\sigma(\pos)} \sum_i \sigma_i(\pos) \, (L_e)_i(\dir).
    \label{eq:emitted_radiance}
\end{equation}

To solve the integral for each Gaussian, its 3D density contribution along the ray is first expressed as a 1D Gaussian in the ray's coordinate system:
\begin{equation}
    \sigma_i(t) = w_i \, G(\pos_t \,|\, \mean_i, \cov_i) = \wi \, e^{-\frac{(t-\mi)^2}{2\si^2}}
\end{equation}
To avoid explicit inversion of $\cov$, we follow \cite{moenne20243d}.
Let $ { \origin_g = \scale^{-1} \, \rot^T \, (\origin-\mean) } $ and $ { \dir_g = \scale^{-1} \, \rot^T \, \dir } $, then the parameters of the 1D Gaussian are given by:
\begin{equation}
    \si^2 = \frac{1}{\dir_g^T \, \dir_g},
    \quad
    \mi = -\frac{\origin_g^T \, \dir_g}{\dir_g^T \, \dir_g},
    \quad
    \wi = w_i \, e^K
\end{equation}
with
\begin{equation}
    K  = -\frac{1}{2} \origin_g^T \origin_g + \frac{1}{2} \frac{(\origin_g^T \dir_g)^2}{\dir_g^T \dir_g}
\end{equation}
Please refer to Appendix \ref{ap:raydensity} for the detailed derivation.
\subsection{Order-Independent Transmittance}
\label{ssec:transmittance}

To derive the quadrature, we first note that when plugging in \cref{eq:emitted_radiance} into \cref{eq:vre}, the density term $\sigma$ cancels out.
Swapping the order of integration and summation isolates the integral in terms of the $i$-th particle and transmittance.
This ray integral is then split into a sum of integrals over continuous, non-overlapping intervals $[t_j, t_{j+1}]$.
We assume a piecewise constant density, such that $ { \sigma(t) = \sigma(t_j) } $ for any $ { t \in [t_j, t_{j+1}] } $, which implies $\sigma_i(t)$ is also piecewise constant.
This leads to the quadrature for the $i$-th particle's contribution to the Volume Rendering Equation (VRE):
\begin{equation}
    L_i \approx \sum_{j=1}^N \left( T(t_j) - T(t_{j+1}) \right) \frac{\sigma_i(t_j)}{\sigma(t_j)} (L_e)_i(\mathbf{d}).
    \label{eq:quadrature}
\end{equation}
with ${\sigma(t_j) = -(t_{j+1} - t_j)^{-1} \, \log(T(t_{j+1}) / T(t_j))}$.
Please refer to Appendix \ref{ap:quadrature} for the detailed derivation.

To solve the quadrature for each particle individually, we adapt the work on order-independent transparency by Münstermann \etal~\cite{munstermann2018moment} to our problem statement.
Their proposed approach builds on the concepts of statistical moments and the moment problem.

Let $ { \tau \colon \mathbb{R} \to \mathbb{R} } $ be a monotonic increasing, right-continuous function, which defines a unique Lebesgue-Stieltjes measure $\mu_\tau$. The $k$-th raw moment $m_k$ of this measure is defined as $ { m_k = \int_{-\infty}^{\infty} x^k \, \dd \mu_\tau(x) } $. When $\tau$ is absolutely continuous, this measure $\mu_\tau$ has a density $ { \sigma(x) = \tau'(x) } $ with respect to the Lebesgue measure, allowing the moment to be computed as 
$ { m_k = \int_{-\infty}^{\infty} x^k \sigma(x) \, \dd x } $. In general, a finite number of moments does not characterize a unique measure. However, a lower and upper bounds on the set of measures characterized by the finite moments can be computed~\cite{tari2005unified}.

\begin{figure}
\centering
\begin{subfigure}{.5\linewidth}
    \centering
    \includegraphics[width=\linewidth]{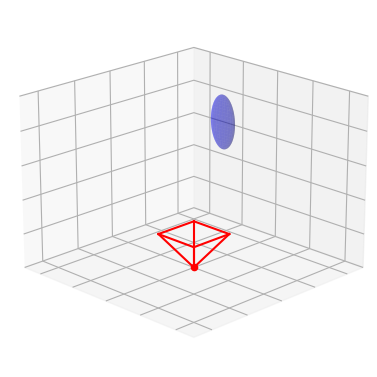}
    \caption{3D scene configuration.}
\end{subfigure}%
\begin{subfigure}{.5\linewidth}
    \centering
    \includegraphics[width=\linewidth]{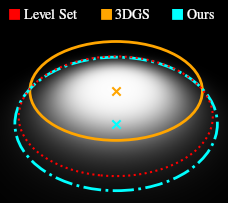}
    \caption{2D screen-space proxies.}
\end{subfigure}
\caption{Comparison of 2D splat proxy accuracy. The affine approximation used by 3DGS provides a poor bound, resulting in a proxy that provides insufficient coverage and is offset from the true perspectively-correct level set. Our method computes a  tighter geometric proxy that conservatively bounds the true projection, ensuring all contributions are correctly rasterized.}
\label{fig:splatting_comparison}
\end{figure}
Münstermann \etal~\cite{munstermann2018moment} model occlusion along a view ray via an absorbance function $ { A(z)=-\ln T(z) } $, with $T(z)$ being the transmittance at depth $z$.
For discrete transparent surfaces at depths $z_l$ with opacities $\alpha_l$, the absorbance function $ { A(z) = \sum_{\substack{l=0, z_l < z}} -\ln(1-\alpha_l) } $ is a monotonic, right-continuous step function and an instance of $\tau$.
The associated Lebesgue-Stieltjes measure $\mu_A$ is a sum of weighted Dirac delta functions, where each surface corresponds to a point mass: $ { \mu_A = \sum_{l=0} w_l \, \delta_{z_l} } $ with weights $ { w_l = -\ln(1-\alpha_l) } $.
Rather than storing the full measure, its first $ { 2n+1 } $ moments with $ { n = 4 } $, $ { m_k=\int z^k \, \dd \mu_A(z) } $, are computed and stored per pixel. 
To reconstruct the bounds on the absorbance at a query depth $\eta$, a unique canonical representation of the moments is constructed.
This representation is itself a discrete measure, $ { \tau_\eta = \sum_{i=0}^n \, w_i' \, \delta_{x_i} } $, with $ { n+1 } $ points of support that has the same moments as $\mu_A$ and is constrained such that one of its support points is the query depth itself, i.e., $ { x_0=\eta } $ \cite{worchel2025moment}.

The problem is thus reduced to finding the unknown locations $\{x_i\}_{i=1}^n$ and weights $\{w_i'\}_{i=0}^n$.
The locations are found as the roots of the degree-$ n $ kernel polynomial $ { K(x)=x^T \,  H^{-1} \, \eta } $, where $H$ is the Hankel matrix of the moments ($ { H_{ij}=m_{i+j} } $).
Once the locations $\{x_i\}$ are known, the weights are found by solving the linear Vandermonde system given by the moment equations $ { m_k=\sum_{i=0}^n w_i' \, x_i^k }$.
Finally, the bounds are computed from this discrete measure:
\begin{equation}
    L(\eta) = \sum_{x_i < \eta} w_i' \quad \text{and} \quad U(\eta) = \sum_{x_i \le \eta} w_i'.
\end{equation}
The transmittance then is $ { T(\tau) = (1-\beta) L + \beta U } $ with $ { \beta=0.25 } $. Recent work has proven that these moment-bounds are differentiable \cite{worchel2025moment}.

In our volumetric setting, the optical depth along the ray, $ { \tau(t) = - \log(T(t)) = \int_{t_n}^t \sigma_t(s) \, \dd s } $, serves as the continuous and differentiable analog to the discrete absorbance function $A(z)$ used by Münstermann \etal.
As $\tau(t)$ is absolutely continuous, the Radon-Nikodym theorem guarantees its associated Lebesgue-Stieltjes measure, $\mu_\tau$, has a density with respect to the Lebesgue measure.
This density is precisely the sum of 1D Gaussians, $\sigma(t)$.
The moments of this measure are therefore computed by integrating against this continuous density:
\begin{equation}
    m_k = \int_{t_n}^{t_f} t^k \, \sigma(t) \, \dd t.
\end{equation}

However, direct computation of $m_k$ is numerically unstable on intervals $[t_n, t_f]$ where $ { t_f \gg 1 } $, as individual particle moments grow rapidly, since $ { (m_k)_i \geq \wi \mi^k \si } $ . To stabilize this, we warp the domain $[t_n, t_f]$ to $[0,1]$ using the transformation $ { \hat{g}(t) = (f(t) - f(t_n))/(f(t_f) - f(t_n)) } $. Münstermann \etal~\cite{munstermann2018moment} proposed this transformation with $ { f(t) = \log(t) } $ to address a similar problem. The choice of the non-linear function $f(t)$ is critical for minimizing linearization error \cite{munstermann2018moment, neff2021donerf, barron2022mip}. As parameterized power transformations are particularly effective when combined with local linear approximations \cite{barron2023zip}, we follow this approach and set $f(t)$ to be the Power-Transform $f_\lambda(2t)$ with $ { \lambda = -1.5 } $. This provides a robust mapping, behaving linearly for near distances while transitioning to an inverse-like function for far distances. For a detailed analysis, see \cite{barron2025power}.

We therefore compute the moments by integration powers of the warped distance against the density: $ { \hat{m}_k = \sum_{i} \int_{t_n}^{t_f} \hat{g}(t)^k \sigma_i(t) \dd t } $. To solve the inner integral $(\hat{m}_k)_i$, we linearize $\hat{g}(t)$ using a Taylor expansion at each particle's mean $\mi$. Assuming an unbounded medium ($ { t_f \to \infty } $), this yields a recurrence for $ { k \geq 2 } $ with closed-form base cases:
\begin{align}
    (\hat{m}_0)_i &= \wi \, \si \, \sqrt{\frac{\pi}{2}} \, \left(1- \erf\left(b_n\right)\right) \label{eq:optical_depth} \\
    (\hat{m}_1)_i &= \hat{g}(\mi) \, (\hat{m}_0)_i - \hat{g}'(\mi)\, \wi \, \si^2 \, e^{-\frac{(t_n - \mi)^2}{2\si^2}} \\
    (\hat{m}_k)_i &= \hat{g}(\mi) \, (\hat{m}_{k-1})_i + \beta\,(k-1) \, (\hat{m}_{k-2})_i - B_i(k)
\end{align}
where $ { \beta = \hat{g}'(\mi)^2 \, \si^2 } $ is scaled variance and near boundary term $ { B_i(k) = -\hat{g}'(\mi) \, \wi \, \si^2 \, u_n^{k-1} \, e^{-b_n^2} } $ with $ { b_n = (t_n-\mi) /(\sqrt{2}\si) } $ and linearized distance $ { u_n = \hat{g}(\mi) + \hat{g}'(\mi) \, (t_n - \mi) } $.

An alternative to the polynomial basis, explored in order-independent occluder literature~\cite{peters2015moment, peters2016beyond, peters2017improved, peters2017non}, are trigonometric moments $\tilde{m}_k$ with a Fourier basis. We adapt this concept to compute trigonometric density moments,
\begin{equation}
    \tilde{m}_k = \sum_{j} \int_{t_n}^{t_f} \left(e^{(2\pi - \theta)i \hat{g}(t)}\right)^k \, \sigma_j(t) \, \dd t
\end{equation}
where $i$ is the imaginary unit. Linearizing $\hat{g}(t)$ at $\mj$ via a Taylor expansion, and again assuming $t_f \to \infty$, provides a closed-form approximation for the inner integral $(\tilde{m}_k)_j$:
\begin{equation}
    (\tilde{m}_k)_j \approx \wj \, \sqrt{\frac{\pi}{2}} \, \sj \, e^{i \alpha \hat{g}(\mj)- \frac{\sj^2 \beta^2}{2}} \, (1 - \erf(v_n)) 
\end{equation}
with $ { v_n = (t_n - \mj) / (\sqrt{2}\sj) - i(\sj \beta)/\sqrt{2} } $, phase $ { \alpha = k(2\pi - \theta) } $, and $ { \beta = \alpha \hat{g}'(\mj) } $. This requires evaluating $\erf(v_n)$ with a complex argument; we approximate this using a first-order Taylor expansion in the imaginary direction. Notably, $\hat{m}_0$ and $\tilde{m}_0$ are exactly equal to the total optical depth $\tau(t_f)$ and involve no approximation. 
Please refer to Appendix \ref{ap:powermoments} for detailed derivations.

Given an order-independent estimate for the optical depth, \cref{eq:quadrature} can be estimated for each particle individually.
To correct for visual opacity fluctuations arising from moment-based transmittance estimation, we renormalize the final radiance. Following Münstermann \etal, we scale the accumulated radiance $\sum L_i$ by the ratio of the true scene opacity, $1 - e^{-m_0}$ (derived from the zeroth moment $m_0$), to the estimated opacity $O$.
This $O$ is accumulated in the alpha-channel alongside the individual radiance contributions $L_i$ by accumulating $ { (L_e)_i = (r,g,b,1) } $
The final, stabilized radiance $L$ is:
\begin{equation}    
    L = \frac{1-e^{-m_0}}{\max\left(\epsilon, O\right)} \sum_{i=1}^n L_i + e^{-m_0} \, L_\textrm{bg}
\end{equation}
where the denominator is clamped by $ { \epsilon > 0 } $ for stability.
\begin{figure}
    \newcommand{\subwidth}{0.49\linewidth}
    \centering
    \begin{subfigure}[t]{\subwidth}
        \includegraphics[width=\textwidth, keepaspectratio, trim={220 150 200 170}, clip]{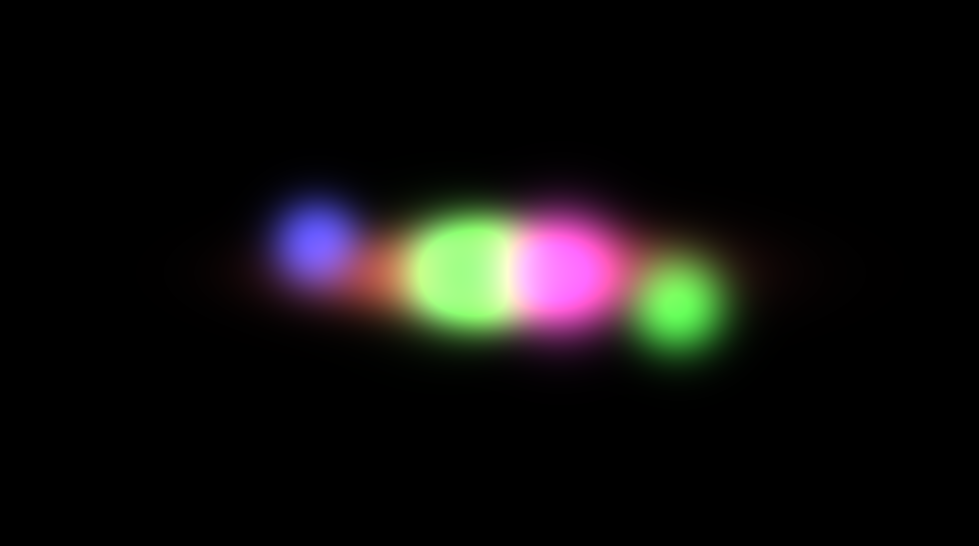}
        \caption{Ground Truth}
    \end{subfigure}
    \begin{subfigure}[t]{\subwidth}
        \includegraphics[width=\textwidth, keepaspectratio, trim={220 150 200 170}, clip]{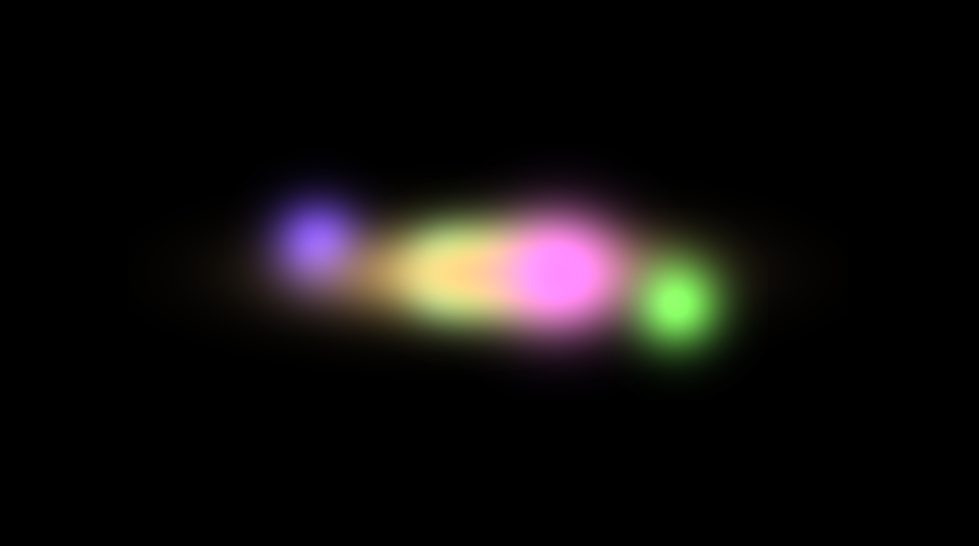}
        \caption{Vol3DGS \cite{talegaonkar2025volumetrically}}
    \end{subfigure}
    \begin{subfigure}[t]{\subwidth}
        \includegraphics[width=\textwidth, keepaspectratio, trim={220 150 200 170}, clip]{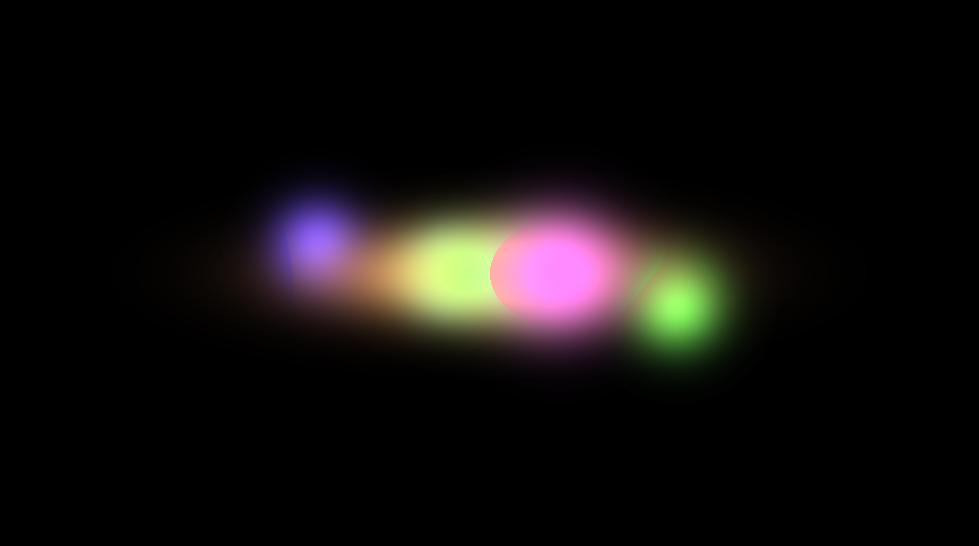}
        \caption{StopThePop \cite{radl2024stopthepop}}
    \end{subfigure}
    \begin{subfigure}[t]{\subwidth}
        \includegraphics[width=\textwidth, keepaspectratio, trim={220 150 200 170}, clip]{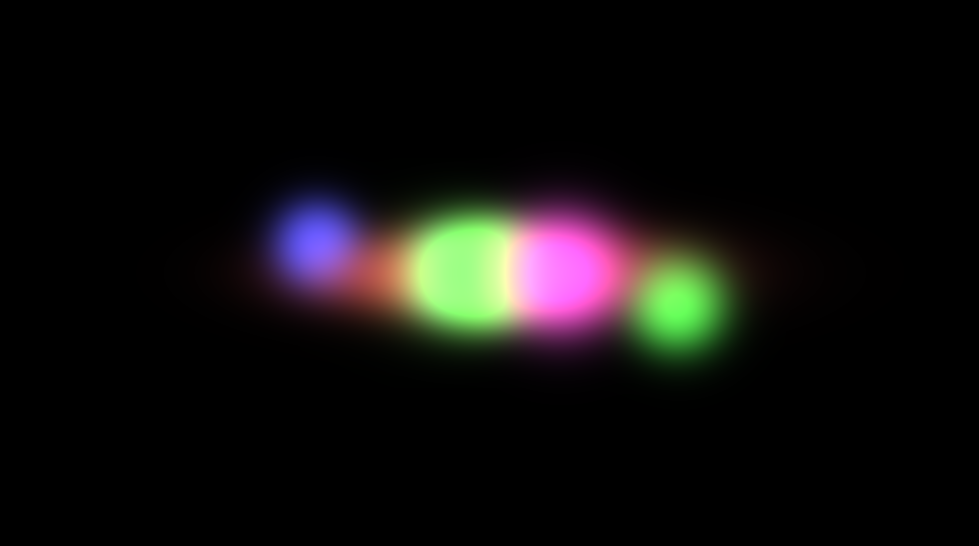}
        \caption{Ours}
    \end{subfigure}
    \caption{Qualitative comparison on a synthetic scene designed to evaluate complex color blending.}
    \label{fig:volumetric_qualitative}
\end{figure}
\subsection{Rasterisation and Adjoint Rendering}
\label{ssec:rasterization}

The rendering process maps efficiently to the GPU rasterization pipeline because both the per-particle moment and radiance quadrature contributions can be evaluated individually and in an arbitrary order.
Our forward rendering approach, illustrated in \cref{fig:fwd-naive-bwd}, proceeds in four main stages.
First, a culling pass visibility tests each 3D Gaussian against the camera frustum using its bounding sphere, defined by $ { r = \max\mathrm{diag}(\bm{S}) } $.
Next, two separate accumulation passes rasterize all visible Gaussians: the moment generation pass evaluates and sums per-Gaussian moments, and the radiance quadrature pass accumulates the per-Gaussian radiance quadrature, with both passes utilizing additive framebuffers.
Finally, a per-pixel normalization pass rescales the observed radiance to ensure correct scene opacity.

\paragraph{Confidence Interval-based Rasterization}
Rasterization of Gaussians requires a screen-space proxy, like a quad, whose shape is derived from projecting the 3D covariance matrix.
In EWA-based splatting~\cite{zwicker2001ewa} and 3DGS~\cite{kerbl20233d}, this projection uses a locally-affine approximation of the perspective transform to map the 3D covariance to its 2D screen-space counterpart.
This shared method cleanly decouples geometry from appearance: the covariance matrix exclusively determines the splat's screen-space footprint, independent of learned scalar parameters (like peak density or opacity) that modulate its final intensity.
This screen-space proxy, however, is unsuitable for our volumetric rendering approach.
The EWA proxy fails to cover all screen areas where the particle has meaningful radiance contributions (see \cref{fig:splatting_comparison}).
This under-coverage becomes especially pronounced as particles approach the camera or if their covariance is ill-conditioned.
We therefore derive a new geometric proxy for a perspective camera that covers all radiance contributions within a confidence threshold.

The required geometric proxy area is dictated by the particle's isolated opacity.
Along a ray, this opacity integral simplifies for a single Gaussian to a closed-form solution:
\begin{equation}
    \int_{t_n}^{\infty} T_i(t_n \to t) \, \sigma_i(t) \, \dd t = 1-e^{-\bar{\tau}_i}
\end{equation}
with optical depth $\bar{\tau}_i$ evaluated using \cref{eq:optical_depth}. 
The 1D Gaussian parameters ($\mi$, $\si$, $\wi$) describe the particle's density distribution along the ray and are functions of that ray's origin and direction.
By extension, the optical depth $\bar{\tau}_i$ and the particle's opacity are also functions of the ray direction.
Under a perspective camera model, a pixel $ { \bm{p}_\mathrm{hom} = (u,v,1)^T } $ maps to a ray direction $ { \dir_{\bm{p}} = \textrm{normalize}(\bm{K}^{-1}\bm{p}_\mathrm{hom}) }$ via the intrinsic matrix $\bm{K}$.
Opacity, being a function of ray direction, can therefore be interpreted as a function of pixel position.

The new geometric proxy encloses the implicit screen-space curve defined by the confidence interval $ { c = 1-e^{-\bar{\tau}_i} } $.
Assuming the Gaussian is at a reasonable distance from the camera (\ie, $ { \mi > t_n - 4 \sqrt{2} \si } $), the full level set equation remains complex due to the interdependence of $\wi$ and $\si$ on the ray direction $\dir$.
To obtain a tractable solution, we replace $\si$ with an upper bound, which produces a valid, larger geometric proxy.
This simplification isolates the level set of $\wi$, allowing it to be expressed in a quadratic form $ { \bm{p}_{hom}^T \, \bm{W} \, \bm{p}_{hom} = 0 } $. The $ { 3 \times 3 } $ symmetric matrix $\bm{W}$ is:
\begin{equation}
    \bm{W} = (\bm{m}\bm{m}^T - \kappa \bm{M})
\end{equation}
with $ { \bm{m} = (\bm{K}^{-1})^T \, \invcov \, \mean } $ , $ { \bm{M} = (\bm{K}^{-1})^T \, \invcov \, \bm{K}^{-1} }$ and 
\begin{equation}
    \kappa = 2 \left(\log\left(C \right) - \log(w)\right) + \mean^T \invcov \mean .
\end{equation}
with ${C = \left(-\log(1-c) \, \sqrt{\bm{u}^T \invcov \bm{u}}\right)/\left(\sqrt{2\pi \,  } \lVert \bm{u} \rVert_2\right)}$ with ${ \bm{u} = \textrm{normalize}(\mean - \origin) } $. This quadratic form yields an ellipse.
We partition $\bm{W}$ into a $2 \times 2$ block $\bm{W}_{2\times2}$, a vector $\bm{w}_{2\times1}$, and a scalar $w_{33}$ to derive the standard statistical representation for a point $\bm{p}=(u,v)^T$ on the ellipse:
\begin{equation}
    (\bm{p} - \mean_{2d})^T \, \cov_{2d}^{-1} \, (\bm{p} - \mean_{2d}) = 1
\end{equation}
where the 2D mean $\mean_{2d}$ is:
\begin{equation}
    \mean_{2d} = -\bm{W}_{2\times2}^{-1} \, \bm{w}_{2\times1}
\end{equation}
and the 2D quadrature matrix $\cov_\textrm{2d}$ is:
\begin{equation}
    \cov_{2d} = (\mean_{2d}^T \, \bm{W}_{2\times2} \, \mean_{2d} - w_{33}) \, \bm{W}_{2\times2}^{-1}
\end{equation}
This geometric proxy is then rasterized as an aligned rectangle  following \cite{zwicker2001ewa} however without additional scaling of the semi-major and semi-minor axis of the ellipse since the eigenvalues of $\cov_{2d}$ already capture all scaling effects.
Please refer to Appendix \ref{ap:rasterisation} for the detailed derivation.

\paragraph{Adjoint Rendering}
Since hardware-accelerated rasterization is not fully differentiable, we require a custom adjoint rendering method.
Our forward pass consists of three stages: a Moment pass to compute a per-pixel moment texture $\bm{m}$ from Gaussian parameters $\Theta$, a Quadrature pass to compute radiance $L$ and penalty $P$, and a rescaling pass that uses the first moment $\bm{m}_0$ to produce the final radiance $\hat{L}$.
A naive backward pass inverting these stages is inefficient.
It requires two separate reduction operations and numerous intermediate adjoint framebuffers, including one for the opacity rescaling derivative $\partial \mathcal{L} /\partial \hat{L} \cdot \partial \hat{L} / \partial \bm{m}_0$.

We introduce an optimized backward pass that resolves these inefficiencies.
We first observe that the derivative from the Rescaling pass, $\partial \hat{L} / \partial \bm{m}_0$, can be re-evaluated and folded into the other backward stages, eliminating the need for three additive framebuffers.
We then consolidate all gradient computations into a single, efficient reduction.
This optimized pass, begins with an Adjoint Moment stage that computes a per-pixel adjoint moment texture $\delta \bm{m}$.
A subsequent Gradient stage re-rasterizes all Gaussians, using $\delta \bm{m}$ and the upstream gradients $\partial \mathcal{L} / \partial L$ and $\partial \mathcal{L} / \partial P$ to perform a single reduction over all covered pixels $(u,v)$, yielding the final gradient $\nabla_{\Theta_i}$:
\begin{equation}
    \nabla_{\Theta_i} = \sum_{u,v} \frac{\partial \mathcal{L}}{\partial L_i} \frac{\partial L_i}{\partial \Theta_i} + \frac{\partial \mathcal{L}}{\partial P_i} \frac{\partial P_i}{\partial \Theta_i} + \frac{\partial \mathcal{L}}{\partial \bm{m}_i}\frac{\partial \bm{m}_i}{\Theta_i}
\end{equation}
Please refer to Appendix \ref{ap:rasterisation} for a more detailed discussion.
\begin{figure}
    \newcommand{\subwidth}{0.49\linewidth}
    \centering
    \begin{subfigure}[t]{\subwidth}
       \includegraphics[width=\textwidth, keepaspectratio, trim={220 150 200 170}, clip]{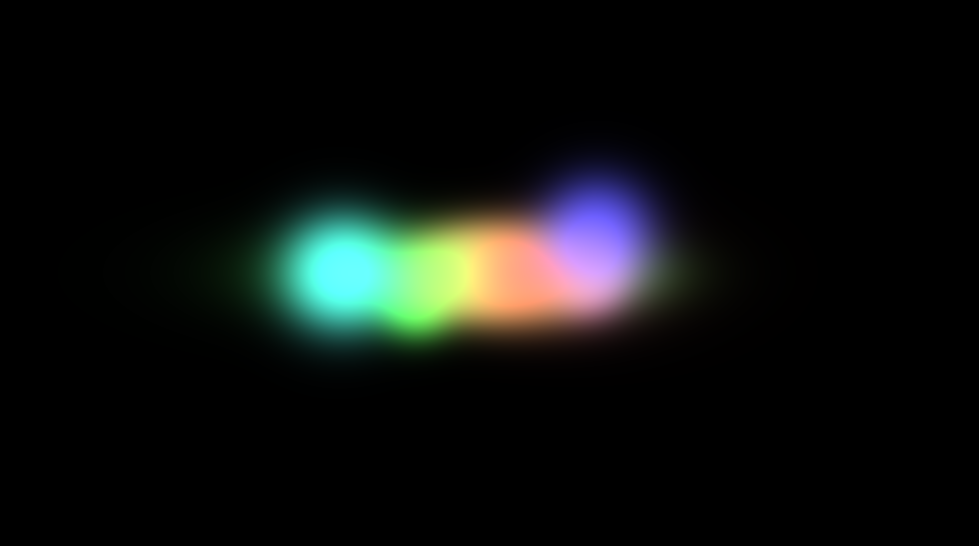}
        \caption{Ground Truth}
    \end{subfigure}
    \begin{subfigure}[t]{\subwidth}
        \includegraphics[width=\textwidth, keepaspectratio, trim={220 150 200 170}, clip]{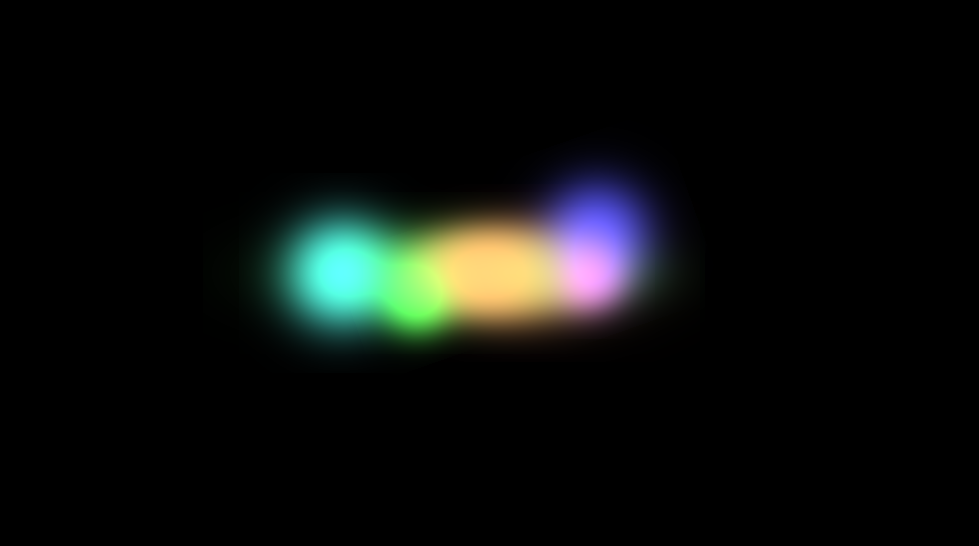}
        \caption{Power Moments $ { N=3 } $}
    \end{subfigure}
    \begin{subfigure}[t]{\subwidth}
        \includegraphics[width=\textwidth, keepaspectratio, trim={220 150 200 170}, clip]{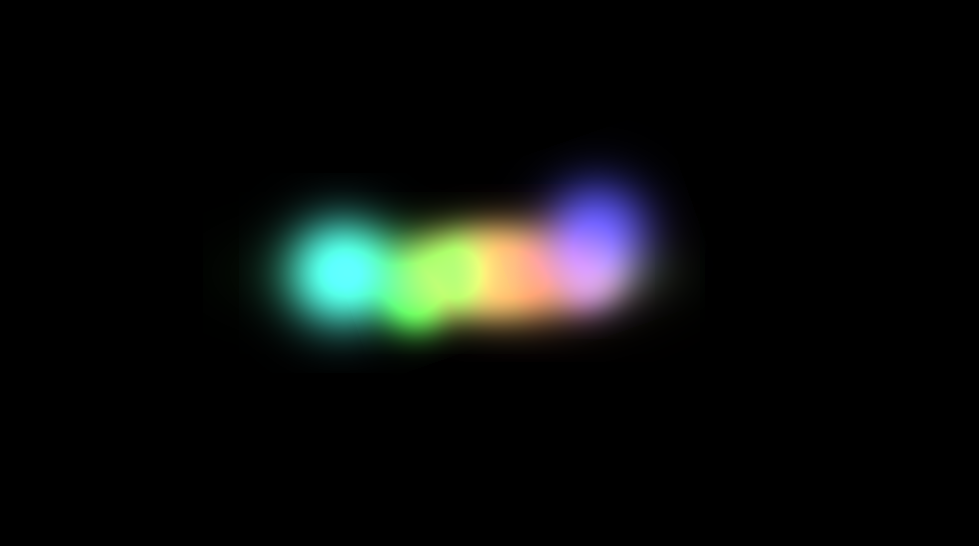}
        \caption{Trig. Moments $ { N=3 } $ }
    \end{subfigure}
    \begin{subfigure}[t]{\subwidth}
       \includegraphics[width=\textwidth, keepaspectratio, trim={220 150 200 170}, clip]{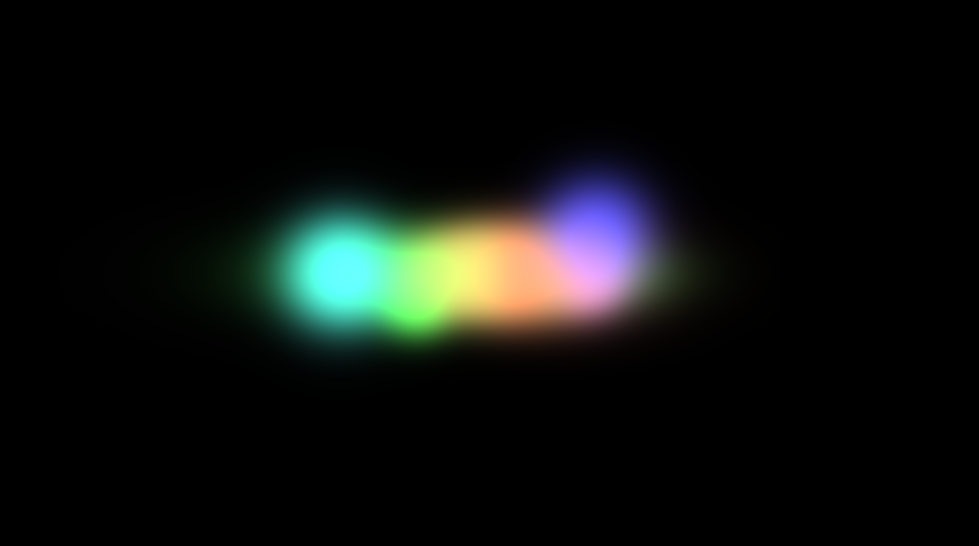}
        \caption{Trig. Moments $ { N=5 } $}
    \end{subfigure}
    \caption{Visual comparison of Power Moments and Trigonometric Moments (using ${ N=3 } $ and $ { N=5 } $ intervals) against the ground truth on synthetic data.}
    \label{fig:ablation_qualitative}
\end{figure}
\begin{figure*}[t]
    \newcommand{\percellwidth}{0.245\linewidth}
    \centering
    \captionsetup[subfigure]{justification=centering}
    
    \def\zoomx{-1.30}
    \def\zoomy{0.60}
    \def\zoomboxx{1.45}
    \def\zoomboxy{-0.5}
    \def\zoomlevel{1.6}

    \begin{subfigure}[t]{\percellwidth}
        \imagewithzoom{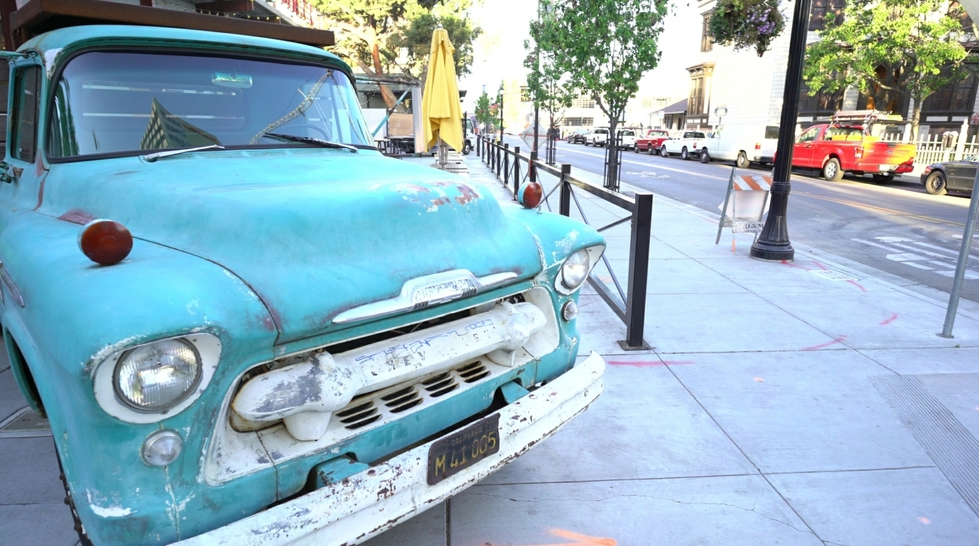}{\zoomx}{\zoomy}{\zoomboxx}{\zoomboxy}{1.2cm}{\zoomlevel}{0 0 0 0}
    \end{subfigure}
    \hfill
    \begin{subfigure}[t]{\percellwidth}
        \imagewithzoom{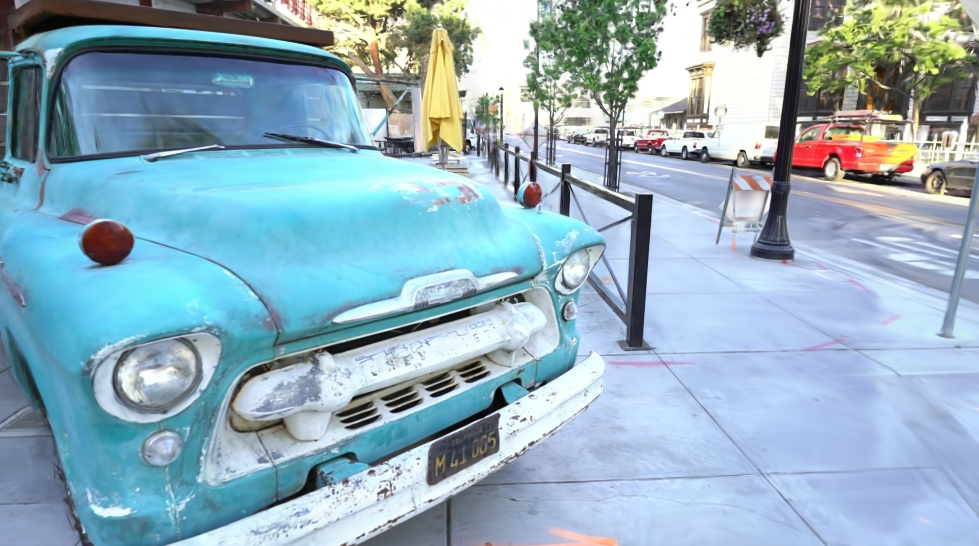}{\zoomx}{\zoomy}{\zoomboxx}{\zoomboxy}{1.2cm}{\zoomlevel}{0 0 0 0}
    \end{subfigure}
    \hfill
    \begin{subfigure}[t]{\percellwidth}
        \imagewithzoom{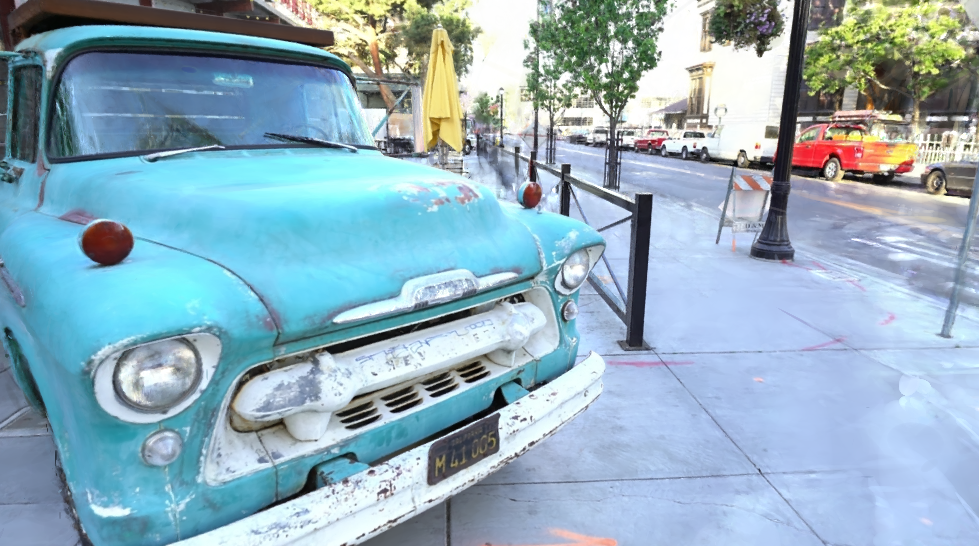}{\zoomx}{\zoomy}{\zoomboxx}{\zoomboxy}{1.2cm}{\zoomlevel}{0 0 0 0}
    \end{subfigure}
    \hfill
    \begin{subfigure}[t]{\percellwidth}
        \imagewithzoom{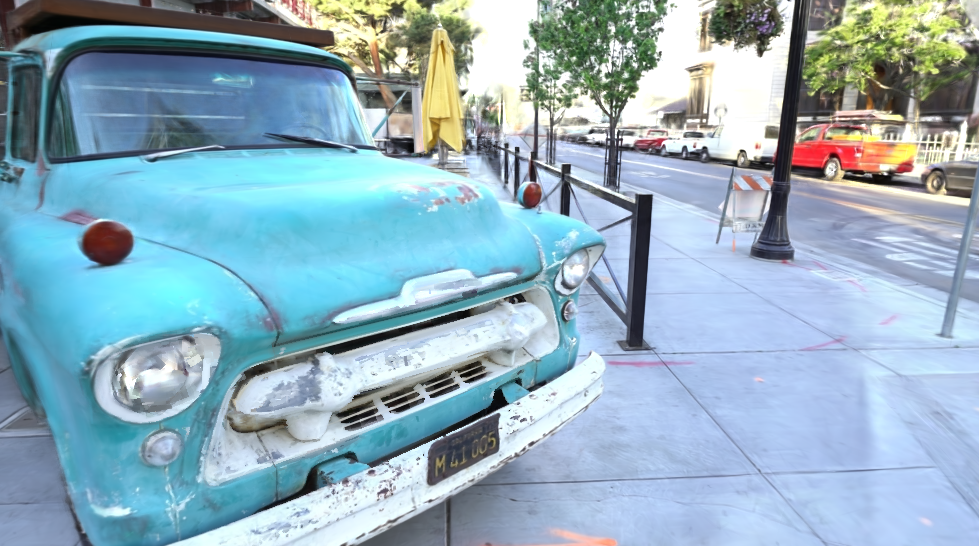}{\zoomx}{\zoomy}{\zoomboxx}{\zoomboxy}{1.2cm}{\zoomlevel}{0 0 0 0}
    \end{subfigure}


    \def\zoomx{0.90}
    \def\zoomy{0.9}
    \def\zoomboxx{-1.45}
    \def\zoomboxy{-0.73}
    \def\zoomlevel{2}

    \begin{subfigure}[t]{\percellwidth}
        \imagewithzoom{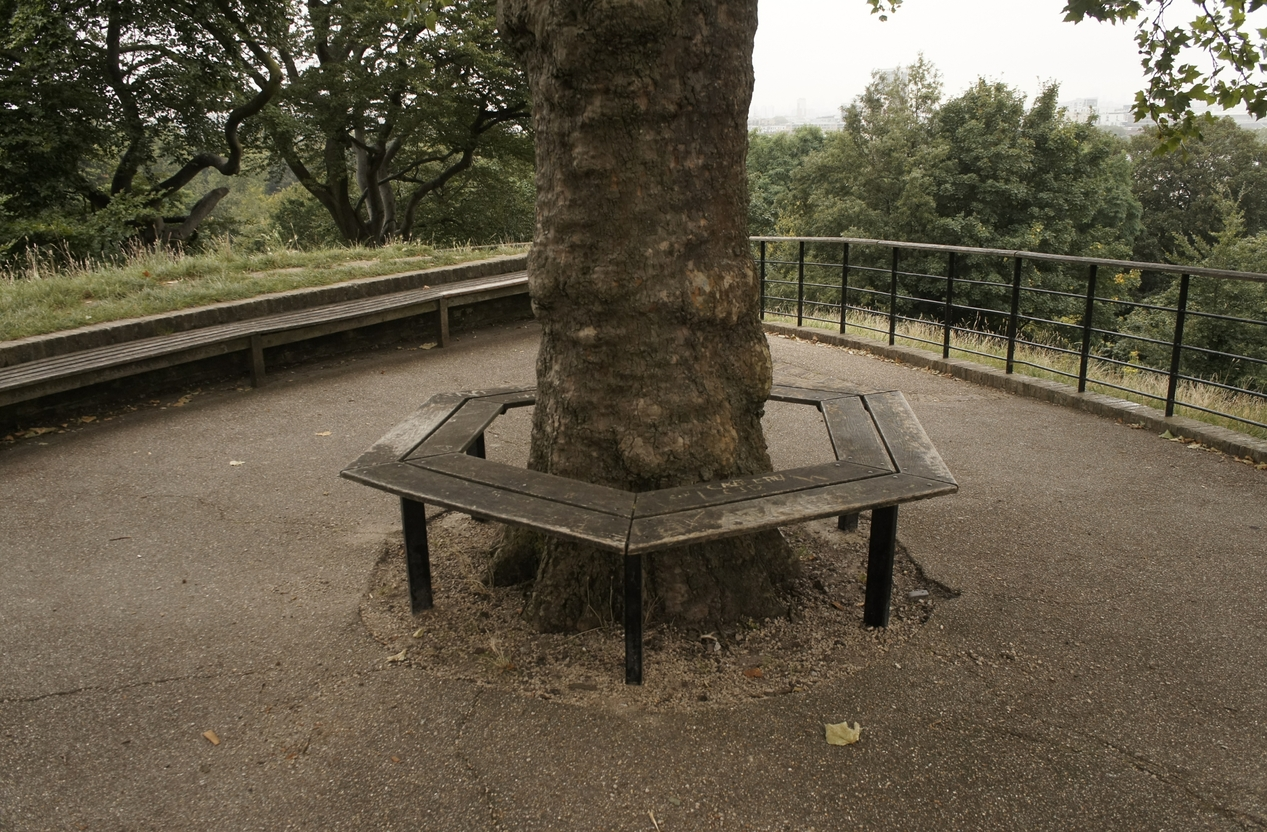}{\zoomx}{\zoomy}{\zoomboxx}{\zoomboxy}{1.2cm}{\zoomlevel}{0 0 0 0}
    \end{subfigure}
    \hfill
    \begin{subfigure}[t]{\percellwidth}
        \imagewithzoom{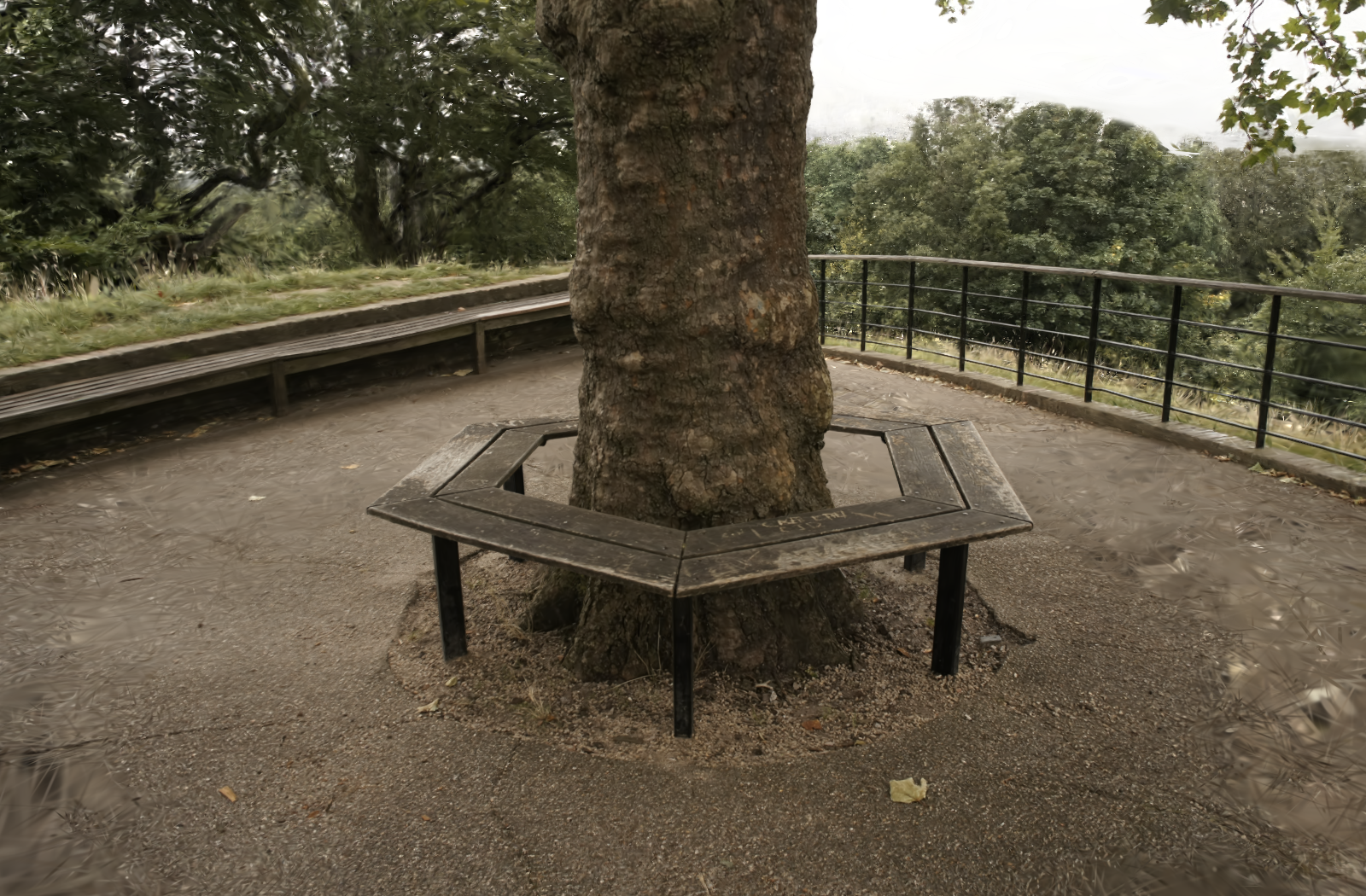}{\zoomx}{\zoomy}{\zoomboxx}{\zoomboxy}{1.2cm}{\zoomlevel}{0 0 0 0}
    \end{subfigure}
    \hfill
    \begin{subfigure}[t]{\percellwidth}
        \imagewithzoom{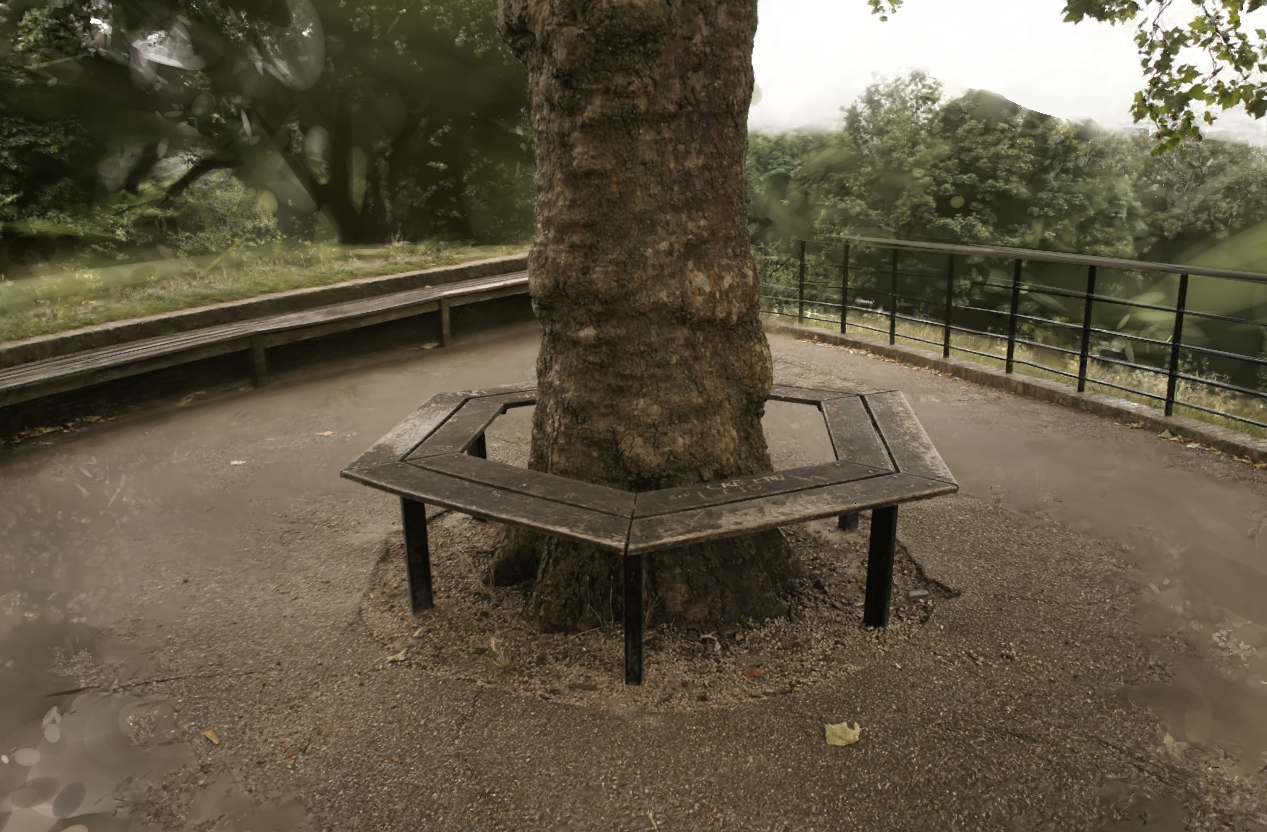}{\zoomx}{\zoomy}{\zoomboxx}{\zoomboxy}{1.2cm}{\zoomlevel}{0 0 0 0}
    \end{subfigure}
    \hfill
    \begin{subfigure}[t]{\percellwidth}
        \imagewithzoom{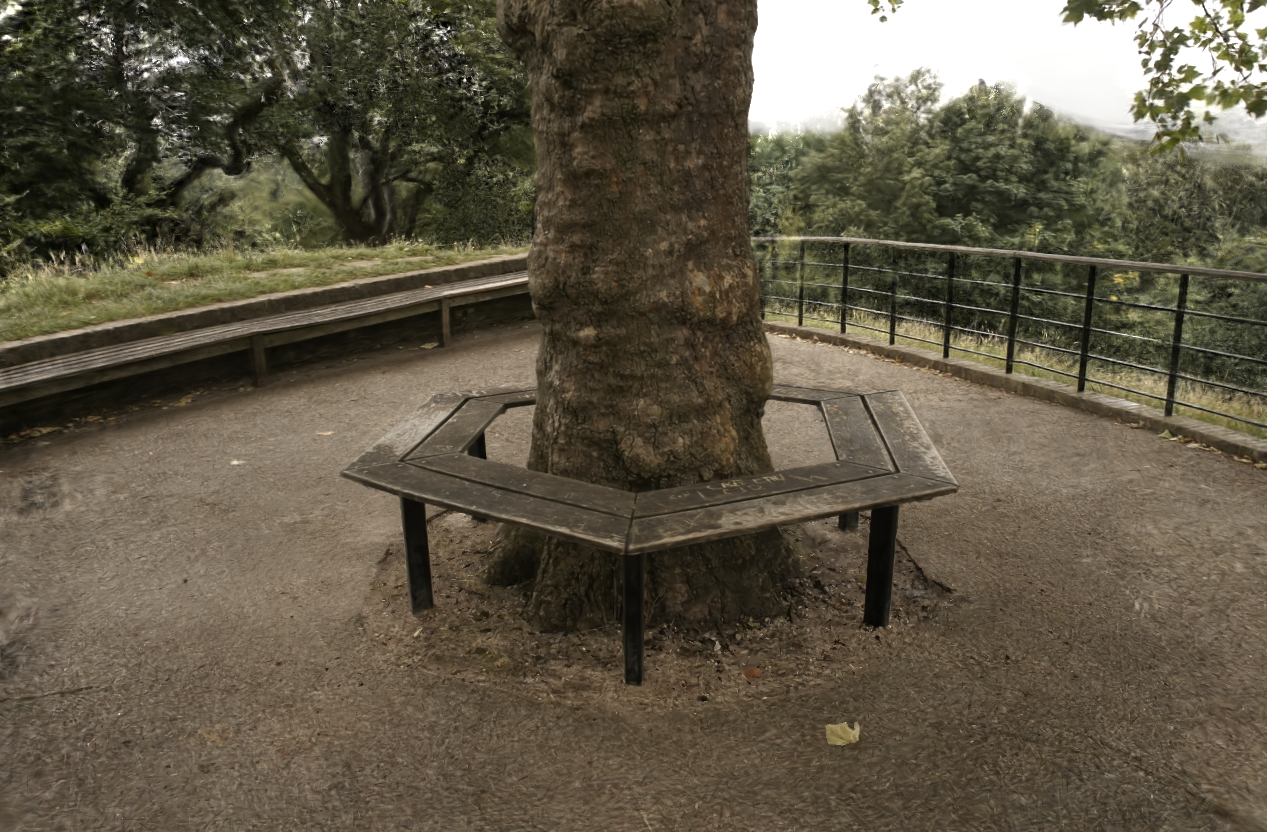}{\zoomx}{\zoomy}{\zoomboxx}{\zoomboxy}{1.2cm}{\zoomlevel}{0 0 0 0}
    \end{subfigure}


    \def\zoomx{-0.9}
    \def\zoomy{-0.4}
    \def\zoomboxx{-1.45}
    \def\zoomboxy{0.77}
    \def\zoomlevel{2.0}

    \begin{subfigure}[t]{\percellwidth}
        \imagewithzoom{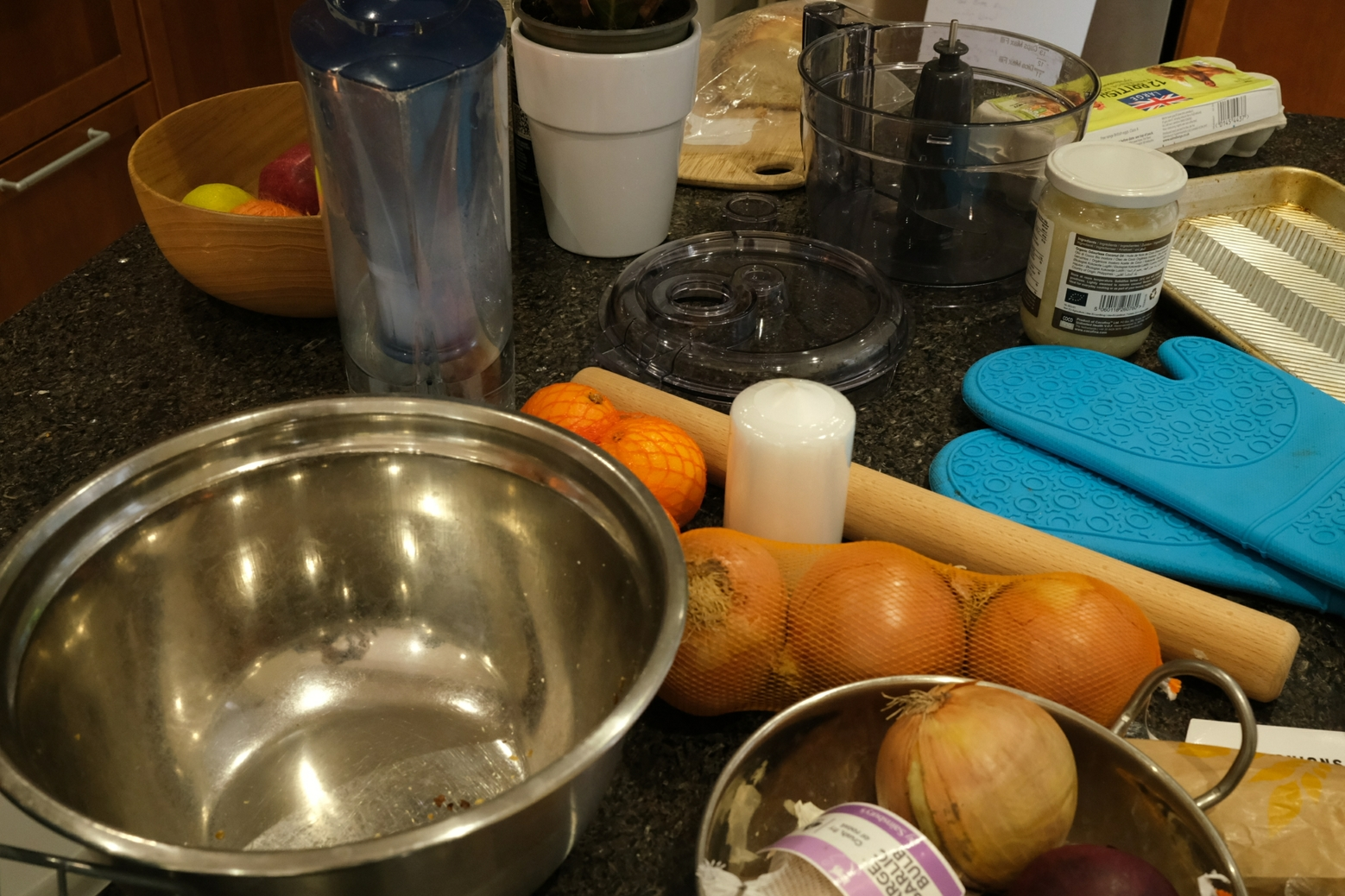}{\zoomx}{\zoomy}{\zoomboxx}{\zoomboxy}{1.2cm}{\zoomlevel}{0 0 0 0}
        \caption{Ground Truth}
    \end{subfigure}
    \hfill
    \begin{subfigure}[t]{\percellwidth}
        \imagewithzoom{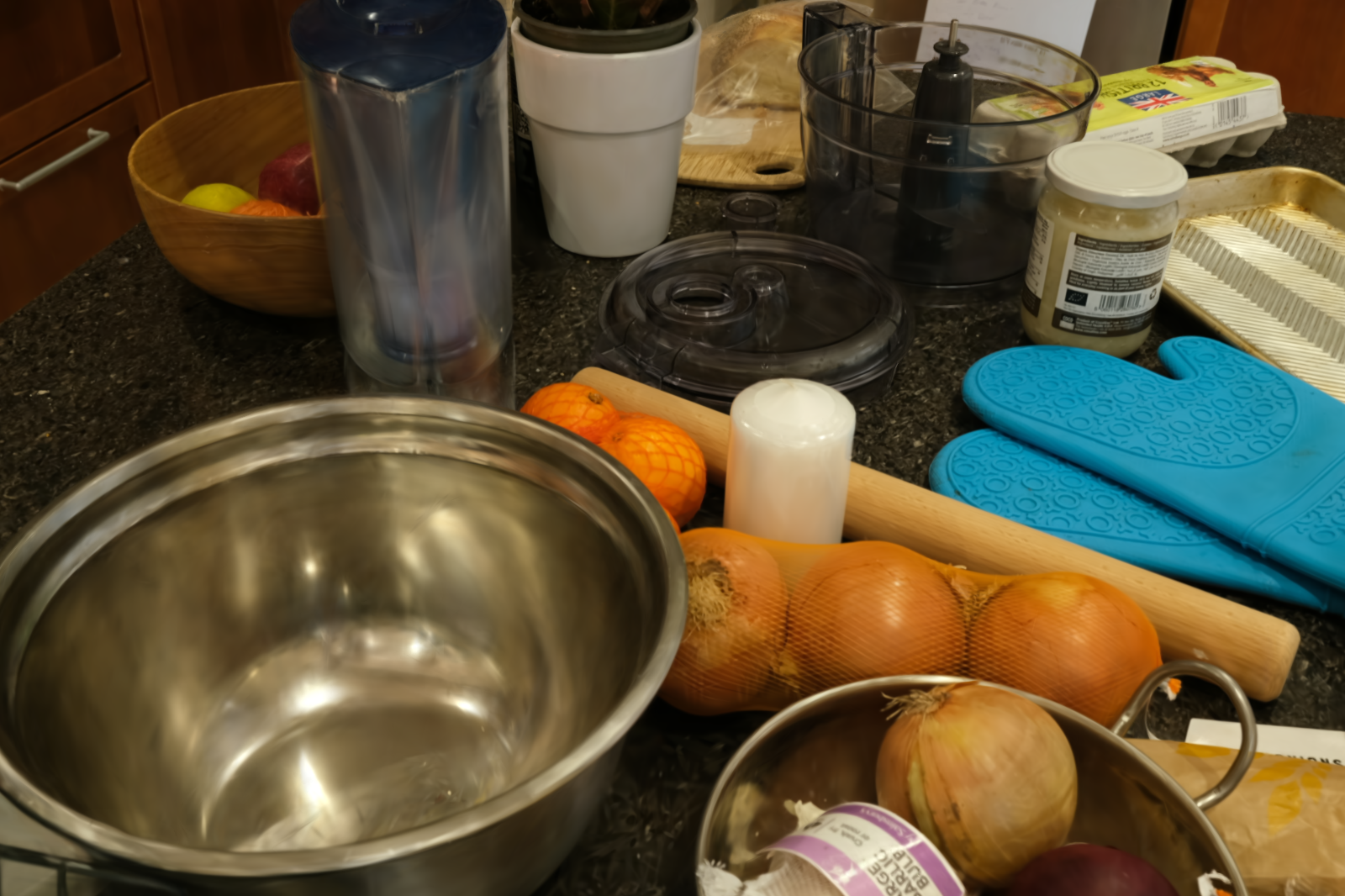}{\zoomx}{\zoomy}{\zoomboxx}{\zoomboxy}{1.2cm}{\zoomlevel}{0 0 0 0}
        \caption{Vol3DGS \cite{talegaonkar2025volumetrically}}
    \end{subfigure}
    \hfill
    \begin{subfigure}[t]{\percellwidth}
        \imagewithzoom{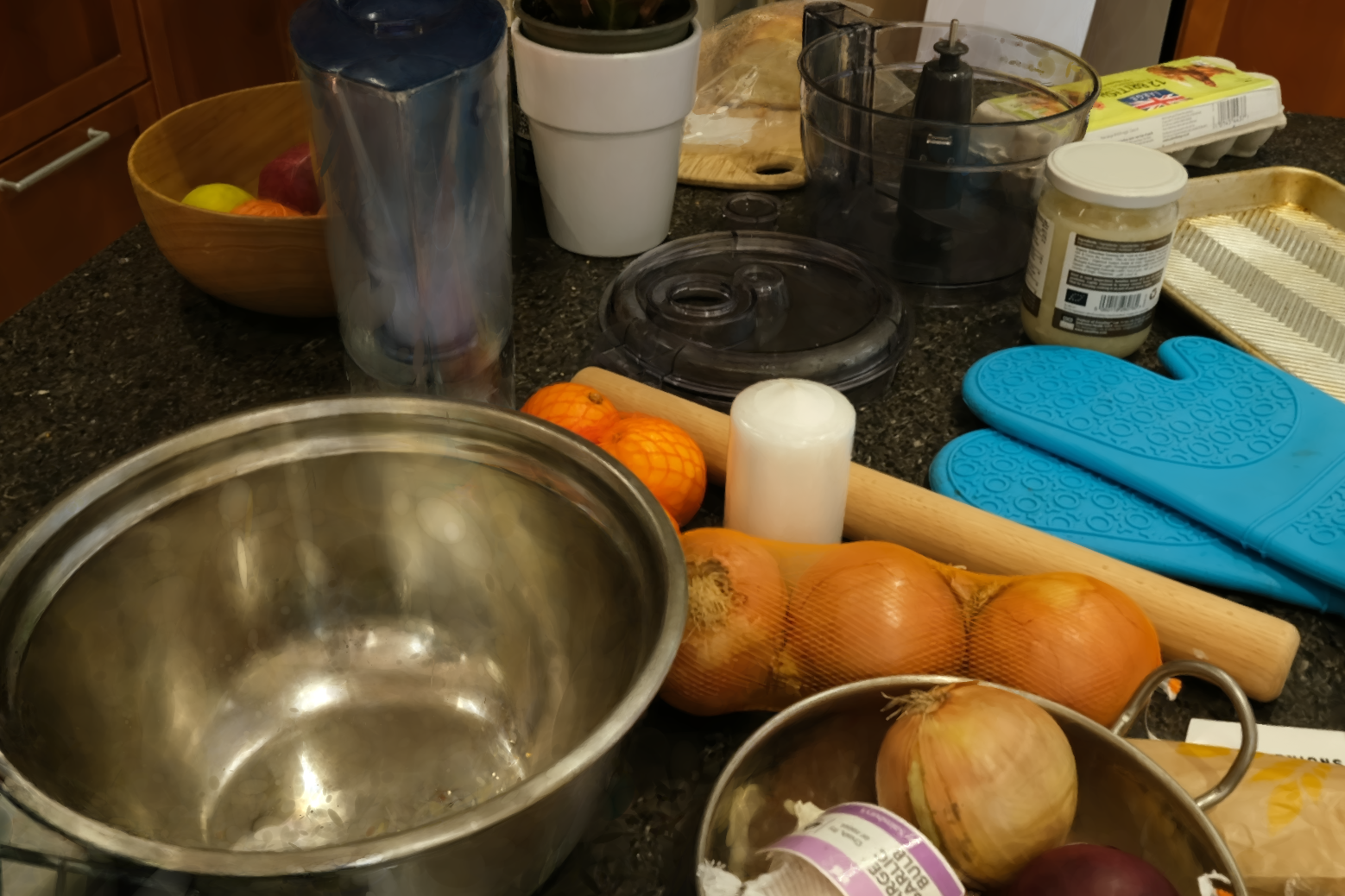}{\zoomx}{\zoomy}{\zoomboxx}{\zoomboxy}{1.2cm}{\zoomlevel}{0 0 0 0}
        \caption{EVER \cite{mai2025ever}}
    \end{subfigure}
    \hfill
    \begin{subfigure}[t]{\percellwidth}
        \imagewithzoom{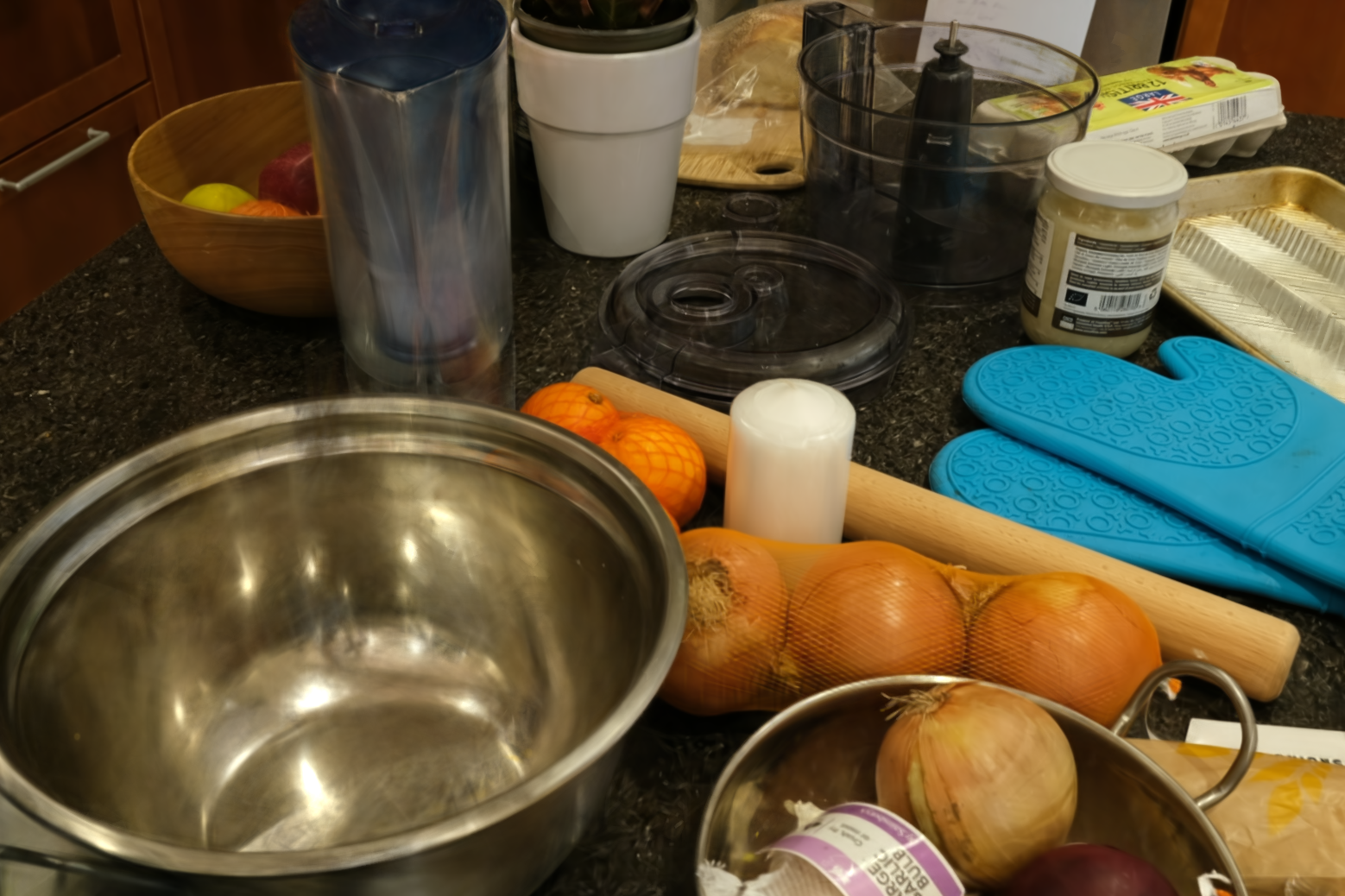}{\zoomx}{\zoomy}{\zoomboxx}{\zoomboxy}{1.2cm}{\zoomlevel}{0 0 0 0}
        \caption{Ours}
    \end{subfigure}

    \caption{Qualitative comparison of our method against two SOTA methods on scenes from Tanks and Temples~\cite{Knapitsch2017} and Mipnerf-360~\cite{barron2022mip}.}
    \label{fig:qualitative_comparison}
\end{figure*}
\subsection{Training and Densification}
\label{ssec:optimisation}
We optimize our model using an objective function that extends the 3DGS losses with a novel consistency regularizer, $\mathcal{L}_\textrm{consistency}$, to mitigate overfitting.
The moment-based transmittance reconstruction can be inaccurate for complex densities, and our regularizer enforces consistency between the predicted transmittance and the analytical density of individual Gaussian particles.
The full objective is
\begin{equation}
    \mathcal{L} = (1-\alpha) \, \mathcal{L} + \alpha \, \mathcal{L}_\textrm{D-SSIM} + \lambda \, \mathcal{L}_\textrm{consistency}.
\end{equation}
with $\alpha = 0.2$ and $\lambda = 0.1$.
The regularization term is derived from the physical constraint that the optical depth over any ray interval $[t_j, t_{j+1}]$, given by $ { \tau(t_n \to t_{j+1}) - \tau(t_n \to t_j) } $, must be greater than or equal to the analytical optical depth $\tau_{ij}$ of any single particle $i$ within that interval.
We penalize violations of this condition using $ { P_i = \sum_{j} \max\left(0, \tau_{ij} - \left(\tau_\theta(t_n \to t_{j+1}) - \tau_\theta(t_n \to t_j)\right) \right)^2 } $.
This penalty, summed over all visible particles, implicitly enforces the physical monotonicity of the learned transmittance $T_\theta$.

We further adapt the 3DGS Adaptive Density Control (ADC) to our density-based medium, where opacity is view-dependent.
Standard pruning fails as it relies on view-independent opacity.
We introduce a robust, view-independent metric for pruning based on a particle's opacity when viewed along its shortest axis: $ { o_i = 1-e^{-\sqrt{2\pi} w_i \min(s_x,s_y,s_z)} } $.
This criterion effectively prunes particles while preventing the creation of thin, view-dependent particles that cause overfitting.
We also invert this equation to initialize particle peak densities $w_\textrm{init}$ from the input point cloud.

Finally, we modify the cloning and splitting operations to preserve density integrity.
When cloning a particle, we correct the introduced density bias by halving its peak density, $ { w_i \gets \frac{1}{2} w_i } $.
We replace the stochastic splitting mechanism with a deterministic operation that splits a particle along its longest eigenvector.
The new means are offset by $ { \mean_\textrm{new} = \mean \pm \delta \, \dir_\textrm{split} } $ and the corresponding scale is reduced by $ { \Sigma_\textrm{new} = \gamma \, \Sigma_\textrm{split} } $.
We derived the optimal parameters ($ { \gamma \approx 0.639 } $, $ { \delta \approx 0.613 \cdot \Sigma_\textrm{split} } $) by numerically minimizing the change in the particle's opacity contribution, significantly improving optimization stability.
Please refer to Appendix \ref{ap:optimisation} for a detailed derivations and discussions.

\section{Evaluation}
\paragraph{Implementation.}
We use PyTorch \cite{paszke2019pytorch} and the 3DGS codebase \cite{kerbl20233d} for training. Our rasterizer is implemented via the Vulkan API, using Slang and its Slang-D extension \cite{bangaru2023slang} for automatic differentiation. 

\paragraph{Results.}
\begin{table*}[t]
  \centering
  \scriptsize
  \setlength{\tabcolsep}{5.5pt}
  \renewcommand{\arraystretch}{1.15}
  \caption{Quantitative comparison of Gaussian Splatting (GS) methods on three datasets.
  Higher is better for PSNR/SSIM ($ \uparrow $) and lower is better for LPIPS ($ \downarrow $). We used the publically available code to reproduce the results, where possible. Results with dagger (\textdagger) are taken from the respective publication instead.}
  \label{tab:eval-quantitative}
  \begin{tabular}{l | cccc cccc cccc}
    \toprule
    Method &
    \multicolumn{4}{c}{MipNeRF-360} &
    \multicolumn{4}{c}{Tanks \& Temples} &
    \multicolumn{4}{c}{DeepBlending} \\
    \cmidrule(lr){2-5} \cmidrule(lr){6-9} \cmidrule(lr){10-13}
    & PSNR $ \uparrow $ & SSIM $ \uparrow $ & LPIPS $ \downarrow $ & \# Points
    & PSNR $ \uparrow $ & SSIM $ \uparrow $ & LPIPS $ \downarrow $ & \# Points
    & PSNR $ \uparrow $ & SSIM $ \uparrow $ & LPIPS $ \downarrow $ & \# Points\\
    \midrule
    3DGS \cite{kerbl20233d}                         & 27.43 & 0.813 & 0.218 & 3.36$\times10^6$ & 23.72 & 0.846 & 0.178 & 1.78$\times10^6$ & 29.46 & 0.900 & 0.247 & 2.98$\times10^6$ \\
    StopThePop \cite{radl2024stopthepop}            & 27.31 & 0.814 & 0.213 & 3.29$\times10^6$ & 23.16 & 0.843 & 0.173 & 1.81$\times10^6$ & 29.92 & 0.905 & 0.234 & 2.81$\times10^6$ \\
    \midrule
    Vol3DGS \cite{talegaonkar2025volumetrically}    & 27.44 & 0.820 & 0.201 & 3.00$\times10^6$ & 23.67 & 0.851 & 0.174 & 1.06$\times10^6$ & 29.61 & 0.905 & 0.242 & 3.60$\times10^6$ \\
    EVER \cite{mai2025ever}                         & 25.60 & 0.772 & 0.299 & 3.89$\times10^6$ & 22.59 & 0.842 & 0.199 & 6.38$\times10^6$ & 28.12 & 0.891 & 0.353 & 2.54$\times10^6$ \\
    Don't Splat\textsuperscript{\textdagger} \cite{condor2025don}                & 27.32 & 0.793 & -- & -- & 22.09 & 0.797 & -- & -- & 28.06 & 0.878 & -- & -- \\
    \midrule
    Ours        & 25.96 & 0.760 & 0.245 & 2.12$\times10^6$ & 22.18 & 0.825 & 0.194 & 1.37$\times10^6$ & 29.14 & 0.900 & 0.248 & 2.69$\times10^6$ \\
    \bottomrule
  \end{tabular}
\end{table*}
We evaluate our method on three established novel view synthesis benchmarks: The nine scenes from Mip-NeRF 360 \cite{barron2022mip}, \emph{train} and \emph{truck} scenes from Tanks and Temples \cite{Knapitsch2017} as well as the \emph{drjohnson} and \emph{playroom} scenes from DeepBlending \cite{DeepBlending2018}.
Following previous literature, we report PSNR, SSIM, and LPIPS computed on held-out target views.
For a fair comparison, all methods are trained on the same input views and evaluated on the same test splits, using each method's provided codebase and hyperparameters (unless otherwise noted).
We compare against the standard Gaussian splatting baseline (3DGS) \cite{kerbl20233d}, StopThePop \cite{radl2024stopthepop}, EVER \cite{mai2025ever}, Vol3DGS \cite{talegaonkar2025volumetrically}, and Don't Splat Your Gaussians \cite{condor2025don}.

The results are summarized in \cref{tab:eval-quantitative}.
The values suggest competitive performance compared to volumetric extensions, whereas standard 3DGS and StopThePop often remain ahead in the evaluated quality metrics.
Nevertheless, \cref{fig:qualitative_comparison} shows that our approach can resolve complex light interactions more robustly than previous volumetric-aware approaches.
Semi-transparent effects like the reflection of the building on the windshield (row 1) and specular highlights on the metal bowl (row 3) are reconstructed with less noise and higher sharpness.
Volumetric-like regions such as distant trees (row 2) exhibit similar improvements.

To illustrate how our approach handles the challenging scenario of intersecting Gaussians, we conducted a small experiment on a synthetic scene consisting of six Gaussians that intersect and produce non-trivial volumetric color blending effects.
For a fair comparison of different renderers, we converted the scene to the Gaussian representation of the respective methods and optimized diffuse color, opacity and scale for 1000 iterations to allow the methods to best fit the data.
\cref{fig:volumetric_qualitative} shows a comparison to a ground-truth volumetric renderer (a).
Non-volumetric techniques like StopThePop~\cite{radl2024stopthepop} are unable to model the intersection of the Gaussians faithfully, as the naive rasterization inevitably has to blend the splats in order along the z-axis (b).
Volumetric-aware extensions like Vol3DGS~\cite{talegaonkar2025volumetrically} reduce sharp boundary artifacts, but still suffer from unrealistic colors (c).
In contrast, our method significantly improves the resulting color blending (d).

\paragraph{Ablation.}
A synthetic comparison (see \cref{fig:ablation_qualitative}) analyzes the choice of moment functions and the number of quadrature intervals ($N$). Power moments paired with our quadrature tend to overestimate splat visibility in certain views. Trigonometric moments prove more robust: $ { N=3 } $ achieves correct splat ordering but inaccurate sizing, while ${ N=5 } $ closely matches the ground truth.

Real-world ablations (see \cref{tab:ablation}) disable individual components. Removing regularization slightly degrades image metrics and increases runtime, attributed to a minor increase in outliers. Using the EWA geometric proxy, rather than our method, also slightly decreases aggregate metrics and increases runtime without significantly altering particle count due to overestimating small, translucent splats. Finally, reverting to the default ADC degrades metrics despite a substantial increase in total particle count and runtime, an effect we attribute to a high outlier count.

\begin{table}
    \centering
    \scriptsize
    \caption{Component analysis on the Tanks and Temples \textit{Truck}~\cite{Knapitsch2017} scene. We report results for ablations of each component.}
    \begin{tabular}{c|ccc}
        \toprule
        & PSNR $ \uparrow $ & Time [h] $ \downarrow$ & \# Points \\
        \midrule
        w/o Reg.  & 24.14 &  \phantom{0}3.45 & 1.08$\times 10^6$ \\ 
        EWA Geom.  & 24.10  & \phantom{0}6.20  & 1.59$\times 10^6$ \\ 
        Default ADC   & 23.50 & 12.88  &  3.82$\times 10^6$ \\ 
        \midrule
        Full Model & 24.25 & \phantom{0}2.83 & 1.06$\times10^6$ \\ 
        \bottomrule
    \end{tabular}
    \label{tab:ablation}
\end{table}

\paragraph{Limitations}
Although our approach improves volumetric consistency over pure splatting, it still inherits limitations from its underlying densification and camera assumptions.
On challenging scenes with fine, highly parallaxed structures such as flowers in Mip-NeRF 360, we observe under-reconstruction and residual blur, which we attribute to conservatively tuned adaptive density control (ADC).
As in other volumetric variants of 3DGS, our quality remains tightly coupled to these heuristic splitting and pruning thresholds, suggesting that more advanced densification strategies could further improve performance.
Moreover, our physically motivated density field makes the method more susceptible to calibration errors.
Similar to Vol3DGS, which reports opaque artifacts on miscalibrated scenes like \emph{treehill} in Mip-NeRF 360, our model tends to explain pose and distortion inconsistencies by localized overfitting, leading to below-average metrics compared to opacity-centric splatting baselines.
We therefore see improving ADC for our volumetric formulation and increasing robustness to imperfect camera calibration as complementary directions for future work.
\section{Conclusion}

We presented MB3DGS, a moment-based formulation for physically accurate, order-independent rendering of 3D Gaussian representations.
By modeling the combined density of overlapping Gaussians and reconstructing a continuous transmittance function from analytically computed moments, our approach overcomes the inherent limitations of alpha-blended splatting.
The proposed power-transformed moment recurrence and confidence-interval–based rasterization enable stable, efficient rendering while preserving the performance benefits of modern GPU rasterization pipelines.
Our results demonstrate significantly improved reconstruction fidelity, particularly in complex translucent and highly detailed regions where traditional splatting fails.
Overall, MB3DGS bridges the gap between rasterization and physically grounded volumetric rendering, providing a practical path toward accurate and real-time Gaussian-based scene representations.

\section{Acknowledgments}
This work has been funded by the Federal Ministry of Research, Technology and Space of Germany and the state of North Rhine-Westphalia as part of the Lamarr Institute for Machine Learning and Artificial Intelligence, by the European Regional Development Fund and the state of North Rhine-Westphalia under grant number EFRE-20801085 (Gen-AIvatar), by the state of North Rhine-Westphalia as part of the Excellency Start-up Center.NRW (U-BO-GROW) under grant number 03ESCNW18B, and additionally by the Ministry of Culture and Science North Rhine-Westphalia under grant number PB22-063A (InVirtuo 4.0: Experimental Research in Virtual Environments).

{
    \small
    \bibliographystyle{ieeenat_fullname}
    \bibliography{chapters/references}

@String{CVPR = "IEEE/CVF Conference on Computer Vision and Pattern Recognition (CVPR)"}

@String{ICCV = "IEEE/CVF International Conference on Computer Vision (ICCV)"}

@String{ECCV = "European Conference on Computer Vision (ECCV)"}

@String{WACV = "IEEE/CVF Winter Conference on Applications of Computer Vision (WACV)"}

@String{NeurIPS = "Advances in Neural Information Processing Systems (NeurIPS)"}

@String{SIGGRAPH = "Annual Conference on Computer Graphics and Interactive Techniques (SIGGRAPH)"}

@String{ICLR = "International Conference on Learning Representations (ICLR)"}

@String{TOG = "ACM Transactions on Graphics (TOG)"}

@String{CGF = "Computer Graphics Forum (CGF)"}

@inproceedings{zwicker2001ewa,
  title={EWA Volume Splatting},
  author={Zwicker, Matthias and Pfister, Hanspeter and Van Baar, Jeroen and Gross, Markus},
  booktitle={Visualization},
  year={2001},
  organization={IEEE}
}

@inproceedings{mildenhall2020nerf,
  title={NeRF: Representing Scenes as Neural Radiance Fields for View Synthesis},
  author={Mildenhall, Ben and Srinivasan, Pratul P and Tancik, Matthew and Barron, Jonathan T and Ramamoorthi, Ravi and Ng, Ren},
  booktitle=ECCV,
  year={2020}
}

@article{kerbl20233d,
  title={3D Gaussian splatting for real-time radiance field rendering.},
  author={Kerbl, Bernhard and Kopanas, Georgios and Leimk{\"u}hler, Thomas and Drettakis, George},
  journal=TOG,
  volume={42},
  number={4},
  year={2023}
}

@inproceedings{celarek2025does,
  title={Does 3D Gaussian Splatting Need Accurate Volumetric Rendering?},
  author={Celarek, Adam and Kopanas, George and Drettakis, George and Wimmer, Michael and Kerbl, Bernhard},
  booktitle=CGF,
  year={2025},
  organization={Wiley Online Library}
}

@inproceedings{talegaonkar2025volumetrically,
  title={Volumetrically Consistent 3D Gaussian Rasterization},
  author={Talegaonkar, Chinmay and Belhe, Yash and Ramamoorthi, Ravi and Antipa, Nicholas},
  booktitle=CVPR,
  year={2025}
}

@article{radl2024stopthepop,
  title={Stopthepop: Sorted gaussian splatting for view-consistent real-time rendering},
  author={Radl, Lukas and Steiner, Michael and Parger, Mathias and Weinrauch, Alexander and Kerbl, Bernhard and Steinberger, Markus},
  journal=TOG,
  volume={43},
  number={4},
  year={2024},
  publisher={ACM New York, NY, USA}
}

@inproceedings{yu2024mip,
  title={Mip-splatting: Alias-free 3d gaussian splatting},
  author={Yu, Zehao and Chen, Anpei and Huang, Binbin and Sattler, Torsten and Geiger, Andreas},
  booktitle=CVPR,
  year={2024}
}

@inproceedings{liang2024analytic,
  title={Analytic-splatting: Anti-aliased 3d gaussian splatting via analytic integration},
  author={Liang, Zhihao and Zhang, Qi and Hu, Wenbo and Zhu, Lei and Feng, Ying and Jia, Kui},
  booktitle=ECCV,
  year={2024},
  organization={Springer}
}

@inproceedings{wu20253dgut,
  title={3dgut: Enabling distorted cameras and secondary rays in gaussian splatting},
  author={Wu, Qi and Esturo, Janick Martinez and Mirzaei, Ashkan and Moenne-Loccoz, Nicolas and Gojcic, Zan},
  booktitle=CVPR,
  year={2025}
}

@inproceedings{huang20242d,
  title={2d gaussian splatting for geometrically accurate radiance fields},
  author={Huang, Binbin and Yu, Zehao and Chen, Anpei and Geiger, Andreas and Gao, Shenghua},
  booktitle={ACM SIGGRAPH Conference Papers},
  year={2024}
}

@inproceedings{hahlbohm2025efficient,
  title={Efficient Perspective-Correct 3D Gaussian Splatting Using Hybrid Transparency},
  author={Hahlbohm, Florian and Friederichs, Fabian and Weyrich, Tim and Franke, Linus and Kappel, Moritz and Castillo, Susana and Stamminger, Marc and Eisemann, Martin and Magnor, Marcus},
  booktitle=CGF,
  year={2025},
  organization={Wiley Online Library}
}

@inproceedings{housort,
  title={Sort-free Gaussian Splatting via Weighted Sum Rendering},
  author={Hou, Qiqi and Rauwendaal, Randall and Li, Zifeng and Le, Hoang and Farhadzadeh, Farzad and Porikli, Fatih and Bourd, Alexei and Said, Amir},
  booktitle=ICLR,
  year={2025}
}

@inproceedings{kheradmand2025stochasticsplats,
  title={StochasticSplats: Stochastic Rasterization for Sorting-Free 3D Gaussian Splatting},
  author={Kheradmand, Shakiba and Vicini, Delio and Kopanas, George and Lagun, Dmitry and Yi, Kwang Moo and Matthews, Mark and Tagliasacchi, Andrea},
  booktitle=ICCV,
  year={2025}
}

@article{moenne20243d,
  title={3d gaussian ray tracing: Fast tracing of particle scenes},
  author={Moenne-Loccoz, Nicolas and Mirzaei, Ashkan and Perel, Or and de Lutio, Riccardo and Martinez Esturo, Janick and State, Gavriel and Fidler, Sanja and Sharp, Nicholas and Gojcic, Zan},
  journal=TOG,
  volume={43},
  number={6},
  year={2024},
  publisher={ACM New York, NY, USA}
}

@article{yu2024gaussian,
  title={Gaussian opacity fields: Efficient adaptive surface reconstruction in unbounded scenes},
  author={Yu, Zehao and Sattler, Torsten and Geiger, Andreas},
  journal=TOG,
  volume={43},
  number={6},
  year={2024},
  publisher={ACM New York, NY, USA}
}

@inproceedings{mai2025ever,
  title={Ever: Exact volumetric ellipsoid rendering for real-time view synthesis},
  author={Mai, Alexander and Hedman, Peter and Kopanas, George and Verbin, Dor and Futschik, David and Xu, Qiangeng and Kuester, Falko and Barron, Jonathan T and Zhang, Yinda},
  booktitle=ICCV,
  year={2025}
}

@article{condor2025don,
  title={Don't Splat your Gaussians: Volumetric Ray-Traced Primitives for Modeling and Rendering Scattering and Emissive Media},
  author={Condor, Jorge and Speierer, Sebastien and Bode, Lukas and Bozic, Aljaz and Green, Simon and Didyk, Piotr and Jarabo, Adrian},
  journal=TOG,
  volume={44},
  number={1},
  year={2025},
  publisher={ACM New York, NY, USA}
}

@inproceedings{blanc2025raygauss,
  title={Raygauss: Volumetric gaussian-based ray casting for photorealistic novel view synthesis},
  author={Blanc, Hugo and Deschaud, Jean-Emmanuel and Paljic, Alexis},
  booktitle=WACV,
  year={2025},
  organization={IEEE}
}

@article{sun2025stochastic,
  title={Stochastic Ray Tracing of 3D Transparent Gaussians},
  author={Sun, Xin and Georgiev, Iliyan and Fei, Yun and Ha{\v{s}}an, Milo{\v{s}}},
  journal={arXiv preprint arXiv:2504.06598},
  year={2025}
}

@inproceedings{peters2015moment,
  title={Moment shadow mapping},
  author={Peters, Christoph and Klein, Reinhard},
  booktitle={ACM SIGGRAPH Symposium on Interactive 3D Graphics and Games (I3D)},
  year={2015}
}

@inproceedings{peters2016beyond,
  title={Beyond hard shadows: Moment shadow maps for single scattering, soft shadows and translucent occluders},
  author={Peters, Christoph and Munstermann, Cedrick and Wetzstein, Nico and Klein, Reinhard},
  booktitle={ACM SIGGRAPH Symposium on Interactive 3D Graphics and Games (I3D)},
  year={2016}
}

@article{peters2017improved,
  title={Improved moment shadow maps for translucent occluders, soft shadows and single scattering},
  author={Peters, Christoph and M{\"u}nstermann, Cedrick and Wetzstein, Nico and Klein, Reinhard},
  journal={Journal of Computer Graphics Techniques (JCGT)},
  volume={6},
  number={1},
  year={2017}
}

@inproceedings{peters2017non,
  title={Non-linearly quantized moment shadow maps},
  author={Peters, Christoph},
  booktitle={High Performance Graphics (HPG)},
  year={2017}
}

@article{munstermann2018moment,
  title={Moment-Based Order-Independent Transparency},
  author={M{\"u}nstermann, Cedrick and Krumpen, Stefan and Klein, Reinhard and Peters, Christoph},
  journal={ACM on Computer Graphics and Interactive Techniques},
  volume={1},
  number={1},
  year={2018}
}

@article{worchel2025moment,
  title={Moment Bounds are Differentiable: Efficiently Approximating Measures in Inverse Rendering},
  author={Worchel, Markus and Alexa, Marc},
  journal=TOG,
  volume={44},
  number={4},
  year={2025},
  publisher={ACM New York, NY, USA}
}

@inproceedings{carpenter1984buffer,
  title={The A-buffer, an antialiased hidden surface method},
  author={Carpenter, Loren},
  booktitle=SIGGRAPH,
  year={1984}
}

@article{everitt2001interactive,
  title={Interactive order-independent transparency},
  author={Everitt, Cass},
  journal={White paper, NVIDIA},
  volume={2},
  number={6},
  year={2001}
}

@article{bavoil2008order,
  title={Order independent transparency with dual depth peeling},
  author={Bavoil, Louis and Myers, Kevin},
  journal={NVIDIA OpenGL SDK},
  volume={1},
  number={12},
  year={2008}
}

@article{mcguire2013weighted,
  title={Weighted blended order-independent transparency},
  author={McGuire, Morgan and Bavoil, Louis},
  journal={Journal of Computer Graphics Techniques (JCGT)},
  volume={2},
  number={4},
  year={2013}
}

@inproceedings{tsopouridis2024deep,
  title={Deep and fast approximate order independent transparency},
  author={Tsopouridis, Grigoris and Vasilakis, Andreas A and Fudos, Ioannis},
  booktitle=CGF,
  volume={43},
  number={6},
  year={2024},
  organization={Wiley Online Library}
}

@article{bangaru2023slang,
  title={Slang. d: Fast, modular and differentiable shader programming},
  author={Bangaru, Sai Praveen and Wu, Lifan and Li, Tzu-Mao and Munkberg, Jacob and Bernstein, Gilbert and Ragan-Kelley, Jonathan and Durand, Fredo and Lefohn, Aaron and He, Yong},
  journal=TOG,
  volume={42},
  number={6},
  year={2023},
  publisher={ACM New York, NY, USA}
}

@inproceedings{barron2021mip,
  title={Mip-nerf: A multiscale representation for anti-aliasing neural radiance fields},
  author={Barron, Jonathan T and Mildenhall, Ben and Tancik, Matthew and Hedman, Peter and Martin-Brualla, Ricardo and Srinivasan, Pratul P},
  booktitle=ICCV,
  pages={5855--5864},
  year={2021}
}

@inproceedings{barron2022mip,
  title={Mip-nerf 360: Unbounded anti-aliased neural radiance fields},
  author={Barron, Jonathan T and Mildenhall, Ben and Verbin, Dor and Srinivasan, Pratul P and Hedman, Peter},
  booktitle=CVPR,
  year={2022}
}

@inproceedings{barron2023zip,
  title={Zip-nerf: Anti-aliased grid-based neural radiance fields},
  author={Barron, Jonathan T and Mildenhall, Ben and Verbin, Dor and Srinivasan, Pratul P and Hedman, Peter},
  booktitle=ICCV,
  year={2023}
}

@article{barron2025power,
  title={A Power Transform},
  author={Barron, Jonathan T},
  journal={arXiv preprint arXiv:2502.10647},
  year={2025}
}

@inproceedings{neff2021donerf,
  title={DONeRF: Towards real-time rendering of compact neural radiance fields using depth oracle networks},
  author={Neff, Thomas and Stadlbauer, Pascal and Parger, Mathias and Kurz, Andreas and Mueller, Joerg H and Chaitanya, Chakravarty R Alla and Kaplanyan, Anton and Steinberger, Markus},
  booktitle=CGF,
  volume={40},
  number={4},
  year={2021},
  organization={Wiley Online Library}
}

@inproceedings{ye2024absgs,
  title={Absgs: Recovering fine details in 3d gaussian splatting},
  author={Ye, Zongxin and Li, Wenyu and Liu, Sidun and Qiao, Peng and Dou, Yong},
  booktitle={ACM International Conference on Multimedia (MM)},
  year={2024}
}

@inproceedings{rota2024revising,
  title={Revising densification in gaussian splatting},
  author={Rota Bul{\`o}, Samuel and Porzi, Lorenzo and Kontschieder, Peter},
  booktitle=ECCV,
  year={2024},
  organization={Springer}
}

@article{Knapitsch2017,
    author    = {Arno Knapitsch and Jaesik Park and Qian-Yi Zhou and Vladlen Koltun},
    title     = {Tanks and Temples: Benchmarking Large-Scale Scene Reconstruction},
    journal   = TOG,
    volume    = {36},
    number    = {4},
    year      = {2017},
}

@article{DeepBlending2018,
  author = {Hedman, Peter and Philip, Julien and Price, True and Frahm, Jan-Michael and Drettakis, George and Brostow, Gabriel},
  title = {Deep Blending for Free-viewpoint Image-based Rendering},
  journal = TOG,
  publisher = {ACM},
  volume    = {37},
  number    = {6},
  year      = {2018}
}

@article{paszke2019pytorch,
  title={Pytorch: An imperative style, high-performance deep learning library},
  author={Paszke, Adam and Gross, Sam and Massa, Francisco and Lerer, Adam and Bradbury, James and Chanan, Gregory and Killeen, Trevor and Lin, Zeming and Gimelshein, Natalia and Antiga, Luca and others},
  journal=NeurIPS,
  volume={32},
  year={2019}
}

@inproceedings{tari2005unified,
  title={A unified approach to the moments based distribution estimation--unbounded support},
  author={Tari, {\'A}rp{\'a}d and Telek, Mikl{\'o}s and Buchholz, Peter},
  booktitle={European Workshop on Performance Engineering},
  pages={79--93},
  year={2005},
  organization={Springer}
}

@book{horn2012matrix,
  title={Matrix analysis},
  author={Horn, Roger A and Johnson, Charles R},
  year={2012},
  publisher={Cambridge university press}
}

@book{gradshteyn2014table,
  title={Table of integrals, series, and products},
  author={Gradshteyn, Izrail Solomonovich and Ryzhik, Iosif Moiseevich},
  year={2014},
  publisher={Academic press}
}
}

\newpage
\setcounter{section}{0}
\renewcommand\thesection{\Alph{section}}
\section{Ray Parameterized Density}
\label{ap:raydensity}
Central to our approach is the re-parameterization of the density as a sum 1D Gaussians along a ray, parameterized by distance (Eq.\ 4-6).
This section provides details about the derivation of this result and contextualizes it with respect to prior work.

Let a ray be defined as ${r(t) = \origin + t \, \dir }$, parameterized by distance $t$ from origin $\origin$ in direction $\dir$. The density at any position $\pos$ is a Gaussian mixture:
\begin{equation*}    
\sigma(\pos) = \sum_{i} w_i \, G(\pos \,|\, \mean_i, \cov_i)
\end{equation*}
where $w_i$, $\mean_i$, and $\cov_i$ are the weight, mean, and covariance of the $i$-th component. 

We first analyze the density of a single component along the ray. Substituting $r(t)$ gives:
\begin{equation*}    
\sigma_i(r(t)) = w_i \, e^{-\frac{1}{2} (\origin + t \dir - \mean_i)^T \cov_i^{-1} (\origin + t \dir - \mean_i)}.
\end{equation*}
As the following derivation applies independently to each component, we now drop the index $i$ from $w_i$, $\mean_i$, $\cov_i$ for notational simplicity.

Let ${ \bm{v} = \origin - \mean }$, substitute it in the exponent and expand it into its quadratic form:
\begin{align*}
    &(\bm{v} + t\cdot \dir)^T \invcov (\bm{v} + t \cdot \dir) \\
    &= \bm{v}^T \invcov \bm{v} + t \, \bm{v}^T \invcov \dir + t \, \dir^T \invcov \bm{v} + t^2 \,\dir^T \invcov \bm{v}
\end{align*}
Since $\invcov$ is symmetric, ${\bm{v}^T \invcov \dir = \dir^T \invcov \bm{v}}$. So, the cross terms combine to ${2 t \, \dir^T \invcov \bm{v}}$. The exponent is thus
\begin{equation*}
    -\frac{1}{2} \left( \bm{v}^T \invcov \bm{v} + 2t\, \dir^T \invcov \bm{v} + t^2 \, \dir^T \invcov \dir\right)
\end{equation*}
This is quadratic in $t$: ${A \, t^2 + B\, t + C}$, where
\begin{align*}
    A = -\frac{1}{2} \dir^T \invcov \dir, \ \
    B = -\dir^T \invcov \bm{v}, \ \
    C = -\frac{1}{2} \bm{v}^T \invcov \bm{v}.
\end{align*}
So, the $i$-th particle density as a function of distance $t$ is ${\sigma(t) = w \, e^{A \, t^2 + B \, t +C}}$. We want to rewrite this into the standard form of a 1D Gaussian 
\begin{equation*}
    \w \, e^{-\frac{(t - \m)^2}{2 \s^2}}
\end{equation*}
Recall the quadratic expansion of the 1D Gaussian exponent:
\begin{equation*}
    -\frac{(t - \m)^2}{2 \s^2} = -\frac{1}{2\s^2}\,t^2 + \frac{\m}{\s^2} \, t - \frac{(\m)^2}{2\s^2} 
\end{equation*}
We derive the 1D parameters by matching the exponent $At^2+Bt+C$ to the target 1D Gaussian form. Comparing the $t^2$ coefficient yields the variance $\s^2$:
\begin{equation*}
    A = -\frac{1}{2\s^2} \Rightarrow \s^2 = -\frac{1}{2A} = \frac{1}{\dir^T \invcov \dir}
\end{equation*}
Comparing the t coefficient yields the mean $\m$:
\begin{equation*}
    B = -\frac{\m}{2\s^2} \Rightarrow \m = B \, \s^2 = -\frac{\dir^T \invcov \bm{v}}{\dir^T \invcov \dir}
\end{equation*}
which leaves ${C = -\mu^2 / (2\s^2)}$ as the coefficient that is independent of $t$.

The 1D Gaussian weight $\w$ is determined by the peak amplitude, which we find by completing the square on the exponent:
\begin{equation*}
    At^2 + Bt + C = A\left(t + \frac{B}{2A}\right)^2 + \left(C - \frac{B^2}{4A}\right).
\end{equation*}
The term independent of $t$, ${K=C- B^2/(4A)}$, is the value of the exponent at its peak where ${ t = \m }$. Substituting the known expressions for $A$, $B$, and $C$ gives:
\begin{equation*}
      K = -\frac{1}{2} \bm{v}^T \invcov \bm{v} + \frac{1}{2} \frac{(\dir^T \invcov \bm{v})^2}{\dir^T \invcov \dir}.
\end{equation*}
The density as a function of ray distance $t$ thus simplifies to ${w \, e^K \, e^{-(t-\m)^2/(2\s^2)}}$. Finally, the 1D Gaussian weight is $\w = w \, e^{K}$.

Recall the covariance definition ${\cov = \bm{R} \bm{S}\bm{S}^T\bm{R}^T}$, where $\bm{R}$ is a rotation matrix and $\bm{S}$ is a diagonal scale matrix. To avoid explicit inversion of $\cov$, we follow \cite{moenne20243d} by defining transformed vectors:
\begin{equation*}    
\origin_g = \bm{S}^{-1}\bm{R}^T(\origin-\mean) \quad \text{and} \quad \dir_g = \bm{S}^{-1}\bm{R}^T\dir
\end{equation*}
This simplifies the above quadratic terms, yielding ${\bm{v}^T \invcov \bm{v} = \origin_g^T \origin_g}$, ${\dir^T \invcov \dir = \dir_g^T \dir_g}$, and ${\dir^T \invcov \bm{v} = \dir_g^T \origin_g}$. Substituting these into the expressions for $\m$, $\s$, and $K$ directly yields Equations 4-6 from the main paper.

This 1D re-parameterization builds on related concepts in prior work. StopThePop~\cite{radl2024stopthepop} and 3D Gaussian Raytracing~\cite{moenne20243d} use the mean of the 1D Gaussian (the point of highest amplitude) to define a single distance from the camera to the particle. Other opacity-based methods, such as Vol3DGS~\cite{talegaonkar2025volumetrically} and Gaussian Opacity Fields~\cite{yu2024gaussian}, also derive a similar 1D parameterization. Gaussian Opacity Fields~\cite{yu2024gaussian} further modifies this 1D Gaussian to mimic a threshold function, which is then used to derive a single particle's transmittance. Crucially, all these methods use alpha-blending for rendering. This approach does not properly resolve intersections between Gaussian particles, a limitation addressed by our volumetric formulation.
\section{Quadrature}
\label{ap:quadrature}

This section derives the quadrature for the volume rendering equation (Eq.\ 7, main paper), details the opacity rescaling (Eq.\ 15), and describes the interval spacing for the Eq.\ 7 quadrature. While the main paper uses compact notation $T(t)$ for transmittance over the interval $[t_n, t]$, this section uses the more explicit notation $T(a \to b)$ to denote transmittance over the interval $[a,b]$.

We substitute the definitions for the medium's density $\sigma(\pos)$ (Eq.\ 1) and emitted radiance $L_e(\pos, \dir)$ (Eq.\ 3) into the integrand of the volume rendering equation (Eq.\ 2). This substitution simplifies the $\sigma(\pos) L_e(\pos, \dir)$ term, as the total density $\sigma(\pos)$ cancels out, leaving a sum of radiance contributions from each particle:
\begin{equation*}
    \sigma(\pos_t) \, L_e(\pos_t, \dir) = \sum_{i=1}^n \sigma_i(\pos_t) \, (L_e)_i(\dir)
\end{equation*}
Substituting this result back into Eq.\ 2, the linearity of integration permits exchanging the summation and the integral:
\begin{equation*}    
L = \sum_{i=1}^n \underbrace{\int_{t_n}^{t_f} T(t_n \to t) \, \sigma_i(\pos_t) \, (L_e)_i(\dir) \, \dd t}_{L_i}
\end{equation*}
Discretize the interval $[t_n, t_f]$ into $N$ small, contiguous segments where the $j$-th segment spans from $t_j$ to $t_{j+1}$ with length ${\Delta_j = t_{j+1} - t_j}$. The integral for $L_i$ becomes a sum over these segments:
\begin{equation*}    
L_i = \sum_{j=1}^{N} \int_{t_j}^{t_{j+1}} T(t_n \to t) \, \sigma_i(\pos_t) \, (L_e)_i(\dir) \, \dd t.
\end{equation*}
The transmittance ${T(t_n \to t)}$ can be split into the product ${T(t_n \to t_j) \, T(t_j \to t)}$. The $j$-th segment of the integral for $L_i$ becomes
\begin{equation*}
L_{i,j} = \int_{t_j}^{t_{j+1}} T(t_n \to t_j) \, T(t_j \to t) \, \sigma_i(\pos_t) \, (L_e)_i(\dir) \, \dd t
\end{equation*}
Within each small segment $[t_j, t_{j+1}]$, \textbf{we assume the density $\sigma(\pos_t)$ is piecewise constant} and equal to its value at the start of the interval, $\sigma(\pos_{t_j})$. 
Since each Gaussian has an infinite support, this assumption about the total density also implies that each particles local density is piecewise constant.
Therefore terms $T(t_n \to t_j)$, $\sigma_i(\pos_{t_j})$, and $(L_e)_i(\dir)$ are constant within this interval and can be pulled out:
\begin{equation*}    
L_{i,j}\approx T(t_n \to t_j) \, \sigma_i(\pos_{t_j}) \, (L_e)_i(\dir) \int_{t_j}^{t_{j+1}} T(t_j \to t) \, \dd t
\end{equation*}
The remaining integral under this assumption evaluates as:
\begin{align*}
    &\int_{t_j}^{t_{j+1}} T(t_j \to t) \, \dd t = \int_{t_j}^{t_{j+1}} e^{-\int_{t_j}^{t} \sigma(\pos_s) \, \dd s} \, \dd t \\
    = \ & \int_{t_j}^{t_{j+1}} e^{-\sigma(\pos_{t_j})(t - t_j)} \, \dd t = \frac{1 - e^{-\sigma(\pos_{t_j})\Delta_j}}{\sigma(\pos_{t_j})}
\end{align*}
The term ${1 - e^{-\sigma(\pos_{t_j})\delta_j}}$ can be expressed using the transmittance over the segment ${T(t_j \to t_{j+1})}$ since with constant density it is ${T(t_j \to t_{j+1}) = e^{-\sigma(\pos_{t_j})\delta_j}}$. Substituting this back, the contribution of segment $j$ to particle $i$ is:
\begin{equation*}
L_{i,j} \approx T(t_n \to t_j) \, \sigma_i(\pos_{t_j}) \, (L_e)_i(\dir) \, \left(\frac{1 - T(t_j \to t_{j+1})}{\sigma(\pos_{t_j})}\right)
\end{equation*}
Using the multiplicative property of the transmittance again, the visibility can be evaluated using transmittance values from the ray's near bound to the interval's left and right edge:
\begin{align*}    
& T(t_n \to t_j)\, (1 - T(t_j \to t_{j+1})) \\
= \ & T(t_n \to t_j) - T(t_n \to t_{j+1}).
\end{align*}
Substituting this back and summing the contributions $L_{i,j}$ over all segments yields the quadrature presented in Eq.\ 7:
\begin{equation*}    
L_{i} \approx \sum_{j=1}^N  \left(T(t_n \to t_j) - T(t_n \to t_{j+1}) \right) \, \frac{\sigma_i(t_j)}{\sigma(t_j)} \, (L_e)_i(d).
\end{equation*}

A direct evaluation of $\sigma(t_j)$ is not possible in a rasterization setting. However, under the assumption of piecewise-constant density, we can solve ${T(t_j \to t_{j+1}) = e^{-\sigma(\pos_{t_j})\Delta_j} }$ for $\sigma(t_j)$ given the transmittance over the interval ${T(t_j \to t_{j+1})}$.
The multiplicative property of the transmittance also allows this expression to be written in terms of transmittance starting from the ray's near bound. We therefore evaluate the total density within $j$-th interval to be
\begin{equation*}
\sigma(t_j) \approx -\frac{1}{\Delta_j} \log\left(\frac{T(t_n \to t_{j+1})}{T(t_n \to t_j)}\right).    
\end{equation*}
Recall that we estimate the optical depth ${\tau(t_n \to t) = \int_{t_n}^t \sigma(x_s) \, \dd s}$ at distance $t$ from the density moments $m_0, \dots, m_k$. The transmittance under this estimation of the optical depth is ${T(t_n \to t) = e^{-\tau(t_n \to t)}}$. Substituting this relationship into the estimate of the total transmittance gives the numerically more stable expression 
\begin{equation*}
    \sigma(t_j) = \frac{1}{\Delta_j}\left( \tau(t_n \to t_{j+1}) - \tau(t_n \to t_j)\right)
\end{equation*}

\paragraph{Opacity Rescaling} 
Previous work on order-independent transparency ~\cite{munstermann2018moment, mcguire2013weighted} observed that an under- or overestimation of the transmittance causes the volumetric scene's overall opacity to fluctuate across the image. To stabilize the visual appearance, they rescale the radiance using the ratio of the true scene opacity ${1-T(t_n \to t_f)}$, to the estimated scene opacity, $O$:
\begin{equation*}    
\frac{1-T(t_n \to t_f)}{O}
\end{equation*}
The true transmittance ${T(t_n \to t_f})$ is recovered from the zeroth moment as ${e^{-m_0}}$. The estimated opacitay $O$ is the sum of particle opacities, ${O = \sum_i O_i}$.

Following prior work, we apply the opacity ratio to rescale the observed radiance. The final radiance prediction in Eq.\ 15 of the main paper is:
\begin{equation*}
\hat{L} =  \frac{1-e^{-m_0}}{\max\left(\epsilon, \sum_{i=1}^n O_i\right)} \sum_{i=1}^n L_i + e^{-m_0} \, L_\text{bg}
\end{equation*}
where the denominator is clamped to ${\epsilon >0}$ to prevent division by zero. Each $O_i$ is computed concurrently with $L_i$ in single render pass by accumulating a homogeneous "color" (\ie appending $1$ to the RGB components). Under the assumption of a piecewise-constant density, this opacity for the $i$-th particle is given by the quadrature
\begin{equation*}
    O_i =  \sum_{j=1}^N \left(T(t_n \to t_j) - T(t_n \to t_{j+1} )\right) \, \frac{\sigma_i(x_{t_j})}{\sigma(t_j)}.
\end{equation*}

\paragraph{Sample Spacing}
To numerically evaluate the integral $L_i$ for each particle, we require an efficient sampling strategy. We use inverse transform sampling  to concentrate samples $\{t_j\}_{j=1}^N$ in the particle's high-density region along the ray.

First, a set of standard normal samples $\{x_j\}$ is pre-computed once by applying the inverse normal CDF, $\Phi^{-1}$, to uniformly spaced samples on the interval $[0,1]$. In the shader, these are then transform using
\begin{equation*}
    t_j = \mi + \kappa \, \si \, x_j
\end{equation*}
where ${ \kappa \geq 1 }$ is a scaling parameter ($\kappa = 3$ is used in all experiments). For ${ \kappa = 1 }$, this corresponds to importance sampling the particle's density $\sigma_i$. Increasing $\kappa$ broadens the sampling distribution, transitioning it towards a more uniform distribution.

Finally, the samples $\{t_j\}$ are sampled to the integration interval $[t_n, t_f]$. This is necessary to respect the integration bounds, as particle density outside this interval does not contribute to $L_i$. For partially visible Gaussians (\ie those truncated by the bounds), this clamping clusters samples at the boundaries $t_n$ or $t_f$, effectively truncating the sampling distribution to the visible segment.

\section{Analytical Density Moments}
\label{ap:powermoments}

This section proves the lower-bound on the density power-moment, which justifies integrating $\hat{g}(t)^k$ against the density instead of $t^k$. We also derive the recurrence relation used to evaluate the power moments (Eqs.\ 10-12 in the main paper) and the closed-form expression for the trigonometric moments (Eq.\ 14 in the main paper). Additionally, this section provides further details on the Taylor approximation used to evaluate $\erf$ for complex arguments and compares our density moments to those described by Münstermann~\etal~\cite{munstermann2018moment}. All derivations utilize the density parameterization with respect to the ray distance presented in Eq.\ 4 of the main paper.

\paragraph{Lower bounds on Power Moments}
If ${ t_f \geq \mi + \sqrt{2}\si }$ and ${ t_n \leq \mi }$ (holds \eg for ${ t_n = 0 }$, ${ t_f = \infty} $ and any ${ \mi \geq 0 }$ but also justifiable in practical settings due to clipping and ${ t_f = 10^6 }$), choose interval ${ I =[\mi, \mi + \sqrt{2}\si] \subset [t_n,t_f] }$. If ${ \sqrt{2}\si > 0 }$, then for ${ t \in I }$ we have ${ t > \mi }$ and by monotonicity of ${ t^k \leq (\mi)^k }$. Therefore,
\begin{equation*}
    (m_k)_i \geq \wi \int_I t^k e^{-\frac{(t-\mi)^2}{2\si^2}} \, \dd t \geq \wi \, \mi^k \, \int_{I} e^{-\frac{(t-\mi)^2}{2\si^2}} \, \dd t.
\end{equation*}
The remaining integral is just a (truncated) Gaussian mass which we can express in terms of the error function. Change variables ${ u=(t-\mi)/(2\si^2) }$. Then
\begin{equation*}
   ( m_k)_i \geq  \wi \, \mi^k \, \sqrt{2} \, \si \, \int_0^1 e^{-u^2} \, \dd u.
\end{equation*}
This lead to a lower-bound in the value of the $k$-th moment
\begin{equation*}
    (m_k)_i \geq c_0 \, \wi \, \mi^k \, \si \quad \text{with} \quad c_0 = \sqrt{\frac{\pi}{2}} \, \erf(1) \approx 1.0562.
\end{equation*}
This lower-bound holds for most splats with a reasonable distance from the camera. Therefore, the moments of most splats to grow at least as fast as $\mi^k$. The exponential increase with $k$ causes numerical instability when using the moments as a medium descriptor. 

\paragraph{Power Moments}
Recall that the warping function $\hat{g}$ maps $[t_n, t_f]$ to $[0,1]$, is differentiable and strictly monotone. To avoid the exponential growth of the power moments with increasing $k$, we define moments that integrate the warped distance against the density: 
\begin{equation*}
\hat{m}_k = \int_{t_n}^{t_f} \hat{g}(t)^k \, \sigma(t) \, \dd t.    
\end{equation*}
By linearity, the k-th power moment can be written as the sum of power moments for the individual particles:
\begin{equation*}    
\hat{m}_k = \sum_i \wi \, \int_{t_n}^{t_f} \hat{g}(t)^k \, e^{-\frac{(t - \mi)^2}{2\si^2}} \, \dd t.
\end{equation*}
The inner integral is solvable in elementary functions and $\textrm{erf}$ for many choices of $\hat{g}$ but can be solved if $\hat{g}$ is a polynomial.~\cite{gradshteyn2014table} Our choice of warping function and most functions that are suitable to map perceptually significant distances to well-resolved floating-point ranges do not permit for a practical closed-form solution. However, each particles contribution to the moment is concentrated around its mean, we therefore use a first-order Taylor approximation of the warped distance: ${u_t = \hat{g}(\mi) + \hat{g}'(\mi) \, (t - \mi)}$ which reduces the problem to a polynomial. Since each integral is to be solved individually we center the linearization at $\mi$.
The moments over the warped distance is therefore approximated as
\begin{equation*}
    \hat{m}_k \approx \sum_i \underbrace{\wi \, \int_{t_n}^{t_f} u_t^k \, e^{-\frac{(t-\mi)^2}{2\Sigma_i^2}} \, \dd t}_{(\hat{m}_k)_i}
\end{equation*}

We derive a recurrence relationship in order to evaluate the inner integral.

Pulling out one factor of the linearized distance from the exponent $u_t^k$ and substitute by is definition, gives
\begin{align*}
    \wi \, \int_{t_n}^{t_f} (\hat{g}(t) + \hat{g}'(\mi) \, (t-\mi)) \, u_t^{k-1} \, e^{-\frac{(t-\mi)^2}{2\si^2}} \, \dd t.
\end{align*}
Expand the integral into two parts: 
\begin{equation}
    \label{eq:recurrence}
    \begin{split}
    (\hat{m}_k)_i = & \ \hat{g}(\mi) \, \wi \int_{t_n}^{t_f} u_t^{k-1} \, e^{-\frac{(t-\mi)^2}{2\si^2}} \, \dd t \\
    &+ \hat{g}'(\mi) \, \wi \int_{t_n}^{t_f} (t-\mi) \, u_t^{k-1} \, e^{-\frac{(t-\mi)^2}{2\si^2}} \, \dd t.
    \end{split}
\end{equation} 
By definition, the first part of the expanded integral is $\hat{g}(\mu_i) (\hat{m}_{k-1})_i$. The remaining integral, $I$, can be expressed in term of the total derivative of the Gaussian since
\begin{align*}    
    &\frac{\partial}{\partial t} e^{-\frac{(t-\mi)^2}{2\si^2}} = -\frac{t-\mi}{\si^2} \, e^{-\frac{(t-\mi)^2}{2 \si^2}}\\
    &\Rightarrow (t-\mi) \, e^{-\frac{(t-\mi)^2}{2\si^2}} = -\si^2 \, \frac{\partial}{\partial t} e^{-\frac{(t-\mi)^2}{2\si^2}}.
\end{align*}
Hence 
\begin{equation*}
    I= -\hat{g}'(\mi) \, \wi \, \si^2 \, \int_{t_n}^{t_f} u_t^{k-1} \, \frac{\partial}{\partial t} \left(e^{-\frac{(t-\mi)^2}{2 \si^2}}\right) \, \dd t.
\end{equation*}
Use integration by parts with functions ${f(t) = u_t^{k-1}}$ and ${g(t) = e^{-\frac{(t-\mi)^2}{2\si^2}}}$ which implies $f'(t) = (k-1) \, g'(\mi) \, u_t^{k-2}$ since ${u_t' = \hat{g}'(\mi)}$.
\begin{align*}
    I = &-\hat{g}'(\mi) \, \wi \, \si^2 \left[u_t^{k-1}  e^{-\frac{(t-\mi)^2}{2\si^2}} \right]_{t_n}^{t_f} \\
    &+ \hat{g}'(\mi)^2 \, \si^2 (k-1) \, \underbrace{\wi \int_{t_n}^{t_f} u_t^{k-2} e^{-\frac{(t-\mi)^2}{2\si^2}} \, \dd t}_{(m_{k-2})_i}.
\end{align*}
Putting everything together, gives for ${ k\geq 2 }$, 
\begin{equation*}    
(\hat{m}_k)_i = \hat{g}(\mi) \, (\hat{m}_{k-1})_{i} + \hat{g}'(\mi)^2 \, \Sigma_i^2 \, (k-1) \, (\hat{m}_{k-2})_i - B_i(k)
\end{equation*}
where $B_i(k)$ is term to correct for density outside of the integration boundary
\begin{equation*}
B_i(k) = \hat{g}'(\mi) \, \wi \, \si^2 \left[u_t^{k-1}  e^{-\frac{(t-\mi)^2}{2\si^2}} \right]_{t_n}^{t_f}.
\end{equation*}
For the base case $(m_0)_i$, we have
\begin{equation*}    
(\hat{m}_0)_i = \wi \int_{t_n}^{t_f} e^{-\frac{(t - \mi)^2}{2\,\si^2}} \, \dd t.
\end{equation*}
Applying change of variable with ${v = (t - \mi) /(\sqrt{2}\,\si)}$ which implies ${t = \mi + \sqrt{2}\,\si\,v}$, and ${\dd t = \sqrt{2}\,\si\,\dd v}$ gives
\begin{equation*}
(\hat{m}_0)_i = \wi \sqrt{2}\,\si \int_{v_n}^{v_f} e^{-v^2} \, \dd v
\end{equation*}
where the near bound is ${v_n = (t_n - \mi)/(\sqrt{2}\,\si)}$ and the far bound ${v_f = (t_f - \mi)/(\sqrt{2}\,\si)}$. Writing the antiderivative of the Gaussian function $e^{-v^2}$ in terms of the error function gives for any $C$
\begin{equation*}
\int e^{-v^2} \, \dd v = \frac{\sqrt{\pi}}{2} \erf\left(v\right) + C.
\end{equation*}
Thus, after simplfying constants 
\begin{equation}
\label{eq:zero_moment}
(\hat{m}_0)_i = \wi \si \sqrt{\frac{\pi}{2}} \left(\erf(v_f) - \erf(v_n) \right).
\end{equation}
The base case $(\hat{m}_1)_i$ follows directly from Equation \eqref{eq:recurrence}:
\begin{equation*}    
(\hat{m}_1)_i = \hat{g}(\mi) \, (\hat{m}_0)_i + \hat{g}'(\mi) \, \wi \, \int_{t_n}^{t_f} (t-\mi) \, e^{-\frac{(t-\mi)^2}{2 \si^2}} \, \dd t
\end{equation*}
Substitute the total derivative in the remaining integral gives
\begin{equation*}
\int_{t_n}^{t_f} (t-\mi) e^{-\frac{(t-\mi)^2}{2 \si^2}} \, \dd t = -\si^2 \int_{t_n}^{t_f} \frac{\partial }{\partial t} \, e^{-\frac{(t-\mi)^2}{2 \si^2}} \, \dd t.    
\end{equation*}
Thus by FTC,
\begin{equation*}
(\hat{m}_1)_i = \hat{g}(\mi) \, (\hat{m}_0)_i - \hat{g}'(\mi)\,\wi\,\si^2 \, \left[ e^{-\frac{(t - \mi)^2}{2\si^2}} \right]_{t_n}^{t_f}.
\end{equation*}
\textbf{Special case $t_f \to \infty$:} Since ${\lim_{t \to \infty} \erf(a+bt) =1}$ for fixed ${a,b > 0}$, we get the first recurrence basis presented in Equation 10 of the main paper:
\begin{equation*}
(\bar{m}_0)_i = \wi \si \sqrt{\frac{\pi}{2}} \left(1 - \erf\left(\frac{t_n - \mi}{\sqrt{2}\,\si}\right)\right).    
\end{equation*}
For ${t \geq 2 \mi}$. ${(t-\mi)^2 \geq (t/2)^2 = t^2/4}$. Hence
\begin{equation*}
    e^{-\frac{(t-\mi)^2}{2\si^2}} \leq e^{-\frac{t^2}{8\si^2}} \quad (t \geq 2\mi).
\end{equation*}
Since ${e^{-ct^2} \to 0}$ as ${t_f \to \infty}$ for any fixed $c > 0$, we get the second recurrence basis present in Equation 11 of the main paper:
\begin{equation*}
(\bar{m}_1)_i =\hat{g}(\mi) \, (\bar{m}_0)_i + \hat{g}'(\mi) \wi \Sigma_i^2 e^{-\frac{(t_n - \mu)^2}{2\Sigma_i^2}}.
\end{equation*}
Since ${(a+b\,t)^k\,e^{-c\,t^2} \to 0}$ as ${t \to \infty}$ for any fixed $k$ and $a,b,c > 0 $ by repeated L'Hopital, it follows that 
\begin{equation*}
    \lim_{t \to \infty} \left(\hat{g}(\mi) + \hat{g}'(\mi)(t -\mi)\right)^{k-1} e^{-\frac{(t_f - \mi)^2}{2 \si^2}} = 0.
\end{equation*}
The boundary term of the recurrence thus simplifies to the boundary term referenced in Equation 12 in the main paper: 
\begin{equation*}
    \bar{B}_i(k) = -\hat{g}'(\mi)\,\wi\,\si^2 \, v_n^{k-1} \, e^{-\frac{(t_n - \mi)^2}{2 \Sigma^2}}.
\end{equation*}

\paragraph{Trigonometric Moments}
An alternative to the polynomial basis for the moments is the Fourier basis for which we derive the trigonometric moments for a Gaussian density function:
\begin{equation*}
    \tilde{m}_k = \sum_{j=1}^N \underbrace{\wj \int_{t_n}^{t_f} (e^{(2\pi-\theta)i \hat{g}(t) })^k e^{-\frac{(t-\mj)^2}{2\sj^2}} \, \dd t}_{= (\tilde{m}_k)_j}
\end{equation*}
Let ${\alpha=k(2\pi-\theta)}$ be the angular frequency term. The inner integral becomes:
\begin{equation*}
(\tilde{m}_k)_j = \wj  \int_{t_n}^{t_f} e^{i\alpha \hat{g}(t)} e^{-\frac{(t-\mj)^2}{2\sj^2}} \, \dd t
\end{equation*}
Substitute the Taylor expansion for $\hat{g}(t)$ into the integral:
\begin{equation*}
(\tilde{m}_k)_j \approx \wj \int_{t_n}^{t_f} e^{i\alpha \left(\hat{g}(\mj) + \hat{g}'(\mj)(t - \mj)\right)} e^{-\frac{(t-\mj)^2}{2\sj^2}} \, \dd t    
\end{equation*}
Split the exponential term:
The term $e^{i \alpha \hat{g}(\mj)}$ is a constant with respect to $t$ and can be moved outside the integral. Let's also define a new constant ${\beta=\alpha\hat{g}'(\mj)}$
\begin{equation*}
(\tilde{m}_k)_j  \approx \wj \, e^{i\alpha \hat{g}(\mj)} \int_{t_n}^{t_f} e^{i\beta (t - \mj)} e^{-\frac{(t-\mj)^2}{2\sj^2}} \, \dd t
\end{equation*}
We now focus on solving the remaining integral, which we'll call $J$:
\begin{equation*}
J = \int_{t_n}^{t_f} e^{i\beta (t - \mj)} e^{-\frac{(t-\mj)^2}{2\sj^2}} \, \dd t
\end{equation*}
Perform a substitution: let ${u=t-\mj}$ which implies differential ${\dd u = \dd t}$. The limits of integration change to ${u \in [t_n-\mj,t_f-\mj]}$:
\begin{equation*}
J = \int_{t_n - \mj}^{t_f - \mj} e^{i \beta u} e^{-\frac{u^2}{2\sj^2}} \, \dd u = \int_{t_n - \mj}^{t_f - \mj} e^{-\frac{u^2}{2\sj^2} + i\beta u} \, \dd u.    
\end{equation*}
The exponent is a quadratic in $u$.
\begin{align*} 
-\frac{u^2}{2\sj^2} + i\beta u &= -\frac{1}{2\sj^2} \left[ u^2 - 2\sj^2 i\beta u \right]
\end{align*}
We can solve this integral by completing the square for the exponent with ${(i\sj^2 \beta)^2}$:
\begin{align*} 
-\frac{u^2}{2\sj^2} + i\beta u = -\frac{1}{2\sj^2} \left[ (u - i\sj^2 \beta)^2 - (i\sj^2 \beta)^2 \right]
\end{align*} 
Simplifying terms using ${i^2 = -1}$ gives the exponent
\begin{align*} 
-\frac{u^2}{2\sj^2} + i\beta u = -\frac{(u - i\sj^2 \beta)^2}{2\sj^2} - \frac{\sj^2 \beta^2}{2}.
\end{align*}
Substituting this exponent back into the integral $J$:
\begin{equation*}
J = \int_{t_n - \mj}^{t_f - \mj} e^{-\frac{(u - i\sj^2 \beta)^2}{2\sj^2} - \frac{\sj^2 \beta^2}{2} } \, \dd u.    
\end{equation*}
The term ${e^{-2\sj^2\beta^2}}$ is a constant and can be factored out:
\begin{equation*}
J = e^{-\frac{\sj^2 \beta^2}{2}} \int_{t_n - \mj}^{t_f - \mj} e^{ -\frac{(u - i\sj^2 \beta)^2}{2\sj^2} } \, \dd u.   
\end{equation*}
To match this form, we use another substitution: Let ${v= ({u-i\sj^2\beta})/({\sqrt{2} \sj})}$ such that $v^2$ matches the exponent. The differential is ${\dd v = {(\dd u)}/{(\sqrt{2}\,\sj)}}$, so ${\dd u = \sqrt{2}\,\sj \, \dd v}$ and the new limits of integration become:
\begin{equation*}
v_n=\frac{(t_n-\mj)-i\sj^2\beta}{\sqrt{2}\sj} \textrm{ and } v_f = \frac{(t_f-\mj)-i\sj^2\beta}{\sqrt{2}\sj}.
\end{equation*}
Substitute these into $J$:
\begin{equation*}
J =\sqrt{2}\,\sj e^{-\frac{\sj^2 \, \beta^2}{2}} \int_{v_n}^{v_f} e^{-v^2} \, \dd v
\end{equation*}
Now, apply the definition of the definite integral using the error function:
\begin{equation*}
\int_{v_n}^{v_f} e^{-v^2} \, \dd v = \left[ \frac{\sqrt{\pi}}{2} \erf\left(v\right) \right]_{v_n}^{v_f}
\end{equation*}
Substitute this back into $J$ and simplifying constants:
\begin{equation*}
J = \sqrt{\frac{\pi}{2}}\sj e^{-\frac{\sj^2 \beta^2}{2}} \left(\erf(v_f) - \erf(v_n)\right)    
\end{equation*}
Finally, we combine all the parts. Recall that ${(m_k)_j \approx \wj e^{i\alpha \hat{g}(\mj)}\cdot J}$.
\begin{equation*}
(\tilde{m}_k)_j \approx\wj e^{i\alpha \hat{g}(\mj)}\left(\sqrt{\frac{\pi}{2}}\sj e^{-\frac{\sj^2\beta^2}{2}}\left(\erf\left(v_f\right) - \erf\left(v_n\right) \right)\right)    
\end{equation*}
Combine the exponential terms:
\begin{equation*}
(\tilde{m}_k)_j \approx e^{i\alpha \hat{g}(\mj)-\frac{\sj^2\beta^2}{2}}\wj \sj \sqrt{\frac{\pi}{2}} \left(\erf\left(v_f\right)-\erf\left(v_n\right)\right)    
\end{equation*}
\textbf{Special case $t_f \to \infty$:} Recall the definition of $v_f$ which has real part $\textrm{Re}(v_f) = (t_f - \mj) /(\sqrt{2} \, \sj)$ and imaginary part $\textrm{Im}(v_f) = -(\sj \, \beta)/\sqrt{2}$.  The error function $\erf(z)$ is defined by the integral:
\begin{equation*}
\text{erf}(z) = \frac{2}{\sqrt{\pi}} \int_0^z e^{-w^2} \dd w
\end{equation*}
The function $e^{-w^2}$ is analytic on the whole complex plane. This means the integral is path-independent; the value depends only on the endpoints. This allows us to split the integral into two parts:
\begin{equation*}
\text{erf}(x_f + iy) = \frac{2}{\sqrt{\pi}} \left[ \int_0^{x_f} e^{-t^2} \dd t + \int_{x_f}^{x_f+iy} e^{-w^2} \dd w \right].
\end{equation*}
The first term is simply the definition of the real error function $\text{erf}(x_f)$. As $x_f \to \infty$, this part converges to 1. Let's call the second term $K$. We parameterize the vertical path by $w=x_f +is$, where $s$ goes from $0$ to $y$. The differential is $\dd w = i \dd s$.
\begin{equation*}
K = \frac{2}{\sqrt{\pi}} \int_0^y e^{-(x_f + is)^2} (i \dd s)  
\end{equation*}
Expand the exponent and factor out the term $e^{-x_f^2}$, which does not depend on $s$:
\begin{equation*}
K = \frac{2i e^{-x_f^2}}{\sqrt{\pi}} \int_0^y e^{s^2} e^{-2\,i\,x_f\,s} \dd s
\end{equation*}
Now, we take the limit as ${x_f \to \infty}$. The term ${e^{-x_f^2}}$ goes to $0$. The integral ${\int_0^y e^{s^2} e^{-2ix_f s} \dd s}$ remains bounded. Since $y$ is a finite constant, ${e^{s^2}}$ is bounded on the interval $[0,y]$. The term ${e^{-2ix_f s}}$ is just an oscillation with magnitude 1. The integral of a bounded function over a finite interval is finite. Because the limit is a term that vanishes times a bounded term, the entire second term vanishes. Thus ${\erf(v_f) \to 1}$ as ${t_f \to \infty}$. Therefore the particle's k-th trigonometric moment presented in Equation 14 of the main paper is
\begin{equation*}
(\tilde{m}_k)_j \approx \wj \sqrt{\frac{\pi}{2}} \sj e^{i\alpha \hat{g}(\mj)-\frac{\sj^2\beta^2}{2}}\left(1-\erf\left(v_n\right)\right).
\end{equation*}
Note that the remaining exponential term is a damped oscillator.  Using euler's formula, the oscillator term can be evaluated as
\begin{equation*}
e^{-\frac{\sj^2 \beta^2}{2}} \left(\cos(\alpha \, \hat{g}(\mj)) + i \sin(\alpha \, \hat{g}(\mj))\right).
\end{equation*}
For $k \geq 1$, $v_n$ is a complex number, requiring the evaluation of the error function $\erf$ with complex argument
\begin{equation*}
\frac{(t_n-\mj)-i\sj^2\beta}{\sqrt{2}\sj} = \underbrace{\frac{t- \mj}{\sqrt{2}\sj}}_{=a} - i\underbrace{\frac{\sj\beta}{\sqrt{2}}}_{=b}  
\end{equation*}
Since this is non-trivial in a real-time setting, we approximate it with a first-order Taylor expansion around $a$ in the imaginary direction:
\begin{equation*}
\erf(a+ib) \approx \erf(a) + ib \left(\frac{2}{\sqrt{\pi}}e^{-a^2}\right).
\end{equation*}

\begin{figure}
\centering
\begin{subfigure}[t]{0.49\linewidth}
    \centering
    \includegraphics[width=\linewidth]{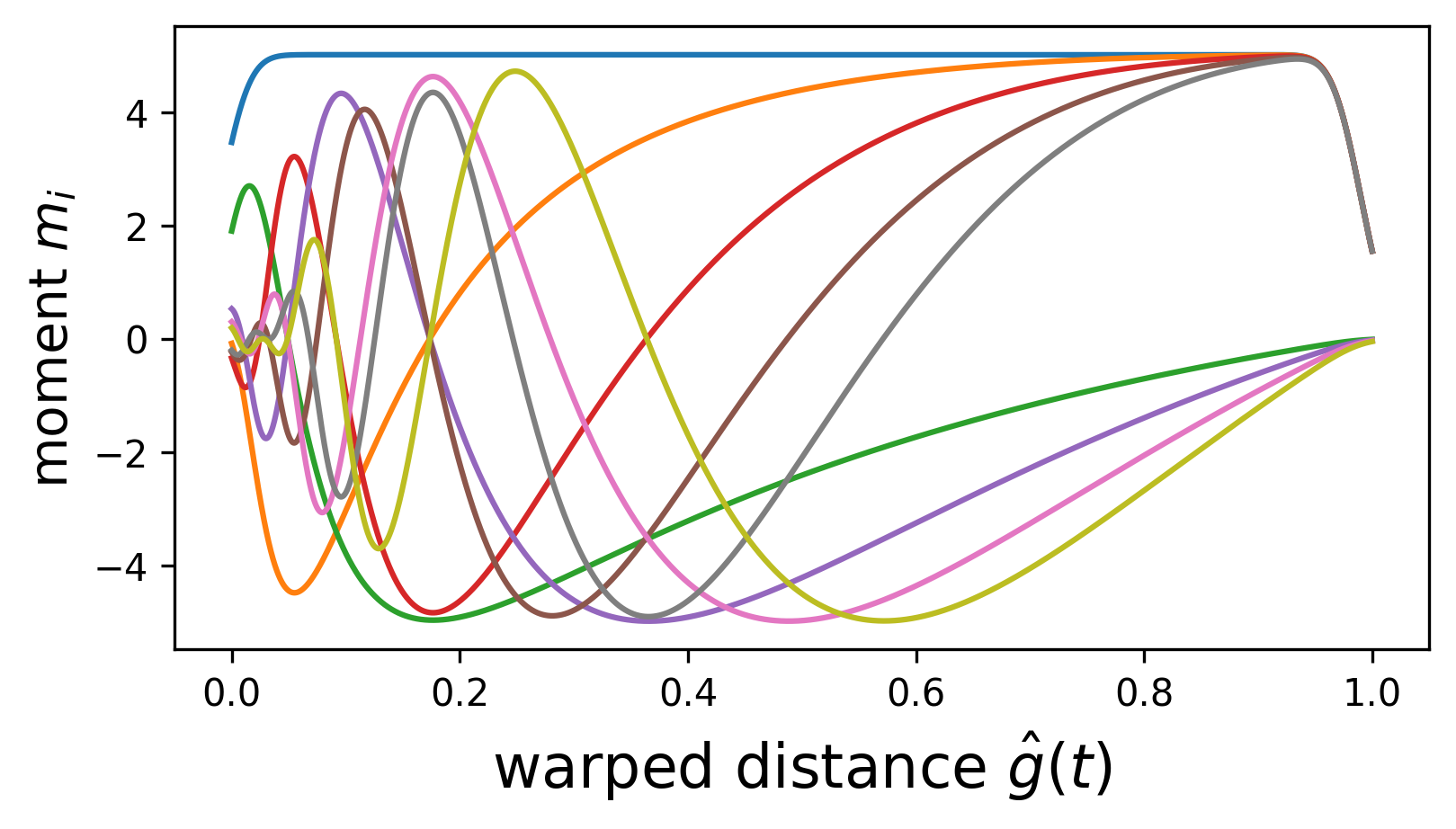}
    \caption{Exact $\erf$ evaluation}
\end{subfigure}
\begin{subfigure}[t]{0.49\linewidth}
    \centering
    \includegraphics[width=\linewidth]{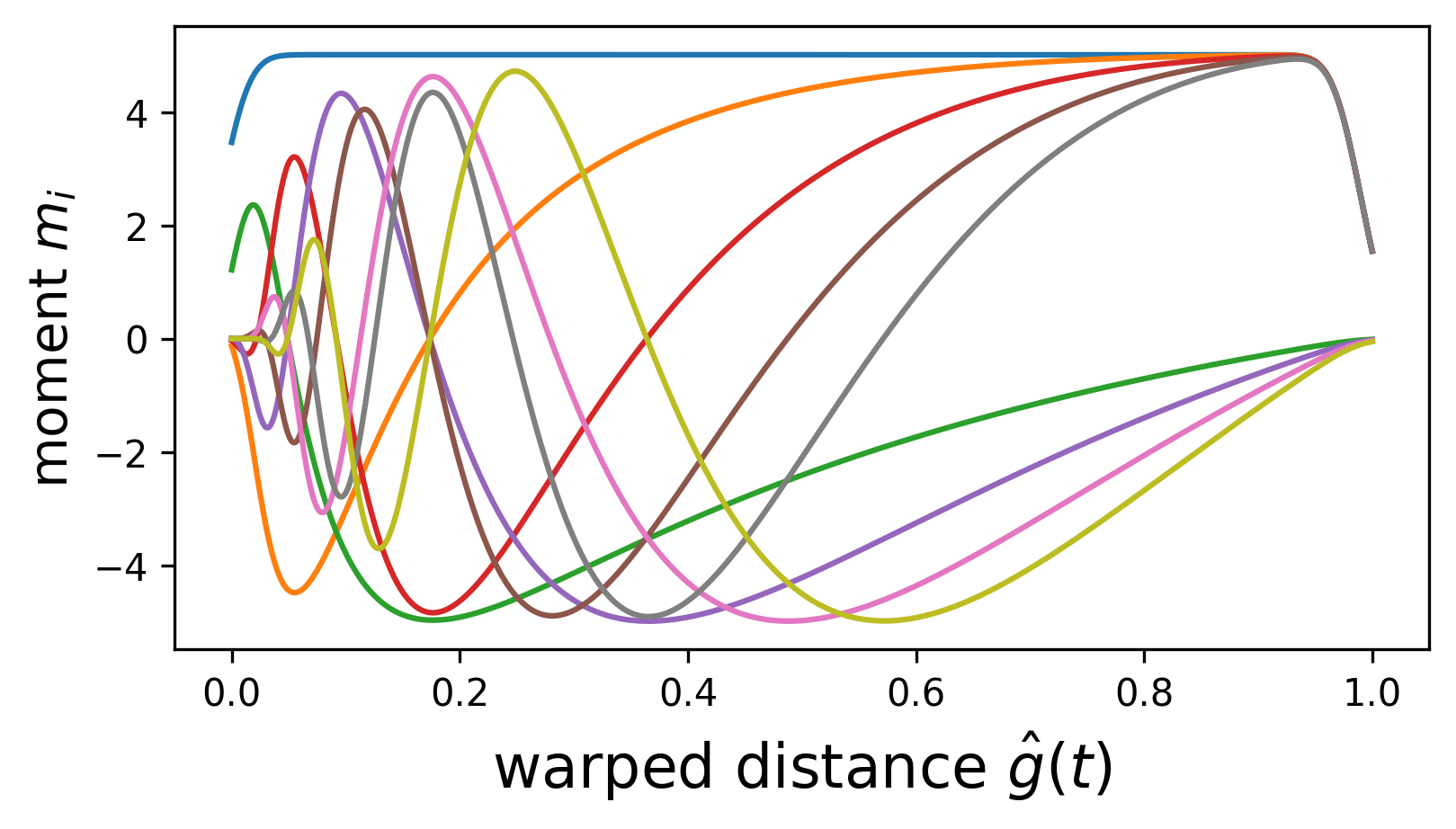}
    \caption{Linearized $\erf$ evaluation}
\end{subfigure}
\caption{Comparison of the trigonometric moment for a single particle, plotted as a function of $\mi$ with fixed particle parameters $\wi$, $\si$. (a) The exact moment, computed using the complex $\erf$ function. (b) Our first-order Taylor approximation. The error introduced by the approximation is visibly localized close to zero.}
\label{fig:trig_moments_erf_approx}
\end{figure}
The first-order expansion is valid when the imaginary component ${b = \sj\beta/\sqrt{2}}$ is small. Furthermore, the approximation error is inherently localized, as demonstrated in \cref{fig:trig_moments_erf_approx}. Since all higher-order terms of the Taylor series include the factor $e^{-a^2}$, the approximation error vanishes rapidly as $|a|$ increases, ensuring high fidelity for $t$ values distant from the mean $\mj$.

\paragraph{Comparison to MBOIT Moments}
\begin{figure}
\centering
\begin{subfigure}{0.49\linewidth}
    \centering
    \includegraphics[width=\linewidth]{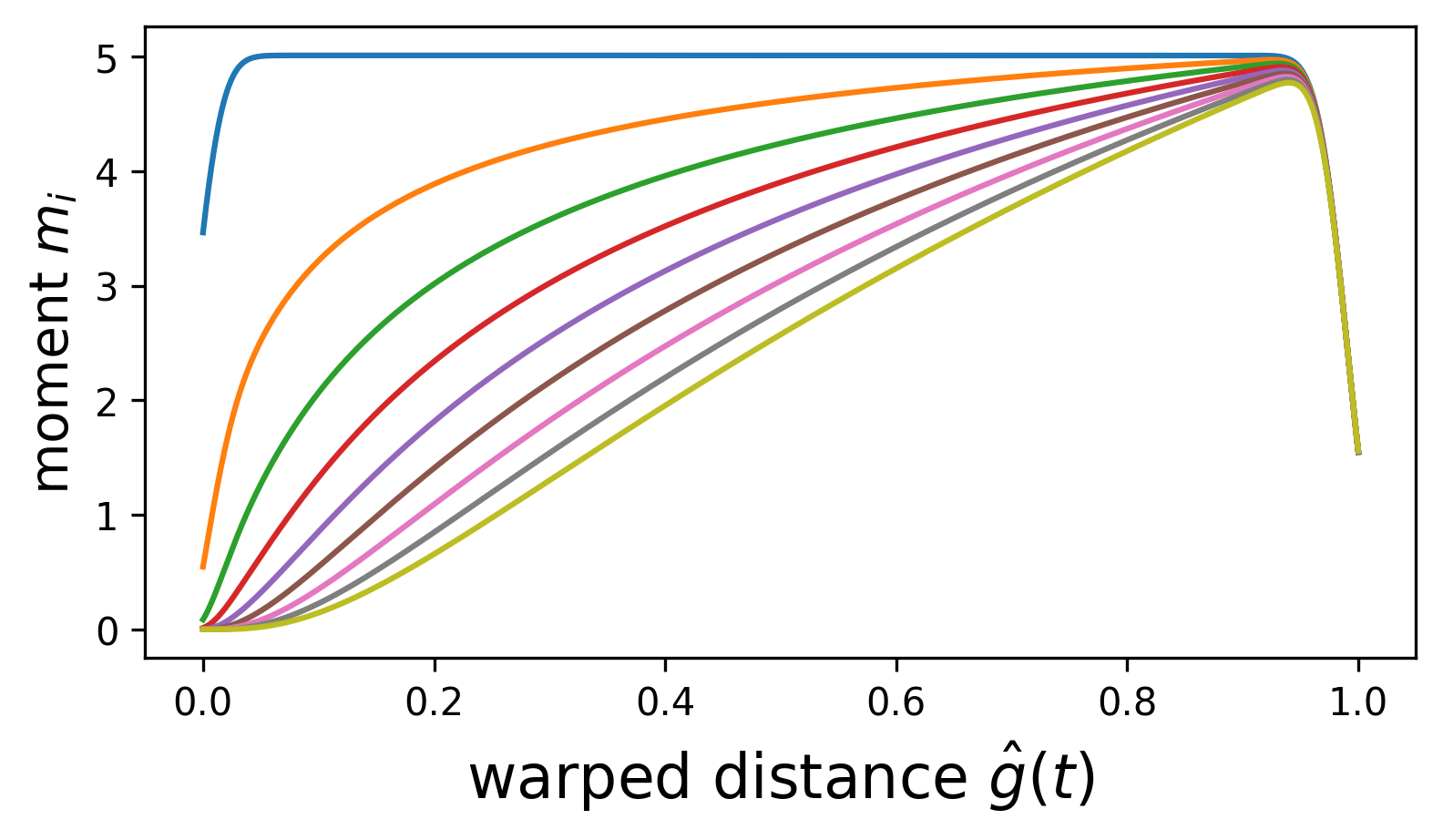}
    \caption{MBOIT power moments.}
\end{subfigure}%
\begin{subfigure}{0.49\linewidth}
    \centering
    \includegraphics[width=\linewidth]{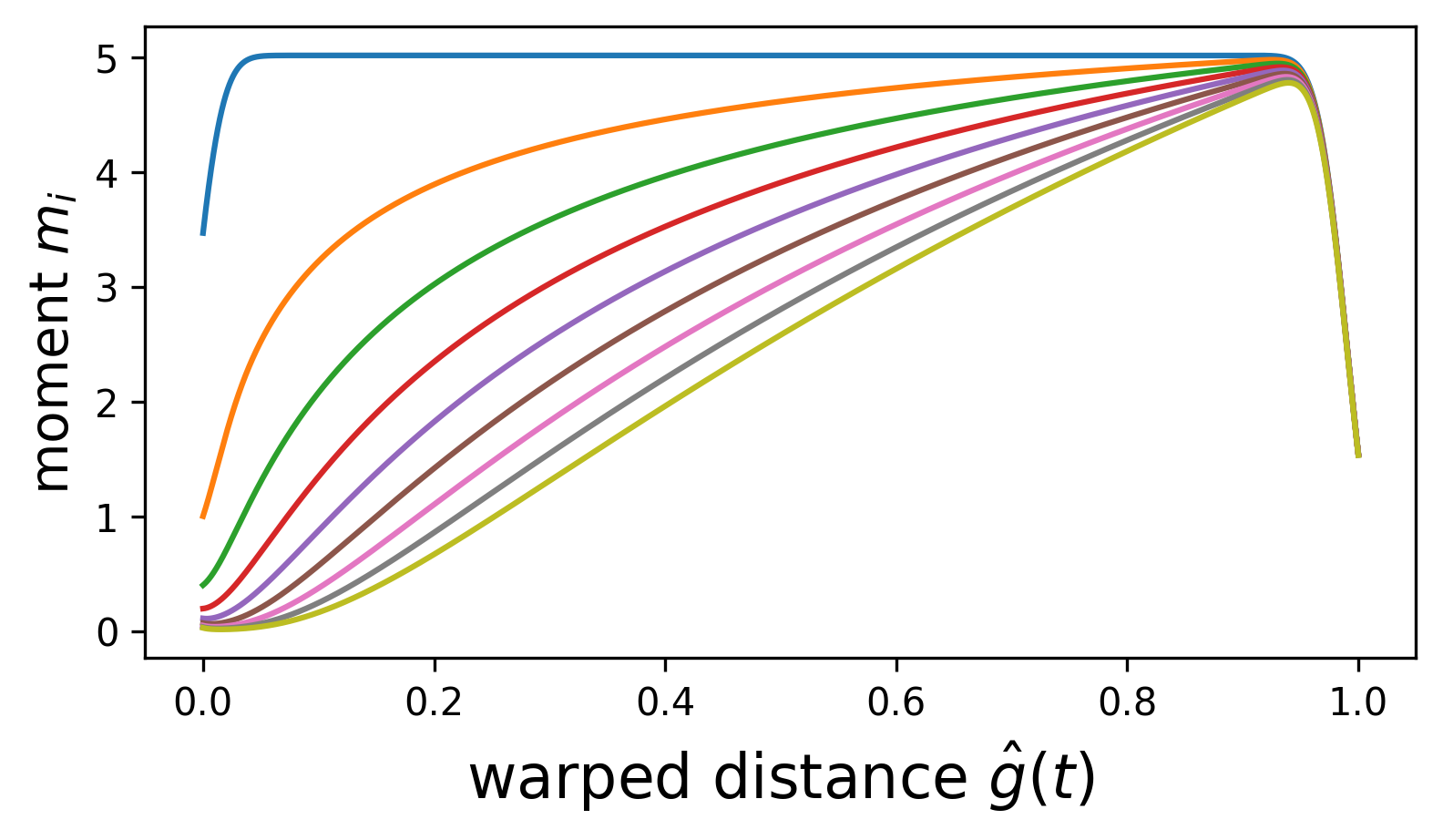}
    \caption{Our power moments.}
\end{subfigure} \newline
\begin{subfigure}{0.49\linewidth}
    \centering
    \includegraphics[width=\linewidth]{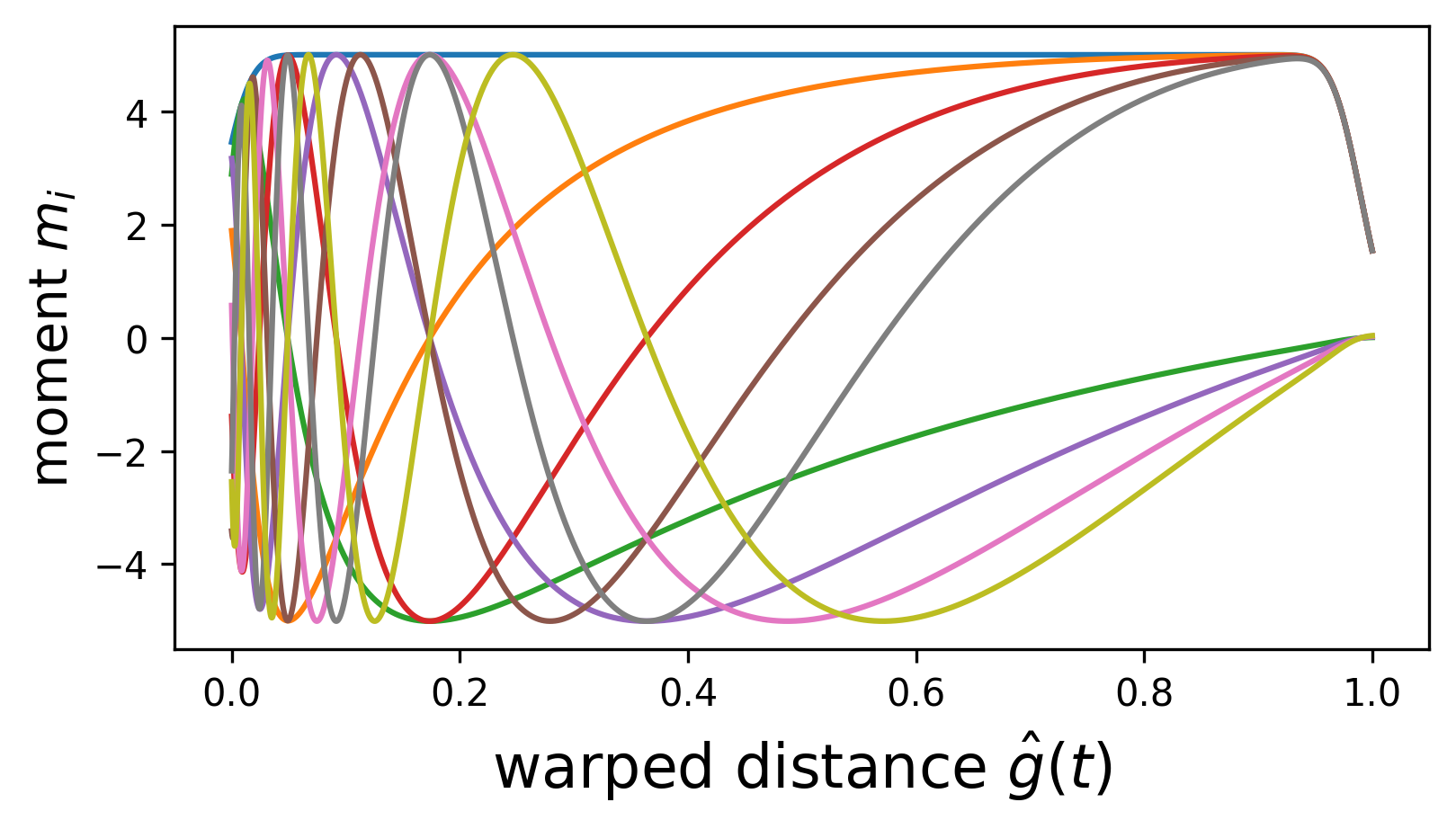}
    \caption{MBOIT trigonometric moments.}
\end{subfigure}%
\begin{subfigure}{0.49\linewidth}
    \centering
    \includegraphics[width=\linewidth]{figures/method/moments/trigonometric_moments_warped_distance_trig_moments.png}
    \caption{Our trigonometric moments.}
\end{subfigure}
\caption{Comparison of our Power and Trigonometric moments with the MBOIT method as a function of $\mi$. Our moments exhibit a clear dampening effect near zero, which becomes more pronounced for higher degrees, as shown by the $k=8$ case (orange). All moments are plotted for a single particle with fixed parameters-}
\label{fig:moments}
\end{figure}
%
Münstermann \etal~\cite{munstermann2018moment} propose moments for rendering infinitesimally thin transparent surfaces. Their power moments are defined as
\begin{equation*}    
m_k = \sum_j -\log(1-\alpha_j) \, z^k
\end{equation*}
and their trigonometric moments as
\begin{equation*}
\tilde{m}_k = \sum_j -\log(1-\alpha_j) e^{(2\pi - \theta ) \,i \, \frac{z+1}{2}}
\end{equation*}
where $z$ is the warped distance, $\alpha_j$ is the surface opacity, and $i$ is the imaginary unit. These moments can be adapted to our domain by setting the particle opacity $\alpha_j = 1-e^{-\bar{\tau}_j}$, where $\bar{\tau}_j$ can be evaluated using \cref{eq:zero_moment} with modified bounds. \cref{fig:moments} shows a comparison to our moments. Our moments exhibit a dampening effect that strengthens with increasing $k$. This dampening is observable in both power and trigonometric moments but is more pronounced for the latter.

\begin{figure}

    \def\zoomx{-0.15}
    \def\zoomy{0.05}
    \def\zoomboxx{1.55}
    \def\zoomboxy{-0.35}
    \def\zoomlevel{1.4}

    \newcommand{\subwidth}{0.49\linewidth}
    \centering
    \begin{subfigure}[t]{\subwidth}
        \centering
        \imagewithzoom{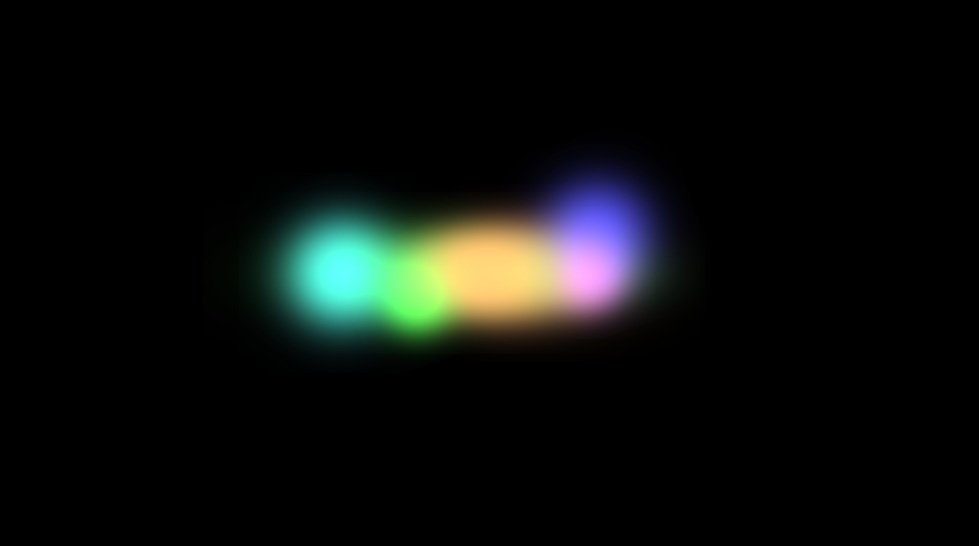}{\zoomx}{\zoomy}{\zoomboxx}{\zoomboxy}{1.0cm}{\zoomlevel}{220 150 200 170}
        \caption{MBOIT power moments}
    \end{subfigure}%
    \hfill
    \begin{subfigure}[t]{\subwidth}
        \centering
        \imagewithzoom{figures/evaluation/eval_image37_power_n3.png}{\zoomx}{\zoomy}{\zoomboxx}{\zoomboxy}{1.0cm}{\zoomlevel}{220 150 200 170}
        \caption{Our power moments}
    \end{subfigure} \newline
    \begin{subfigure}[t]{\subwidth}
        \centering
        \imagewithzoom{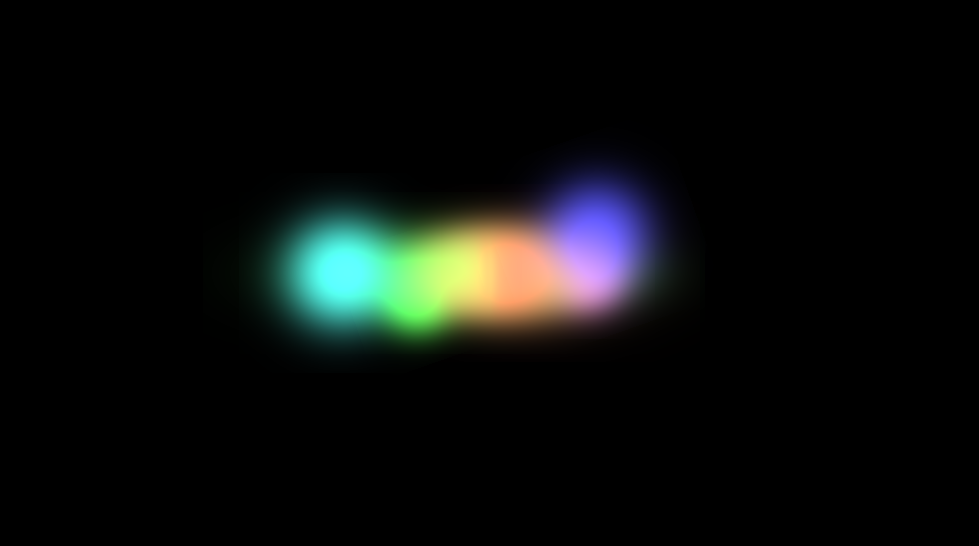}{\zoomx}{\zoomy}{\zoomboxx}{\zoomboxy}{1.0cm}{\zoomlevel}{220 150 200 170}
        \caption{MBOIT trigonometric moments}
    \end{subfigure}%
    \hfill
    \begin{subfigure}[t]{\subwidth}
        \centering
        \imagewithzoom{figures/evaluation/eval_image37_trig_n3.png}{\zoomx}{\zoomy}{\zoomboxx}{\zoomboxy}{1.0cm}{\zoomlevel}{220 150 200 170}
        \caption{Our trigonometric moments}
    \end{subfigure}
    \begin{subfigure}[t]{\subwidth}
        \centering
        \imagewithzoom{figures/evaluation/expected_image37.png}{\zoomx}{\zoomy}{\zoomboxx}{\zoomboxy}{1.0cm}{\zoomlevel}{220 150 200 170}
        \caption{Ground truth}
    \end{subfigure}
    \caption{Visual comparison of MBOIT~\cite{munstermann2018moment} Power and Trigonometric Moments to our Moments on synthetic data. All method use $N =3$ quadrature intervals.}
    \label{fig:mboit_moments_compare}
\end{figure}
A comparison on synthetic data (\cref{fig:mboit_moments_compare}) reveals negligible differences between MBOIT power moments and our proposed power moments. Both formulations underestimate splat occlusion, resulting in excessive color blending relative to the ground truth. While MBOIT trigonometric moments achieve more accurate occlusion, they introduce visual artifacts, particularly at the transition between the green and red particles. Our trigonometric moments do not exhibit this behavior.
\section{Rasterisation and Adjoint Rendering}
\label{ap:rasterisation}
This section derives the geometric proxy for particle rasterization (Eqs. 17–21 in the main paper) and details the proposed rasterization pipeline. Furthermore, we describe a streamlined backward pass that employs adjoint rendering for gradient computation.

\paragraph{Confidence-Interval Rasterisation}
For the sake of clarity, we will assume that the Gaussian parameters $(\mean, \cov)$ are defined directly in the camera's coordinate system. The opacity with which the camera observes the Gaussian is given by the integral. For a single Gaussian component, the integral can be written as:
\begin{equation*}
\int_{t_n}^{\infty} T(t) \sigma(t) \, \dd t  = 1-e^{-\bar{\tau}}
\end{equation*}
Our geometric proxy is designed to encloses the confidence interval 
\begin{equation*}
    c =  1-e^{-\bar{\tau}}
\end{equation*}
for a given $c \in (0,1)$ in the camera's screen-space (we use $c = 0.01$ in all experiments). Rearrange term to be an implicit equation of the optical depth in an unbounded medium
\begin{equation*}
    -\log(1-c) = \w \s \sqrt{\frac{\pi}{2}} \left(1-\erf\left(\frac{t_n - \m}{\sqrt{2} \s}\right)\right).
\end{equation*}

To simplify the analysis, we assume the Gaussian is sufficiently far from the camera's near plane. That is  the Gaussian's mean has a distance to the ray's near bound greater than $\sqrt{32} \s$. Under this assumption we can neglect the effects of the $\erf{}$ term and the implicit equation becomes 
\begin{equation*}
    -\log(1-c) = \w \s \sqrt{2\pi}.
\end{equation*}
This assumption is justified since most visible particles already fulfill this assumption and for those which are closer to the cameras near plane neglecting the $\erf{}$ term only leads to an overestimation of its screen-space size. This might harm performance but does not lead to a wrong appearance of these splats.

Now, only $\w$ and $\s$ are functions of the ray direction. The weight-term $\w$ varies exponentially with how aligned $\dir$ is to the $\bm{v}$ in the $\invcov$-inner product, while $\s$ varies only by a bounded square root factor determined by the eigenvalues of the precision matrix $\invcov$. To avoid the complexity of analyzing the product of these two terms, the derivation replaces $\s$ by a practical approximation evaluated along the view vector. While not a strict upper-bound, this approximation is empirically justified because $\w$ decays exponentially as the ray direction $\dir$ diverges from $\bm{v}$; thus, contributions where the approximation loosens are negligible. We approximate the directional dependency as:
\begin{equation}    
    \dir^T \cov^{-1} \dir \approx \frac{1}{\lVert \bm{v} \rVert_2^2} \bm{v}^T \cov^{-1} \bm{v}
\end{equation}
Consequently, we substitute $\s \approx  \lVert \bm{v} \rVert_2 / \sqrt{\bm{v}^T \invcov \bm{v}}$, which leaves us with analyzing the implicit curve:
\begin{equation*}    
    -\log(1-c) = \w \frac{\lVert \bm{v} \rVert_2}{ \sqrt{\bm{v}^T \invcov \bm{v}}} \sqrt{2\pi}.
\end{equation*}

Let ${C = \left(-\log(1-c) \, \sqrt{\bm{v}^T \invcov \bm{v}}\right)/\left(\sqrt{2\pi \,  } \lVert \bm{v} \rVert_2\right)}$ capture all constants for a given confidence level. The implicit curve $\w = C$ can be expressed in terms of the ray direction $\dir$ by taking the logarithm:
\begin{equation*}
\log(C) = \log(w) - \frac{1}{2} \mean^T \invcov \mean + \frac{1}{2} \frac{\left(\dir^T\invcov \mean\right)^2}{\dir^T \invcov \dir}
\end{equation*}
Rearranging to isolate the terms dependent on $\dir$ yields:
\begin{equation*}
\frac{\left(\dir^T\invcov \mean\right)^2}{\dir^T \invcov \dir} = 2(\log(C) - \log(w)) + \mean^T \invcov \mean
\end{equation*}
The right-hand side is constant for a given level set. Let's call this constant $\kappa$. Letting $\bm{A} = \invcov$, the equation defining the surface of valid ray directions is:
\begin{equation*}
(\dir^T \bm{A} \mean)^2 - \kappa (\dir^T \bm{A} \dir) = 0
\end{equation*}

A pixel with coordinates $\bm{p} = (u,v)^T$ corresponds to a homogeneous vector $\bm{p}_{\text{hom}} = (u,v,1)^T$. The un-normalized ray direction is $\dir_{\text{un}} = \bm{K}^{-1}\bm{p}_{\text{hom}}$ where $\bm{K}$ is the intrinsic matrix of a perspective camera. Since the level set equation is scale-invariant with respect to $\dir$, we can substitute $\dir_{\text{un}}$ directly:
\begin{equation*}
(\bm{p}_{\text{hom}}^T (\bm{K}^{-1})^T \bm{A} \mean)^2 - \kappa(\bm{p}_{\text{hom}}^T (\bm{K}^{-1})^T \bm{A} \bm{K}^{-1} \bm{p}_{\text{hom}}) = 0
\end{equation*}
This is a quadratic form, $\bm{p}_{\text{hom}}^T W \bm{p}_{\text{hom}} = 0$, where the $3 \times 3$ symmetric matrix $\bm{W}$ defines the resulting conic section in the image plane:
\begin{equation*}
\bm{W} = (\bm{m}\bm{m}^T - \kappa \bm{M})
\end{equation*}
with vector ${\bm{m} = (\bm{K}^{-1})^T \bm{A} \mean}$ and matrix ${\bm{M} = (\bm{K}^{-1})^T \bm{A} \bm{K}^{-1}}$.

To prove that the quadratic form $\bm{p}_{\text{hom}}^T \bm{W} \bm{p}_{\text{hom}} = 0$ describes a real ellipse, we examine the eigenvalues of $-\bm{W} = \kappa \bm{M} - \bm{m}\bm{m}^T$. Since $\bm{M}$ is congruent to the positive definite precision matrix $\invcov$, the base matrix $\kappa \bm{M}$ has strictly positive eigenvalues $0 < \mu_1 \le \mu_2 \le \mu_3$. Applying Cauchy’s Interlacing Theorem for a rank-1 subtraction (Corollary 4.3.7 \cite{horn2012matrix}) yields eigenvalues $\lambda_i$ for $-\bm{W}$ that satisfy $\lambda_1 \le \mu_1 \le \lambda_2 \le \mu_2 \le \lambda_3 \le \mu_3$, which immediately guarantees that the two largest eigenvalues are positive ($\lambda_2, \lambda_3 > 0$). The sign of the smallest eigenvalue $\lambda_1$ is determined by the secular equation at zero, $f(0) = 1 - \bm{m}^T (\kappa \bm{M})^{-1} \bm{m}$, which reduces to $1 - \kappa^{-1} \mean^T \invcov \mean$. Because the camera center is assumed to be outside the Gaussian’s confidence interval ($\mean^T \invcov \mean > \kappa$), $f(0)$ is negative, forcing the smallest root $\lambda_1 < 0$. The resulting signature $(-, +, +)$ for $-\bm{W}$ implies a signature of $(+, -, -)$ for $\bm{W}$, thereby characterizing a real ellipse.

To transform the quadratic form into the standard representation of an 2D ellipse
\begin{equation*}
(\bm{p} - \mean_{2d})^T \invcov_{2d} (\bm{p} - \mean_{2d}) = 1,
\end{equation*}
such that $\bm{p} = (u,v)^T$ is on the ellipse, we first partition the conic matrix $\bm{W}$ into a $2 \times 2$ block $\bm{W}_{2\times2}$, a $2 \times 1$ vector $\bm{w}_{2\times1}$, and a scalar $w_{33}$:
\begin{equation*}
\bm{W} = \begin{pmatrix} \bm{W}_{2\times2} & \bm{w}_{2\times1} \\ \bm{w}_{2\times1}^T & w_{33} \end{pmatrix}.
\end{equation*}
Then, starting from the quadratic form ${\bm{p}_{\text{hom}}^T W \bm{p}_{\text{hom}} = 0}$, we expand the left term and use the symmetry of scalar product:
\begin{equation*}
\bm{p}^T \bm{W}_{2\times2} \bm{p} + 2 \bm{w}_{2\times1}^T \bm{p} + w_{33} = 0
\end{equation*}
Define the screen-space center ${\mean_{2d} = - \bm{W}_{2 \times 2}^{-1} \bm{w}_{2 \times 1}}$ and substitute ${\bm{w}_{2 \times 1}^T = -\mean_{2d}^T \bm{W}_{2 \times 2}}$:
\begin{equation*}
\bm{p}^T \bm{W}_{2\times2} \bm{p} - 2\mean_{2d}^T \bm{W}_{2 \times 2} \bm{p} + w_{33} = 0
\end{equation*}
Completing the square with ${\mean_{2d}^T \bm{W}_{2 \times 2} \mean_{2d}}$:
\begin{equation*}
(\bm{p} - \mean_{2d})^T \bm{W}_{2 \times 2} (\bm{p} - \mean_{2d}) - \mean_{2d}^T \bm{W}_{2 \times 2} \mean_{2d} + w_{33} = 0
\end{equation*}
Adding ${\mean_{2d}^T \bm{W}_{2 \times 2} \mean_{2d} - w_{33}}$ and dividing by the same scalar gives
\begin{equation*}
    (\bm{p} - \mean_{2d})^T \frac{\bm{W}_{2 \times 2}}{\mean_{2d}^T \bm{W}_{2 \times 2} \mean_{2d} - w_33} (\bm{p} - \mean_{2d}) = 1.
\end{equation*}
Finally, define the new "covariance" matrix for the ellipse to be ${\cov_{2d} = (\mean_{2d}^T \bm{W}_{2 \times 2} \mean_{2d} - w_{33}) \bm{W}_{2\times 2}^{-1}}$, where its inverse is given by
\begin{equation*}
    \invcov_{2d} = \frac{\bm{W}_{2 \times 2}}{\mean_{2d}^T \bm{W}_{2 \times 2} \mean_{2d} - w_33}.
\end{equation*}

To find the tightest oriented bounding box for the ellipse defined by $(\mean_{2d}, \cov_{2d})$, we perform an eigen-decomposition of $\cov_{2d}$. The box is centered at $\mean_{2d}$ and spanned by the semi-axes $\bm{v}_i = \sqrt{\lambda_i} \bm{e}_i$, where $(\lambda_i, \bm{e}_i)$ are the eigenpairs of $\cov_{2d}$.

\paragraph{Adjoint rendering}
Since the rasterisation stage is implemented using hardware-accelerated rasterisation, full auto-differentiation through all operations is not available. Instead, adjoint rendering first computes adjoint moments which allows to compute the gradients for particle parameters in a single reduction pass.

The forward pass processes camera matrices $V, K$ and Gaussian parameters $\Theta$, illustrated in Fig 1. of the main paper. A Moment pass first rasterizes all Gaussians; a pixel shader computes the moment vector $\bm{m}_i$ for each splat, which are accumulated in an additive framebuffer to obtain the per-pixel moment texture $\bm{m}$. A subsequent Quadrature pass, using $V, K, \Theta$, and $\bm{m}$, re-rasterizes the Gaussians. Its pixel shader evaluates the radiance quadrature $L_i$ and consistency penalty $P_i$, which are accumulated to produce the observed radiance $L$. A Rescaling pass then uses a screen-filling quad to pixel-wise rescale $L$ using the first moment $\bm{m}_0$ for correct opacity, yielding the final output radiance $\hat{L}$. We omit the culling stage description for brevity, as non-visible Gaussians simply receive a zero gradient. The resulting image $\hat{L}$ and per-pixel penalty values are passed to PyTorch to compute the final loss $\mathcal{L}$.

A naive backward pass inverts the forward operations, see Fig 1. in the main paper, assuming all shader operations are differentiated via automatic differentiation. Starting from the input gradients $\partial \mathcal{L}/\partial \hat{L}$ and $\partial \mathcal{L} / \partial P$, the backward Rescaling pass computes $\partial \mathcal{L}/\partial L$ and $\partial \mathcal{L}/\partial \hat{L} \cdot \partial \hat{L}/\partial \bm{m}_0$. The trivial derivative of the additive framebuffer accumulation ($\partial V / \partial V_i = 1$ for a contributing splat $i$) is implicitly handled by the backward rasterization, which gathers the per-pixel textures $\partial \mathcal{L}/\partial L$ and $\partial \mathcal{L} / \partial P$ at pixels covered by each splat.
The backward Quadrature pass computes the contribution to the adjoint moments, $\sum_i \frac{\partial \mathcal{L}}{\partial L_i} \frac{\partial L_i}{\partial \bm{m}} + \frac{\partial \mathcal{L}}{\partial P_i} \frac{\partial P_i}{\partial \bm{m}}$, accumulating them in an additive framebuffer. It also computes the derivative of the quadrature and penalty terms w.r.t. the splat parameters, which are reduced over all pixels covered by the splat's geometry: $\sum_{u,v} \frac{\partial \mathcal{L}}{\partial L_i} \frac{\partial L_i}{\partial \Theta_i} + \frac{\partial \mathcal{L}}{\partial P_i} \frac{\partial P_i}{\partial \Theta_i}$.
Finally, the backward Moment pass takes the adjoint moment texture from the quadrature pass and $\partial \mathcal{L}/\partial \hat{L} \cdot \partial \hat{L}/\partial \bm{m}_0$ as input. It evaluates and reduces the derivative of the density moment vectors w.r.t. the splat parameters over all covered pixels: $\sum_{u,v} \frac{\partial \mathcal{L}}{\partial \bm{m}_i}\frac{\partial \bm{m}_i}{\Theta_i}$.

The naive backward pass presents several issues: it requires two reduction operations, creates intermediate adjoint textures for all forward pass framebuffers, and necessitates an additional framebuffer for $\partial \mathcal{L} /\partial \hat{L} \cdot \partial \hat{L} / \partial \bm{m}_0$. We simplify this by observing that the derivative of the per-pixel opacity re-scaling from the Rescaling pass can be pulled into the other two stages, as it is evaluable from the forward pass results $L$ and $\bm{m}$. This optimization eliminates three additive framebuffer objects. Furthermore, given the adjoint moments, the per-splat gradients can be accumulated in a single pass, avoiding two separate, expensive reduction operations.

\begin{figure*}
    \centering
    \vspace{-0.8cm}
    \includegraphics{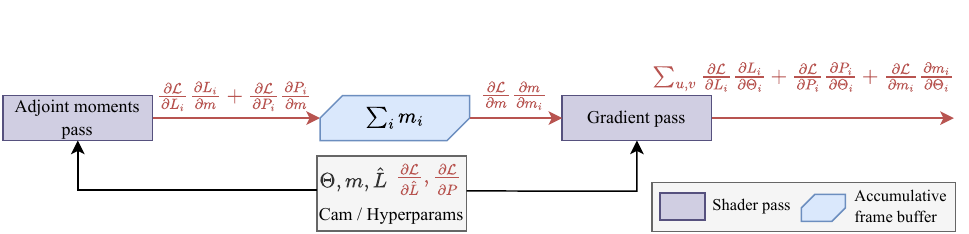}
    \caption{Architecture of the adjoint rendering pass. The pipeline executes in two passes: the computation of the adjoint moments, followed by the calculation of the Gaussian parameter gradients. The backward step for opacity rescaling is folded into both passes for computational efficiency. Inputs include the Gaussian parameters $\Theta$, the forward pass outputs (moment texture $\bm{m}$ and radiance map $\bm{L}$), and the upstream gradients $\partial \mathcal{L} / \partial \hat{L}$ and $\partial \mathcal{L} / \partial P$. }
    \label{fig:our-bwd}
\end{figure*}

Our optimized backward pass therefore consists of two stages, illustrated in \cref{fig:our-bwd}  First, an Adjoint Moment pass takes $V, K, \Theta$, the forward results $\bm{m}, L, P$, and the upstream gradients $\partial \mathcal{L} / \partial L, \partial \mathcal{L} / \partial P$ as input. It rasterizes all Gaussians, computing a per-splat, per-pixel adjoint moment vector $\delta \bm{m}_i$, which is then accumulated into a per-pixel texture $\delta \bm{m}$. Second, a Gradient pass inputs all arguments from the previous pass, plus the adjoint moment texture $\delta \bm{m}$. It re-rasterizes all Gaussians, where the pixel shader computes the per-splat gradient and performs the reduction over all pixels covered by the splat's proxy geometry, yielding the final gradient $\nabla_{\Theta_i}$:
\begin{equation*}
\nabla_{\Theta_i} = \sum_{u,v} \frac{\partial \mathcal{L}}{\partial L_i} \frac{\partial L_i}{\partial \Theta_i} + \frac{\partial \mathcal{L}}{\partial P_i} \frac{\partial P_i}{\partial \Theta_i} + \frac{\partial \mathcal{L}}{\partial \bm{m}_i}\frac{\partial \bm{m}_i}{\Theta_i}    
\end{equation*}
\section{Optimisation}
\label{ap:optimisation}

We derive the penalty term enforcing consistency between density and moment-based transmittance. Additionally, we detail the adaptation of adaptive density control to our volumetric framework.

\paragraph{Regularisation}
The moment-based reconstruction of the transmittance function is limited in its complexity by the number of moments provided to the reconstruction algorithm. Consequently, the reconstructed transmittance over- or might underestimates its true value along rays with highly varying density. This allows the optimization process to converge to minima which overfit on the training views.

This behaviour is avoided by extending the 3DGS trainings objective with an additional regularisation objective to ensure consistency between the predicted transmittance and the actual density distribution along each ray. 
\begin{equation*}
    \mathcal{L} = (1-\alpha) \mathcal{L} + \alpha \mathcal{L}_\textrm{D-SSIM} + \lambda \mathcal{L}_\textrm{consistency}
\end{equation*}

Recall that the density is a sum of individual density values. Thus we can rewrite the transmittance on any interval $[t_j, t_{j+1}]$ as the product of transmittance values for the density of the i-th particle on the same interval:
\begin{equation*}
    T(t_j \to t_{j+1}) = \prod_i e^{-\int_{t_j}^{t_{j+1}} \sigma_i(s) ds} = \prod_i T_i(t_j \to t_{j+1}).
\end{equation*}
where $T_i(t_j \to t_{j+1})$ is the transmittance when solely considering the density of the i-th particle. Since $\sigma_i(x_t) > 0$ for any $t$, we have in particular the inequality
\begin{equation*}
    T(t_j \to t_{j+1}) < T_i(t_j \to t_{j+1})
\end{equation*}
which hold for any interval $[t_j, t_{j+1}]$ and any particle. 

This inequality is the basis of our consistency regularization that we apply to each interval used in the quadrature. However, our order-independent transmittance  predicts the transmittance from $t_n$ to any $t$. We therefore reframe the inequality to only involve the transmittance from $t_n$ to the interval edges by first multiply both sides of the inequality by ${T(t_n \to t_j)}$:
\begin{equation*}
   T(t_n \to t_j)\,T(t_j\to t_{j+1}) <  T(t_n \to t_j)\,T_i(t_j \to t_{j+1})
\end{equation*}
Using the multiplicative property of the transmittance, i.e. ${T(t_n \to t_j) T(t_j \to t_{j+1})}$ is equal to ${T(t_n \to t_{j+1})}$, substituting this into the inequality and dividing by ${T(t_n \to t_j)}$ again gives the inequality
\begin{equation*}
    \frac{T(t_n \to t_{j+1})}{T(t_n \to t_j)} <  T_i(t_j \to t_{j+1}).
\end{equation*}

The fraction between the transmittance values is numerically disadvantageous and can be avoided by rewriting the inequality in terms of the optical depth. The right side of the inequality can be solved analytically for each particle and each interval:
\begin{equation*}
    T_i(t_j \to t_{j+1}) = e^{-\tau_{ij}}
\end{equation*}
where $\tau_{ij}$ is the optical depth over the $j$-interval for the $i$-th particle which can be expressed closed-form via the error function. Further, applying the logarithm to both sides to the inequality and multiplying both sides with $-1$ gives an equivalent inequality for the optical depth
\begin{equation*}
    \tau(t_n \to t_{j+1}) - \tau(t_n \to t_j) \geq \tau_{ij}.
\end{equation*}

We reframe this inequality into a regularization term that penalizes inconsistencies between the optical depth predictions and the density of the i-th particle:
\begin{equation*}
    P_i = \sum_{j=1}^N \max\left(0, \tau_{ij} - \left(\tau(t_n \to t_{j+1}) - \tau(t_n \to t_j)\right) \right)^2
\end{equation*}
We use squared penalty to more strongly punish strong outliers. The penalty is evaluated over the fixed quadrature intervals $[t_j, t_{j+1}]$. The penalty loss then computes an average over all pixels and particles cover the pixel 
\begin{equation*}
    \mathcal{L}_\textrm{consistency} = \frac{1}{N_\textrm{pixel}} \sum_{u,v} \sum_{i} P_i
\end{equation*}

\paragraph{Adaptive Density Control}
To progressively increase the number of particles, we employ the adaptive density control (ADC) presented in the original 3DGS publication and introduce only modifications to make it consistent with our density-based medium. While more effective densification schemes have been presented, we adapt the original ADC formulation for better comparability to the 3DGS baseline and more recent methods which also adapt this formulation. If not state otherwise, our densification follows the steps outlined in \cite{kerbl20233d} and uses the same hyperparameters.

\textbf{View-Independent Opacity}
ADC uses the splats opacity parameter as the criterion whether or not to prune a particle. This criterion relies on the fact that 3DGS is amplitude preserving for all viewing directions \cite{celarek2025does}, that is, only the splats shape changes but its "transparency" is the same from all angles.

Recall that our medium parameterisation assigns each particle a maximal extinction instead of an opacity value. Furthermore, our volumetric rendering approach determines a particles visibility via a numerical integration. A key consequence of this approach is that an individual particle's opacity becomes a view-dependent quantity.

Here, even a particle with low extinction but high standard deviation might have a high visible contribution when viewed in direction of its longest axis. Contrary a particle with high extinction can have a negligible visual contribution when it standard deviation is small enough. Therefore, pruning based on particle's maximal extinction or eigenvalues alone does not yield satisfactory results. Instead, we use the particles highest opacity when viewed along its shortest axis through its center for the opacity pruning. We specifically use the shortest axis to reduce the potential for overfitting where thin elongated particles are created during optimization which only have a visible contribution from very specific viewing directions.

Recall that the particle's individual opacity along a ray while ignoring its surrounding medium is ${1-e^{-\tau(t_f)}}$ with
\begin{equation*}
     \tau(t_f) = \w \s \sqrt{\frac{\pi}{2}} \left(\erf\left(\frac{t_f - \m}{\sqrt{2} \s}\right) - \erf\left(\frac{t_n - \m}{\sqrt{2} \s}\right)\right)
\end{equation*}
The $\textrm{erf}$ terms describe the visibility falloff  the ray boundaries. For the view-independent opacity we can upper-bound this term by $2$, which corresponds to the boundary value for a particle that is in perfect view. The simplified term becomes
\begin{equation*}
1-e^{-\w \s \sqrt{2\pi}}.
\end{equation*}
where $\w$ and $\s$ are given by Eq. 5 and Eq. 6 in the main paper. 

The assumption that the particle is viewed along its shortest axis implies that the viewing direction $\dir$ is the unit eigenvector corresponding to the smallest eigenvalue $\lambda_\textrm{min}(\cov)$ of the covariance matrix $\cov$. Consequently, the quadratic form in the variance along this ray simplifies to:
\begin{equation*}
\dir^T \invcov \dir = \frac{1}{\lambda_\textrm{min}(\cov)}.
\end{equation*}
By construction of $\cov$ we have ${\lambda_\textrm{min}(\cov) = \min(s_x^2, s_y^2,s_z^2)}$ and therefore ${\s = \sqrt{\lambda_\textrm{min}(\cov)} = \min{(s_x,s_y,s_z)}}$.
Given that we view the particle through its center, the camera-to-particle direction $\bm{v}$ is co-linear with $\dir$, such that $\bm{v} = c \cdot \dir$ for $c \geq 0$. This co-linearity causes the exponent term in $w$ to vanish:
\begin{align*}
&-\frac{1}{2} \bm{v}^T \invcov \bm{v} + \frac{1}{2}\frac{\left(\dir^T \invcov \bm{v}\right)^2}{\dir^T \invcov \dir} \\
&= -\frac{1}{2} c^2 \dir^T \invcov \dir + c^2\frac{1}{2}\frac{\left(\dir^T \invcov \dir\right)^2}{\dir^T \invcov \dir} = 0.
\end{align*}
Therefore $\w$ simplifies to $w$. Combining these results yields the the view-independent opacity $o$: 
\begin{equation*}    
o = 1-e^{-\sqrt{2\pi}\,w\, \min(s_x,s_y,s_z)}.
\end{equation*}

We use the view-independent opacity to convert the initial point opacity into an initial density value when creating the scene from the colmap scan. By solving the above equation for the peak density $w$, we get
\begin{equation*}
    w_\textrm{init} = \frac{-\log(1- o_\textrm{init})}{\sqrt{2 \pi} \textrm{avg}{(s_x, s_y, s_z)}}
\end{equation*}
we observed that using the average instead of minimal scaling parameter slightly improved convergence.
The original ADC resets the opacity to a value slightly above the pruning threshold at regular intervals, similar to \cite{talegaonkar2025volumetrically}, we found that this does not improve our results and disabled the operation.

\textbf{Cloning and Splitting}
ADC compares an average of the screen-space gradient lengths to a threshold in addition to a size criteria in order to decide whether to clone or split a point. However, as a result of our modified rasterization approach, there are no screen-space gradients, which are originally used in ADC to determine weather to clone or split a point. Instead we only have access to the 3D gradient of the mean positions; where we observed stronger cancellation of opposing gradient directions compared to screen-space gradient which results in an overall lower magnitude. To compensate for these effects, we employ the strategy proposed in \cite{ye2024absgs} and accumulate the absolute value of each gradient component and use the norm of the accumulated directional gradient magnitudes as the decision criterion. Furthermore, we reduce the threshold from $2e-4$ to $1e-4$.

The cloning operation simply duplicates a selected particle without further changes to its parameters. This operation introduces a bias, since the cloned particle's color contribution becomes overly pronounced but can be corrected for by change the splats opacity parameter. However, previously introduced corrections~\cite{rota2024revising} are only valid for the alpha-blending used in 3DGS and do not extend to our volumetric setting. Fortunately, the concept that the contribution of a particle should not be increased by cloning, is easy to adapt to the density since it is a linear quantity. We correct for the bias by updating the peak density with: 
\begin{equation*}
    w \gets \frac{1}{2} w.
\end{equation*}
It is easy to verify that this correction avoids the over representation of cloned points. Without the correction, the cloning operations is appearance-wise equivalent to doubling a cloned particles density. That is the density of a cloned point would become
\begin{equation*}
    \sigma(t) = 2 \w e^{-\frac{(t-\m)}{2 \s^2}}.
\end{equation*}
Consequently, its opacity without taking the surrounding medium into consideration would be 
\begin{equation*}
    \int_{t_n}^{t_f} T(t_n \to t)\sigma(t) \dd t = 1-e^{2 \tau(t_f)}
\end{equation*}
where $\tau(t_f)$ is the particles optical depth. Recall 
\begin{equation*}
    \tau(t_f) = \w \Sigma \sqrt{\frac{\pi}{2}} \left(\textrm{erf}\left(\frac{t_f - \m}{\sqrt{2}\Sigma}\right) -\textrm{erf}\left(\frac{t_n - \m}{\sqrt{2}\Sigma}\right)\right).
\end{equation*}
and ${\w = w e^{-K}}$. Thus scaling the maximal extinction by half before cloning yields a density in which the cloned points are not visually over-represent.

The splitting operation samples the mean for two new points for each selected point and copies all other parameters from the original except for the scaling parameter. The scaling parameters of the newly created points are down-scaled by constant factor of $5/8$. These newly created point have a high likelihood to have a substantial overlap since they are drawn from a Gaussian distribution. As pointed out when addressing the clone-operation, creating overlapping points without rescaling the density introduces a significant bias that can destabilize the optimisation. We observed in experiments that scaling the particles size parameters by a constant factor was not sufficient to reduce the bias to get a stable training behaviour. Furthermore, the randomness within the splitting operation makes an analysis of the bias difficult.
We introduce a modified splitting operation which allows us to minimize the bias introduced by this operation: Each particle that satisfies the splitting-condition, is divided into two new particles along the direction $\dir_\textrm{split}$ of its eigenvector with the highest eigenvalue. The means of the new particles are shifted by a fixed offset $\delta > 0$ with differing sign along this direction:
\begin{equation*}
    \mean_\textrm{new} = \mean \pm \delta \dir_\textrm{split}
\end{equation*}
To minimize the change in opacity of the newly created points when viewed along $\dir_\textrm{split}$, we downscale the largest scale parameter that with $\gamma \leq 1$:  
\begin{equation*}
    \Sigma_\textrm{new} = \gamma \Sigma_\textrm{split}
\end{equation*}
All other parameters are simply copied from the split-up particle to the new ones. 
Splitting and modifying parameters only in one direction allows us to analyse the problem for 1D Gaussians. The density of the particle before splitting is referred to as
\[
    \sigma_\textrm{old}(t) = e^{-\frac{(t - \m)^2}{2 \s_\mathrm{split}^2}}
\]
while the density after splitting the particle in two is
\[
    \sigma_\textrm{new}(t) = e^{-\frac{(t - (\m - \delta))^2}{2 (\gamma \s_\mathrm{split})^2}} + e^{-\frac{(t - (\m + \delta))^2}{2 (\gamma \s_\mathrm{split})^2}}.
\]
The main source of bias is the increased visual contribution of the newly created points over the original point. We therefore choose $\delta, \gamma$ such that we minimize the changes to the opacity:
\[
    \min_{\delta, \gamma} \int_{-\infty}^{\infty} \left( T_\textrm{old}(t_n \to t) \sigma_\textrm{old}(t) - T_\textrm{new}(t_n \to t) \sigma_\textrm{new}(t)\right)^2 \dd t
\]
By substituting $ t' = \frac{t - \mu}{\Sigma_\mathrm{split}} $, we can simplify this problem. The density before splitting is a standard Gaussian
\[
    \sigma_\textrm{old}(t') = e^{-\frac{1}{2} t'^2}
\]
and the density after the splitting operation is
\[
    \sigma_\textrm{new}(t') = e^{-\frac{\left(t' - \frac{\delta}{\Sigma_\mathrm{split}}\right)^2}{2 \gamma^2}} + e^{-\frac{\left(t' + \frac{\delta}{\Sigma_\mathrm{split}}\right)^2}{2 \gamma^2}}
\]
(Note, that $\gamma$ is a dimensionless scaling parameter and thus directly applies to the original problem domain.)
Directly, solving this optimization problem would return the trivial solution (${\delta=0}$).
So instead, we constrain the optimization problem such that the confidence intervals $[-c \Sigma_\mathrm{new}, c \Sigma_\mathrm{new}] = [-c \gamma \Sigma_\mathrm{split}, c \gamma \Sigma_\mathrm{split}]$ of each split particle lies within the confidence interval $[-c \Sigma_\mathrm{split}, c \Sigma_\mathrm{split}]$ of the original particle.
In particular, we want to place the particles as far apart from each other as possible while staying within the confidence intervals, which leads to the following connection between $\gamma$ and $\delta'$:
\[
    \gamma = 1 - \frac{\delta}{c \Sigma_\mathrm{split}} \Rightarrow \delta = c \Sigma_\mathrm{split} \, (1-\gamma)
\]
We therefore substitute $\delta$ leading to
\[
    \sigma_\textrm{new}(t') = e^{-\frac{(t' - c \, (1-\gamma))^2}{2 \gamma^2}} + e^{-\frac{(t' + c \, (1-\gamma))^2}{2 \gamma^2}}
\]
and solve the above minimization problem with variables $c, \gamma$.
Unlike the original problem, shrinking the particle by $ \gamma $ always implies a corresponding shift away from the center.
A numerical optimization of this problem converges to the solution:
\[
    \gamma = 0.6385502815246582
    \text{ and }
    \delta = 0.6128153090966912
\]

\begin{table}[t]
    \centering
    \scriptsize
    \caption{Per-scene metrics for our presented method.}
    \begin{tabular}{c|ccccc}
        \toprule
        Scene & PSNR $ \uparrow $ & SSIM $ \uparrow $ & LPIPS $ \downarrow $ & \# Points & Time [h]\\
        \midrule
        Bicycle & 23.15	& 0.646 & 0.291 & 3\,404\,626 & \phantom{0}5.69 \\
        Bonsai & 31.17 & 0.938 & 0.196 & \phantom{0}\,951\,805 & \phantom{0}4.69 \\
        Counter & 28.13 & 0.900 & 0.197 & 1\,310\,940 & \phantom{0}9.98 \\
        Garden & 26.50 & 0.841 & 0.141 & 2\,046\,786 & \phantom{0}5.82 \\
        Flowers & 18.99 & 0.488 & 0.374 & 2\,766\,098 & 11.06 \\
        Stump & 24.51 & 0.682 & 0.286 & 1\,994\,725 & \phantom{0}4.86 \\
        Treehill & 20.95 & 0.520 & 0.365 & 3\,347\,791 & \phantom{0}7.83 \\
        Room & 30.88 & 0.923 & 0.202 & 1\,569\,865 & \phantom{0}4.90 \\
        Kitchen & 29.35 & 0.906 & 0.150 & 1\,645\,792 & \phantom{0}9.25 \\
        \midrule
        Train & 20.11 & 0.781 & 0.242 & 1\,675\,025 & \phantom{0}5.37 \\
        Truck & 24.25 & 0.869 & 0.145 & 1\,065\,513 & \phantom{0}2.83 \\
        \midrule
        Playroom & 29.60 & 0.908 & 0.242 & 1\,562\,347 & 14.85 \\
        DrJohnson & 28.67 & 0.891 & 0.254 & 3\,819\,750 & \phantom{0}6.21 \\
        \bottomrule
    \end{tabular}
    \label{tab:full_results}
\end{table}

\section{Results}
An overview of all reported metrics on a per-scene basis is reported in  \cref{tab:full_results}.

\end{document}